\documentclass[lettersize,journal]{IEEEtran}
\usepackage{amsmath,amsfonts}
\usepackage{algorithmic}
\usepackage{algorithm}
\usepackage{url}
\usepackage{graphicx}
\usepackage{amsfonts}
\usepackage{color}
\usepackage{stfloats}
\usepackage{color,soul}
\usepackage{amsmath}
\usepackage{multirow}
\usepackage{hhline}
\usepackage{hyperref}
\usepackage{cite}
\usepackage{amssymb}
\usepackage{amsthm}
\usepackage{mathtools}
\usepackage{multicol}
\usepackage{xcolor,colortbl}

\usepackage{blindtext}

\definecolor{Gray}{gray}{0.9}
\usepackage{tikz}
\newcolumntype{M}[1]{>{\centering\arraybackslash}m{#1}}

\usepackage{array}
\usepackage{textcomp}
\usepackage{stfloats}
\usepackage{url}
\usepackage{verbatim}
\usepackage{graphicx}
\usepackage{textcomp}

\usepackage{tabularx}
\usepackage{graphicx}
\usepackage{adjustbox}
\usepackage{tabularx}
\usepackage{paralist}
\usepackage{longtable}
\usepackage{tikz}
\usepackage{forest}
\usetikzlibrary{trees,positioning,shapes,shadows,arrows.meta}

\makeatletter
\let\savespace\@minipagetrue
\makeatother
\usepackage{array}

\usepackage{xspace}
\usepackage{booktabs}
\newcommand{\tabitem}{~~\llap{\textbullet}~~}
\usepackage{float}

\usepackage{xspace}
\usepackage{textcomp}
\usepackage{fancyhdr}
\usepackage{amssymb,amsmath, amsfonts}
\usepackage{textcomp} 
\usepackage{bbding}
\usepackage{pifont}
\newcommand{\cmark}{\ding{51}}%
\newcommand{\xmark}{\ding{55}}%

\usepackage{comment}

\usepackage{graphicx}

\usepackage{amssymb}

\usepackage{rotating}
\usepackage{ifthen}
\usepackage{keyval}
\usepackage{trig}
\usepackage{listings}
\usepackage[utf8x]{inputenc}
\graphicspath{ {./images/} }
\usepackage{colortbl}
\ifCLASSOPTIONcompsoc
    \usepackage[caption=false, font=normalsize, labelfont=sf, textfont=sf]{subfig}
\else
\usepackage[caption=false, font=footnotesize]{subfig}

\fi

\newcolumntype{L}{>{\centering\arraybackslash}m{3cm}}
\usepackage[all=normal,paragraphs=tight,leading=tight,
wordspacing=tight,
floats=tight,
bibnotes=tight,
indent=tight,lists=tight
]{savetrees}

\begin{document}

\title{AI and 6G into the Metaverse: Fundamentals, Challenges and Future Research Trends}

\author{Muhammad~Zawish,~\IEEEmembership{Student Member,~IEEE,} Fayaz Ali Dharejo,~\IEEEmembership{Student Member,~IEEE,} Sunder Ali Khowaja,~\IEEEmembership{Senior Member,~IEEE,} Kapal Dev,~\IEEEmembership{Senior Member,~IEEE,} Steven Davy,~\IEEEmembership{Member,~IEEE,} Nawab Muhammad Faseeh Qureshi,~\IEEEmembership{Senior Member,~IEEE}, Paolo Bellavista,~\IEEEmembership{Senior Member,~IEEE}
\thanks{M. Zawish and S. Davy are with Walton Institute for Information and Communication Systems Science, South East Technological University, Ireland, E-mails: \{muhammad.zawish, steven.davy\}@waltoninstitute.ie}
\thanks{F.A Dharejo is with CNIC, University of Chinese Academy of Sciences, Beijing, 100190, China. E-mail: fayazdharejo@cnic.cn }
\thanks{S.A. Khowaja is with Faculty of Engineering and Technology, University of Sindh, Jamshoro, Pakistan. E-mail: sandar.ali@usindh.edu.pk }

\thanks{K. Dev is associated with the Department of Computer science, Munster Technological University, Ireland and Institute of Intelligent Systems, University of Johannesburg, South Africa.  E-mail: kapal.dev@ieee.org}

\thanks{N. M. F. Qureshi is associated with the Department of Computer Education, Sungkyunkwan University, Seoul, South Korea. E-mail: faseeh@skku.edu}
\thanks{Paolo Bellavista is associated with the Department of Computer Science and Engineering, University of Bologna, 40136 Bologna, Italy. E-mail:paolo.bellavista@unibo.it}
}

\markboth{Journal of \LaTeX\ Class Files,~Vol.~14, No.~8, August~2021}%
{Shell \MakeLowercase{\textit{et al.}}: A Sample Article Using IEEEtran.cls for IEEE Journals}


\maketitle

\begin{abstract}
Since Facebook was renamed Meta, a lot of attention, debate, and exploration have intensified about what the Metaverse is, how it works, and the possible ways to exploit it. It is anticipated that Metaverse will be a continuum of rapidly emerging technologies, usecases, capabilities, and experiences that will make it up for the next evolution of the Internet. Several researchers have already surveyed the literature on artificial intelligence (AI) and wireless communications in realizing the Metaverse. However, due to the rapid emergence and continuous evolution of technologies, there is a need for a comprehensive and in-depth survey of the role of AI, 6G, and the nexus of both in realizing the immersive experiences of Metaverse. Therefore, in this survey, we first introduce the background and ongoing progress in augmented reality (AR), virtual reality (VR), mixed reality (MR) and spatial computing, followed by the technical aspects of AI and 6G. Then, we survey the role of AI in the Metaverse by reviewing the state-of-the-art in deep learning, computer vision, and Edge AI to extract the requirements of 6G in Metaverse. Next, we investigate the promising services of B5G/6G towards Metaverse, followed by identifying the role of \textit{AI in 6G networks} and \textit{6G networks for AI} in support of Metaverse applications, and the need for sustainability in Metaverse. Finally, we enlist the existing and potential applications, usecases, and projects to highlight the importance of progress in the Metaverse. Moreover, in order to provide potential research directions to researchers, we underline the challenges, research gaps, and lessons learned identified from the literature review of the aforementioned technologies. 
\end{abstract}


\section{Introduction}
\IEEEPARstart{I}{n} 2021, Metaverse started influencing the real-world due to: i) pandemic lifestyle, and ii) the announcements of Meta, Amazon, Apple, Netflix, and Google (MAANG) to release Metaverse-related features and projects for their users. Since then, Metaverse has sought sheer attention from both academia and industry. At the current point in time, the Metaverse can be generally viewed as a pool of extended reality (XR)\footnote{Note that important acronyms are defined in Table. \ref{table:Acronyms}} spaces in which humans and their digital counterparts interact in a fully immersive manner \cite{khan2022Metaverse,ning2021survey, lee2021all, lee2021creators}. According to a recent survey\footnote{https://www.pewresearch.org/internet/2022/06/30/the-Metaverse-in-2040/}, the majority of technology experts believe that by 2040 the Metaverse will be more refined and seamless in its operation so that people around the world will be able to fully engage in its full immersion capabilities as an integral part of their daily lives. In essence, Metaverse is anticipated to integrate all essential aspects of cyberspace or the world wide web, such as B5G/6G, cloud and edge computing, social media, online gaming, augmented reality (AR), virtual reality (VR), cryptocurrencies, and artificial intelligence (AI)/machine learning (ML)/deep learning (DL) platforms and applications, to allow users to interact virtually \cite{khan2022Metaverse,bhattacharya2022Metaverse,gadekallu2022blockchain, cao2022decentralized, ilyina2022Metaverse}. 

Some of the early stage applications of Metaverse, namely Roblox\footnote{https://www.roblox.com/}, VRChat\footnote{https://hello.vrchat.com/}, Zepeto\footnote{https://zepeto.me/} or Second Life\footnote{https://secondlife.com/} have been allowing users to live in “different" or simulated lives, such as making friends and socializing with new avatars. These platforms have incorporated AR, VR and MR as a few elements of the Metaverse. The VR technology replaces the real world around users with a computer-generated digital scene using various software and communication devices, such as a head-mounted display (HMD). While in AR, the virtual world is seamlessly connected with the real world in order to create new interactive experiences. Finally, MR emerges as the combination of AR and VR and their underlying technologies. It is also noteworthy to mention that the popularity of these technologies has led to the universal availability of AR and VR equipment at reasonable prices while continually improving the quality of experience (QoE) for their users \cite{vretos2019exploiting}. 

However, in order to achieve the processing power and communication speed that are required by VR services to deliver smooth and immersive experiences, most HMDs still require the users to be tethered to a PC or gaming console. The only device to break free from this cable constraint is Meta's Oculus Quest 2\footnote{https://store.facebook.com/ie/quest/products/quest-2/}. \textit{Over the next few years, a number of improvements will be made to visual content and the untethered experience of mobile devices as processors get faster and wireless communication technologies to become latency-free.} In particular, with the existence of 5G, there will be a proliferation of devices connected to the network, which will have a profound impact on the growth of Metaverse \cite{lim2022realizing}. 5G has enabled real-time communication and information exchange among all of the connected devices with its low latency. It is evident that 5G offers faster speeds than 4G, but most importantly, it offers a variety of other benefits that go beyond speed alone \cite{torres2020immersive}. Specifically, Metaverse developers will be able to benefit from 5G's low latency by creating applications that can transmit $360^\circ$ content in near real-time. 

Metaverse will also proliferate the trend of human-centric/data-centric intelligent systems. This trend poses a number of constraints on existing 5G communication systems, making them less efficient and unreliable. For example, considering the 0.1ms delay requirements of haptic-based Metaverse applications such as teleportation or teleoperations, 5G can only provide the air interface latency of $<$1ms, which becomes inconsiderable for such applications \cite{al2018experimental, polachan2022assessing,tariq2022toward,aijaz2016realizing}. Moreover, according to Cisco \cite{barnett2018cisco}, mobile data traffic has grown 17-fold over the past five years and is expected to continue to grow. In particular, as of 2022, the 5G traffic will comprise 12\% of the total mobile traffic in the world. 5G networks were designed to be able to cover large areas of the spectrum, such as millimeter-waves (up to 300 GHz), and as a result, can handle large quantities of wireless traffic \cite{ana2021study,ghoshal2022can, roh2014millimeter, lin2021wireless}. 

Since Metaverse will bring several applications that may require data rates that are higher than Tbps, which is not likely to be the case with mmWave systems, such as holographic telepresence (HT), haptic sensory communications, brain-computer interface (BCI), and XR \cite{Gu2020,maier2022art,hilty2020review}. As a result, researchers have resorted to exploring the Terahertz (THz) frequency band (0.1-10 THz) in order to achieve the objectives of Tbps data rates. 6G communication system is expected to provide 1Tbps of data rates, operating at 3Thz of bandwidth in order to support data-intensive applications such as online gaming, live streaming of high definition videos, the transmission of holographic content, and real-time avatar interactions \cite{pengnoo2020digital,adhikari20226g,Liu_2021,fantacci2021edge}. 6G will also provide ubiquitous coverage and ultra-low latency (less than 1ms) as well as support around 100 km/hour mobility by integrating the space-air-ground-sea networks \cite{chude2022enabling,chakrabarti2021deep,tang2022roadmap}. 

Nevertheless, Metaverse applications such as virtual education and training, precise navigation and localization, immersive gaming, and remote healthcare applications would all be enabled by AI and 6G, making the Metaverse more successful. In particular, state-of-the-art in computer vision can be utilized to provide animated 3D
human models or realistic animated faces, and even for the creation of holograms \cite{aggarwal2021generative,huang2019process,li2021animated}. However, the key challenge in utilizing the 3D content for Metaverse services lies in the scalability of the existing infrastructure. Therefore, the rise of Metaverse makes it crucial to develop the key tools and infrastructure to enable Metaverse developers to build better and more scalable 3D/AR/VR experiences, regardless of platforms or purposes. In principle, researchers need to take a leap from the supervised learning paradigm of AI and resort to diversified learning strategies such as reinforcement learning and self-supervised learning in order to scale in a Metaverse environment. Moreover, the edge computing capabilities of 6G networks can be combined with the AI to provide edge intelligence, hence reducing the network delays and privacy issues for enhanced QoE in Metaverse \cite{loven2019edgeai}. \textit{Motivated by the existing technologies and foreseen challenges, we survey the state-of-the-art in AI and 6G in the context of Metaverse to answer the following question: how to provide a better and sustainable Metaverse experience by leveraging the ongoing progress in AI, 6G, and the nexus of both?} In the following subsections, we review the related surveys and outline the contribution and structure of this survey. 

\definecolor{Gray}{gray}{1}
\definecolor{Gray1}{gray}{0.94}
\definecolor{Gray2}{gray}{0.88}
{\renewcommand{\arraystretch}{1.1}
\begin{table}[t!]
\centering
	\caption{List of Important Acronyms} 
	\scalebox{0.75}{
	\setlength\tabcolsep{10pt}
 
 
  

\begin{tabular}{|ll|}
\hline 
 \rowcolor{Gray2} \textbf{Acronym} & \textbf{Definition}  \\ \hline\hline
 
 ICT & \begin{tabular}[c]{@{}c@{}}Information and Communications Technology \end{tabular} \\
 
 AI & \begin{tabular}[c]{@{}c@{}}Artificial Intelligence \end{tabular} \\
 
 6G & \begin{tabular}[c]{@{}c@{}}Sixth-Generation \end{tabular} \\ 
 
 B5G & \begin{tabular}[c]{@{}c@{}}Beyond 5G \end{tabular} \\
 
 KPI & \begin{tabular}[c]{@{}c@{}}Key Performance Indicator \end{tabular} \\
 
 NIB& \begin{tabular}[c]{@{}c@{}}Network-in-Box \end{tabular} \\ 
 
 SDN & \begin{tabular}[c]{@{}c@{}}Software-defined Networking \end{tabular} \\
 
 NFV & \begin{tabular}[c]{@{}c@{}}Network Functions Virtualization \end{tabular} \\
 
 O-RAN & \begin{tabular}[c]{@{}c@{}}Open Radio Access Networks \end{tabular} \\
 
 3GPP & \begin{tabular}[c]{@{}c@{}}3rd Generation Partnership Project \end{tabular} \\
 
 AI & \begin{tabular}[c]{@{}c@{}}Artificial Intelligence \end{tabular} \\ 
 
 LTE & \begin{tabular}[c]{@{}c@{}}Long Term Evolution \end{tabular} \\ 

 CPS & \begin{tabular}[c]{@{}c@{}}Cyber-Physical Systems \end{tabular} \\ 
 
 DT & \begin{tabular}[c]{@{}c@{}}Digital Twin \end{tabular} \\ 
 
 IIoT & \begin{tabular}[c]{@{}c@{}}Industrial Internet-of-Things \end{tabular} \\
 
 UAVs & \begin{tabular}[c]{@{}c@{}}Unmanned Aerial Vehicles \end{tabular} \\
 
 mMIMO & \begin{tabular}[c]{@{}c@{}}Massive Multiple-Input and Multiple-Output \end{tabular} \\
 
 THz and mmWave & \begin{tabular}[c]{@{}c@{}}Terahertz and Millimeter Wave \end{tabular} \\
 
 RIS & \begin{tabular}[c]{@{}c@{}}Reflecting Intelligent Surfaces \end{tabular} \\
 
 IoT & \begin{tabular}[c]{@{}c@{}}Internet of Things \end{tabular} \\
 
 mLLMTC& \begin{tabular}[c]{@{}c@{}}Massive Low Latency Machine Type Communication\end{tabular} \\ 
 
 UmMTC & \begin{tabular}[c]{@{}c@{}}Ultra-massive Machine Type Communication\end{tabular} \\ 
 
 eRLLC & \begin{tabular}[c]{@{}c@{}}Extremely Reliable and Low Latency Communications \end{tabular} \\ 
 
 eMBB & \begin{tabular}[c]{@{}c@{}}Enhanced Mobile Broadband \end{tabular} \\

 MEC & \begin{tabular}[c]{@{}c@{}}Multi-access Edge Computing \end{tabular} \\

 HCI & \begin{tabular}[c]{@{}c@{}}Hyper-computer Interface \end{tabular} \\ 
 
 HT & \begin{tabular}[c]{@{}c@{}}Holographic Telepresence \end{tabular} \\ 
 
 GPUs & \begin{tabular}[c]{@{}c@{}}Graphical Processing Units \end{tabular} \\ 
 
 HMD & \begin{tabular}[c]{@{}c@{}}Head-mounted Display \end{tabular} \\ 
 
 FoV & \begin{tabular}[c]{@{}c@{}}Field of View \end{tabular} \\ 
 FPS & \begin{tabular}[c]{@{}c@{}}Frames per second\end{tabular} \\ 
 
 SAGSIN & \begin{tabular}[c]{@{}c@{}}Space-air-ground-sea Integrated Network
 
 \end{tabular} \\

 QoS and QoE & \begin{tabular}[c]{@{}c@{}}Quality-of-Services and Quality-of-Experiences \end{tabular} \\ 
 
 GBs and GHz & \begin{tabular}[c]{@{}c@{}} Gigabytes and Gigahertz \end{tabular} \\ 
 HDR & \begin{tabular}[c]{@{}c@{}} High Dynamic Range \end{tabular} \\ 

 AIInfra & \begin{tabular}[c]{@{}c@{}} AI-based Infrastructure \end{tabular} \\ 
 
 AIInt & \begin{tabular}[c]{@{}c@{}} AI-based Interfaces \end{tabular} \\ 
 
 AICont & \begin{tabular}[c]{@{}c@{}} AI-based Smart Contracts \end{tabular} \\ 
 
 AIVWorld & \begin{tabular}[c]{@{}c@{}} AI-based virtual worlds \end{tabular} \\ 

 AIART-E & \begin{tabular}[c]{@{}c@{}} AI-based Art for economic enrichment \end{tabular} \\
 
 SocialAI & \begin{tabular}[c]{@{}c@{}} Improved social network experience with AI \end{tabular} \\ 
 
 PersonalizedAI & \begin{tabular}[c]{@{}c@{}} AI for personalized immersive experience \end{tabular} \\ 
 
 GPT & \begin{tabular}[c]{@{}c@{}} Generative Pre-Trained Transformer \end{tabular} \\ 
 
 GauGAN & \begin{tabular}[c]{@{}c@{}} Paul Gauguin Generative Adversarial Networks \end{tabular} \\ 

 GPT & \begin{tabular}[c]{@{}c@{}}Generative Pre-trained Transformer \end{tabular} \\

 ML & \begin{tabular}[c]{@{}c@{}}Machine Learning \end{tabular} \\

 DL & \begin{tabular}[c]{@{}c@{}}Deep Learning \end{tabular} \\ 
  FL & \begin{tabular}[c]{@{}c@{}}Federated Learning\end{tabular} \\ 
 
  RSI & \begin{tabular}[c]{@{}c@{}} Real Super Intelligence \end{tabular} \\ 
 
 NLP & \begin{tabular}[c]{@{}c@{}} Natural Language Processing \end{tabular} \\ 

FCC & \begin{tabular}[c]{@{}c@{}} Federal
Communications Commission \end{tabular} \\ 

HEMTs & \begin{tabular}[c]{@{}c@{}} High-electron-Mobility Transistors \end{tabular} \\ 

AR & \begin{tabular}[c]{@{}c@{}} Augmented Reality \end{tabular} \\ 
VR & \begin{tabular}[c]{@{}c@{}} Virtual Reality \end{tabular} \\ 
XR & \begin{tabular}[c]{@{}c@{}} Extended Reality \end{tabular} \\

\hline

\end{tabular}}
\label{table:Acronyms}
\vspace{-10pt}
\end{table}
}

\definecolor{Gray}{gray}{1}
\definecolor{Gray1}{gray}{0.94}
\definecolor{Gray2}{gray}{0.88}
\bgroup
{\renewcommand{\arraystretch}{0.9}
\begin{table*}[!ht]
\centering
\caption{Summary and categorization of the reviewed studies concerning Metaverse}
\scalebox{0.75}{
\begin{tabular}{|c|c|c|c|c|c|c|c|c|c|c|c|c|} 
\hline
\rowcolor{Gray2}\textbf{Authors \& Year} & {\rotatebox[origin=c]{90}{\textbf{Reference Numbers}}}   & {\rotatebox[origin=c]{90}{\textbf{AR, VR, XR, Spatial Computing}}} & {\rotatebox[origin=c]{90}{\textbf{Layered Metaverse Architecture}}} & {\rotatebox[origin=c]{90}{\textbf{Computer Vision and Learning Paradigms}}} & {\rotatebox[origin=c]{90}{\textbf{5G/6G, MEC, mmWave, THz}}} & {\rotatebox[origin=c]{90}{\textbf{AI for 6G Networks}}} & {\rotatebox[origin=c]{90}{\textbf{6G networks for AI}}} & {\rotatebox[origin=c]{90}{\textbf{Sustainability}}} & {\rotatebox[origin=c]{90}{\textbf{Applications and Usecases}}} & {\rotatebox[origin=c]{90}{\textbf{Frameworks, Projects \& Demos}}} & {\rotatebox[origin=c]{90}{\textbf{Challenges, Research Directions, \& Lessons Learned}}} 
& \begin{tabular}[c]{@{}c@{}}\textbf{Remarks} \\ \textbf{(Relevance with Metaverse)} \end{tabular}
\\ 
\hline\hline
\begin{tabular}[c]{@{}c@{}}Yang~\textit{et al}., \\ 2022 \end{tabular} & \cite{yang2022fusing}         & \cellcolor{red!20} L  & \cellcolor{red!35} \xmark & \cellcolor{green!20} \cmark & \cellcolor{red!35} \xmark & \cellcolor{red!35} \xmark & \cellcolor{red!35} \xmark  & \cellcolor{blue!20} M & \cellcolor{red!35} \xmark & \cellcolor{red!35} \xmark & \cellcolor{blue!20} M & Integration of Blockchain and AI for Metaverse    \\ \hline

\rowcolor{Gray1} \begin{tabular}[c]{@{}c@{}}Wang~\textit{et al}., \\ 2022 \end{tabular} & \cite{wang2022survey}   &\cellcolor{blue!20} M  &\cellcolor{red!35} \xmark  & \cellcolor{red!20} L    &\cellcolor{green!20} \cmark & \cellcolor{red!35} \xmark  & \cellcolor{red!35} \xmark  & \cellcolor{red!20} L  & \cellcolor{green!20} \cmark & \cellcolor{red!20} L & \cellcolor{green!20} \cmark & Security and Privacy for Metaverse   \\ \hline

\begin{tabular}[c]{@{}c@{}}Huynh~\textit{et al}., \\ 2022 \end{tabular} & \cite{huynh2022artificial}   &\cellcolor{green!20} \cmark     &\cellcolor{red!35} \xmark   &\cellcolor{green!20} \cmark   &  \cellcolor{red!20} L      &\cellcolor{green!20} \cmark  & \cellcolor{red!20} L  & \cellcolor{red!20} L  & \cellcolor{green!20} \cmark & \cellcolor{blue!20} M & \cellcolor{blue!20} M & Artificial Intelligence for Metaverse   \\ \hline

\rowcolor{Gray1} \begin{tabular}[c]{@{}c@{}}Lim~\textit{et al}., \\ 2022 \end{tabular} & \cite{lim2022realizing}   &\cellcolor{green!20} \cmark &\cellcolor{green!20} \cmark   & \cellcolor{blue!20} M    &\cellcolor{green!20} \cmark & \cellcolor{blue!20} M  & \cellcolor{green!20} \cmark  & \cellcolor{red!20} L  & \cellcolor{red!20} L & \cellcolor{red!20} L & \cellcolor{green!20} \cmark  & Edge Intelligence for Metaverse   \\ \hline

 \begin{tabular}[c]{@{}c@{}}Khan~\textit{et al}., \\ 2022 \end{tabular} & \cite{khan2022Metaverse}  &\cellcolor{red!20} L  &\cellcolor{red!35} \xmark      &\cellcolor{red!20} L    &\cellcolor{green!20} \cmark      &\cellcolor{red!35} \xmark  & \cellcolor{green!20} \cmark & \cellcolor{red!20} L  & \cellcolor{blue!20} M & \cellcolor{red!20} L & \cellcolor{red!20} L    & Wireless Architectures for Metaverse  \\ \hline

\rowcolor{Gray1} \begin{tabular}[c]{@{}c@{}}Chang~\textit{et al}., \\ 2022 \end{tabular} & \cite{chang20226g}   &\cellcolor{red!20} L   &\cellcolor{red!20} L      &   \cellcolor{red!20} L    &\cellcolor{green!20} \cmark     &  \cellcolor{red!20} L & \cellcolor{green!20} \cmark  & \cellcolor{red!35} \xmark  & \cellcolor{blue!20} M & \cellcolor{red!20} L & \cellcolor{blue!20} M  & 6G-powered Edge AI for Metaverse   \\ \hline

\begin{tabular}[c]{@{}c@{}}Jagatheesaperumal~\textit{et al}., \\ 2022 \end{tabular} & \cite{jagatheesaperumal2022advancing}       &\cellcolor{green!20} \cmark    &\cellcolor{red!35} \xmark &\cellcolor{red!35} \xmark   &  \cellcolor{red!20} L   &  \cellcolor{red!35} \xmark    & \cellcolor{red!20} L  & \cellcolor{red!20} L  & \cellcolor{green!20} \cmark & \cellcolor{red!20} L & \cellcolor{green!20} \cmark   & XR and IoE for Education in Metaverse \\ \hline


\rowcolor{Gray1} \begin{tabular}[c]{@{}c@{}}Dhelim~\textit{et al}., \\ 2022 \end{tabular} & \cite{dhelim2022edge}     &\cellcolor{red!20} L   &\cellcolor{red!20} L      &   \cellcolor{red!20} L        &  \cellcolor{green!20} \cmark       &\cellcolor{red!20} L & \cellcolor{green!20} \cmark  & \cellcolor{red!35} \xmark & \cellcolor{blue!20} M & \cellcolor{red!35} \xmark & \cellcolor{red!20} L & Mobile Edge Computing for Metaverse   \\ \hline

 \begin{tabular}[c]{@{}c@{}}Park~\textit{et al}., \\ 2022 \end{tabular} & \cite{park2022Metaverse}      &\cellcolor{green!20} \cmark          &\cellcolor{red!35} \xmark     &  \cellcolor{green!20} \cmark    &  \cellcolor{red!20} L    &\cellcolor{green!20} \cmark  & \cellcolor{red!20} L & \cellcolor{red!20} L & \cellcolor{green!20} \cmark & \cellcolor{blue!20} M & \cellcolor{blue!20} M & User interactions, Implementations, and Applications in Metaverse   \\\hline

\rowcolor{Gray1} \begin{tabular}[c]{@{}c@{}}Tang~\textit{et al}., \\ 2022 \end{tabular} & \cite{tang2022roadmap}      &\cellcolor{red!20} L &\cellcolor{red!35} \xmark       &  \cellcolor{red!20} L        &\cellcolor{blue!20} M       &  \cellcolor{red!20} L  & \cellcolor{green!20} \cmark  & \cellcolor{red!35} \xmark & \cellcolor{red!35} \xmark & \cellcolor{red!35} \xmark & \cellcolor{green!20} \cmark  & URLLC, Digital Twins, and SAGSIN for Metaverse    \\ \hline



\begin{tabular}[c]{@{}c@{}}Gadekallu~\textit{et al}., \\ 2022 \end{tabular} & \cite{gadekallu2022blockchain}      &\cellcolor{green!20} \cmark   &\cellcolor{red!35} \xmark   &  \cellcolor{red!20} L    &\cellcolor{red!20} L    &  \cellcolor{red!35} \xmark  & \cellcolor{red!35} \xmark & \cellcolor{red!20} L & \cellcolor{red!20} L & \cellcolor{green!20} \cmark & \cellcolor{green!20} \cmark   & Blockchain Technology for Metaverse     \\ \hline

\rowcolor{Gray1} \begin{tabular}[c]{@{}c@{}}Mozumder~\textit{et al}., \\ 2022 \end{tabular} & \cite{mozumder2022overview}      &\cellcolor{red!20} L     &\cellcolor{red!35} \xmark     &  \cellcolor{red!20} L      &   \cellcolor{red!20} L     &\cellcolor{blue!20} M    & \cellcolor{blue!20} M  & \cellcolor{red!35} \xmark & \cellcolor{red!35} \xmark & \cellcolor{red!35} \xmark & \cellcolor{red!35} \xmark  & Technology Roadmap for Healthcare in Metaverse  \\ \hline

 \begin{tabular}[c]{@{}c@{}}Ning~\textit{et al}., \\ 2021 \end{tabular} & \cite{ning2021survey}     & \cellcolor{green!20} \cmark   & \cellcolor{red!35} \xmark  & \cellcolor{blue!20} M  & \cellcolor{red!20} L   & \cellcolor{red!35} \xmark  & \cellcolor{red!35} \xmark & \cellcolor{blue!20} M & \cellcolor{green!20} \cmark & \cellcolor{green!20} \cmark & \cellcolor{green!20} \cmark & Social Value and Technology convergence in Metaverse   \\ \hline

\rowcolor{Gray1} \begin{tabular}[c]{@{}c@{}}Lee~\textit{et al}., \\ 2021 \end{tabular} & \cite{lee2021all}     & \cellcolor{green!20} \cmark &\cellcolor{blue!20} M   &  \cellcolor{green!20} \cmark   &\cellcolor{blue!20} M & \cellcolor{red!20} L  & \cellcolor{red!20} L & \cellcolor{blue!20} M & \cellcolor{green!20} \cmark & \cellcolor{red!20} L & \cellcolor{green!20} \cmark  & Technological Ecosystem for Metaverse   \\ \hline

\begin{tabular}[c]{@{}c@{}}Lee~\textit{et al}., \\ 2021 \end{tabular} & \cite{lee2021creators}  &  \cellcolor{green!20} \cmark  &\cellcolor{red!20} L  &\cellcolor{green!20} \cmark   &\cellcolor{red!35} \xmark   &   \cellcolor{red!35} \xmark   & \cellcolor{red!35} \xmark   & \cellcolor{red!20} L  & \cellcolor{blue!20} M  & \cellcolor{green!20} \cmark & \cellcolor{blue!20} M &  Digital Artworks for Metaverse   \\ \hline

\rowcolor{Gray1} \textbf{Our Article} & \textbf{---}        &\cellcolor{green!20} \cmark  &\cellcolor{green!20} \cmark &\cellcolor{green!20} \cmark &\cellcolor{green!20} \cmark &\cellcolor{green!20} \cmark  & \cellcolor{green!20} \cmark & \cellcolor{green!20} \cmark  & \cellcolor{green!20} \cmark & \cellcolor{green!20} \cmark & \cellcolor{green!20} \cmark  &  \begin{tabular}[c]{@{}c@{}}This survey investigates the underlying technologies of AI and 6G, such as advancements in computer vision,\\learning paradigms, and wireless communication technologies in the context of Metaverse. \\The joint role of AI and 6G in obtaining ubiquitous intelligence, tactile feedbacks,\\ and self-optimising capabilities for Metaverse is explored. \\The sustainability aspect of Metaverse followed by the applications, usecases, and on-going projects is given \\ Lastly, numerous open issues, future directions, and lessons learned are enlightened for \\ potential researchers and developers of Metaverse
applications and services. \end{tabular} 
 \\ \hline
\end{tabular}}

\vspace*{3pt}

\fbox{\begin{tabular}{cccccccc}
\cellcolor{green!20}\cmark & High Coverage &  \cellcolor{blue!20} M & Medium Coverage & \cellcolor{red!20} L & Low Coverage & \cellcolor{red!35} \xmark & Absent/Unavailable
\end{tabular}}

\label{table:reviewedstudiessummary}
\end{table*}
}

\subsection{Related surveys}
In this subsection, we review the state-of-the-art surveys conducted on several technologies in the context of Metaverse. Recently, there has been many surveys and research studies conducted to investigate the potential technology, its implementation, applications, and future research directions \cite{yang2022fusing,ning2021survey,lee2021all,wang2022survey,lee2021creators,huynh2022artificial,lim2022realizing,khan2022Metaverse,chang20226g,jagatheesaperumal2022advancing,dhelim2022edge,park2022Metaverse,tang2022roadmap,gadekallu2022blockchain,mozumder2022overview}. For example, Yang et al., \cite{yang2022fusing} investigate the integrated role of AI and Blockchain for the Metaverse environments. In particular, authors in \cite{yang2022fusing} propose that Metaverse is a 3D virtual reality platform that covers all aspects of social and economic activities and allows everyone to participate in these activities in a safe and free environment that transcends the limitations of our real world by making use of AI and blockchain technology. Similarly, \cite{ning2021survey} surveys the security and privacy concerns for Metaverse users and proposes potential directions to provide secure services to Metaverse users. On the other hand, researchers in \cite{lee2021all,wang2022survey,huynh2022artificial} investigate the role of AI in the Metaverse. In particular, \cite{lee2021all} surveys the existing state-of-the-art in AI, computer vision, and deep learning in the context of Metaverse, while \cite{wang2022survey} and \cite{huynh2022artificial} explore the role of edge intelligence and 6G-powered edge AI for Metaverse, respectively. In contrast, authors in \cite{khan2022Metaverse,jagatheesaperumal2022advancing,dhelim2022edge,tang2022roadmap} scrutinize the wireless communication architectures and technologies in support of Metaverse. For instance, \cite{tang2022roadmap} reviews the advancements in ultra reliable and low latency
communication (uRLLC), digital twins, and SAGSIN for providing ubiquitous intelligence in Metaverse, while \cite{dhelim2022edge} and \cite{jagatheesaperumal2022advancing} provides the overview of Mobile Edge Computing (MEC) and Internet of Everything (IoE) for Metaverse. Different from above studies, authors in \cite{park2022Metaverse,mozumder2022overview,ning2021survey,lee2021creators,lee2021all} provide the overview of societal, economical, technological, and digital value in Metaverse applications. In Table \ref{table:reviewedstudiessummary}, we compare and contrast the above-discussed state-of-the-art with this article in terms of different aspects of Metaverse.

\subsection{Contribution and Structure of this survey}
It is clear from the above discussion that some of the existing surveys have focused on very narrow perspectives of AI, 6G, and relevant technologies, while others have investigated the role of the Metaverse in terms of societal, economic, and digital value. As opposed to them, in this review, we provide the following key contributions: 
\begin{itemize}
\item We describe the underlying components of Metaverse, i.e., VR, MR, AR, and Spatial computing. The fundamentals of the aforementioned technologies are presented to provide readers with an understanding of technical aspects and state-of-the-art, leading to a fully immersive Metaverse. 
\item We present an extensive review of the state-of-the-art in AI and examine the role of AI in realizing the Metaverse with an aim to extract communication requirements for 6G in Metaverse. Essentially, we define the role of AI in the layered architecture of Metaverse, followed by the state-of-the-art in computer vision applications, learning paradigms and Edge AI for Metaverse. 
\item We explain the role of B5G/6G in realizing Metaverse by firstly answering the following key questions: i) \textit{ is B5G/6G need of an hour?} and ii) \textit{what services B5G/6G can bring to Metaverse?} Next, we review the state-of-the-art for immersive experiences and holographic telepresence over wireless by exclusively considering the role of 5G-NR, URLLC, mmWave, MEC, THzComms, and their interplay.
\item We also consider the integrated role of AI and 6G towards realization of the Metaverse. In principle, we examine the potential of \textit{AI for 6G networks} and \textit{6G networks for AI} in support of Metaverse. Next, we investigate the sustainability of Metaverse in the context of the aforementioned technologies, followed by the potential applications and usecases and ongoing projects. Lastly, we present the challenges and future research directions along with the lessons learned from this survey.
\end{itemize}
The rest of the article is organized as follows. Section II includes the background and technical aspects of all relevant technologies, including AI, 6G, VR, MR, AR, and Spatial computing. Section III presents the role of AI in the Metaverse, while Section IV covers the role of B5G/6G in the Metaverse. Section V highlights the integrated role of AI and 6G in the context of Metaverse, followed by the sustainability perspective of Metaverse in Section VI. Applications and Usecases are explored in Section VII, followed by the description of projects in Section VIII. Challenges and future research directions are entailed in Section IX, while Section X presents the lessons learned from this survey. Finally, Section XI concludes the survey. In Figure \ref{Fig1}, we present our paper's complete pictorial structure.


\section{AI and 6G for Metaverse: Background and Technical aspects} 

This section of the article will provide a detailed account of the background on the metaverse from the point of view of technical analysis. The term  “Metaverse" is a combination of the prefix  “meta" (meaning   “beyond") and the suffix  “verse" (short for  “universe"). As a result, it refers to a universe beyond the physical world. This “universe beyond" alludes to a computer-generated environment instead of metaphysical or mystical notions of domains beyond the physical reality \cite{Coinbase:2021}. Metaverses are fully immersive three-dimensional digital environments, as opposed to cyberspace, which refers to all online spaces. We provide a critical evaluation of AI and the Metaverse in this section and some essential prerequisites for understanding the Metaverse 6G prospects \cite{huang2022analysis}. This comprehensive overview will cover the fundamental algorithms, terminology, and current and future product applications. We take a brief introduction to AI, which constructed a sophisticated black-box for high-level tasks such as detection and classification. The Metaverse role is then conducted to investigate what the Metaverse requires of AI key stakeholders for more trustworthy communications \cite{David:2021}.

\subsection{The background of AI for Metaverse }
This secction explain the background of metaverse. The metaverse  incorporates numerous technologies such as 3D animation, VR, AR, spatial computing, blockchain, and many more hot technologies\cite{maccallum2019teacher, sparkes2021Metaverse}. It is the next technological future from a socioeconomic standpoint. Many enterprises have already devoted significant resources to Metaverse initiatives to give new digital services to the globe and be the first to deliver metaverse landscapes. Why do we say metaverse is the hottest technological and socioeconomic topic right now? Take the example of Facebook, which changed its name to Meta from Facebook, demonstrating that the metaverse will be the next big mainstream technology for industries.
The future of Internet in metaverse will be a large-scaled, resilient, dynamic, and platform for socializing, trading, transferring data, and entertaining people in real-time. Hitherto, a lot of effort was made on Web1 and Web2 \cite{kshetri2022web}. Web1 was guided on browsing plain online pages, whereas Web2  explained how to connect to social platforms within controlled ecosystems. In the metaverse, NFTs will take the place of digital ownership in an open, decentralized environment \cite{valaskova2022virtual}. This will enable people to link digitally with the rest of the world and generate a healthy economy in which people can execute any work that can be performed in the physical world. Much has been said about the Internet and its future, but it is vital to distinguish between the metaverse and Web3. Web3 is about applying advanced digital services, but instead of being controlled by large technology corporations as in Web2, these services will be created and governed by the community, returning to the ethos of Web1, when the Internet's value was generated by users at the network's edge, predominantly in a forward mode. Virtual worlds, massive scalability, persistence, synchronicity, a functional economy, openness and decentralization, interoperability, and so on should all be included in a Metaverse platform \cite{David:2021}. The metaverse begins to make artificial intelligence (AI) more ubiquitous in reality, which will power numerous metaverse technology layers such as spatial computing, artist scaffolding, and new emerging types of storytelling.
We have seen that AI is advancing at a comparable level, such as transformers; transformers are neural networks that allow machines to interact with natural language, and the graph below shows the exponential expansion of deep learning transformers \cite{Jon1:2021}\cite{Jon2:2021}. In Figure \ref{Fig2}, we show how Growth of Deep Leaning Transformers from 2018 to 2021.

\begin{figure*}[!ht]
\includegraphics[width= 1.0\textwidth]{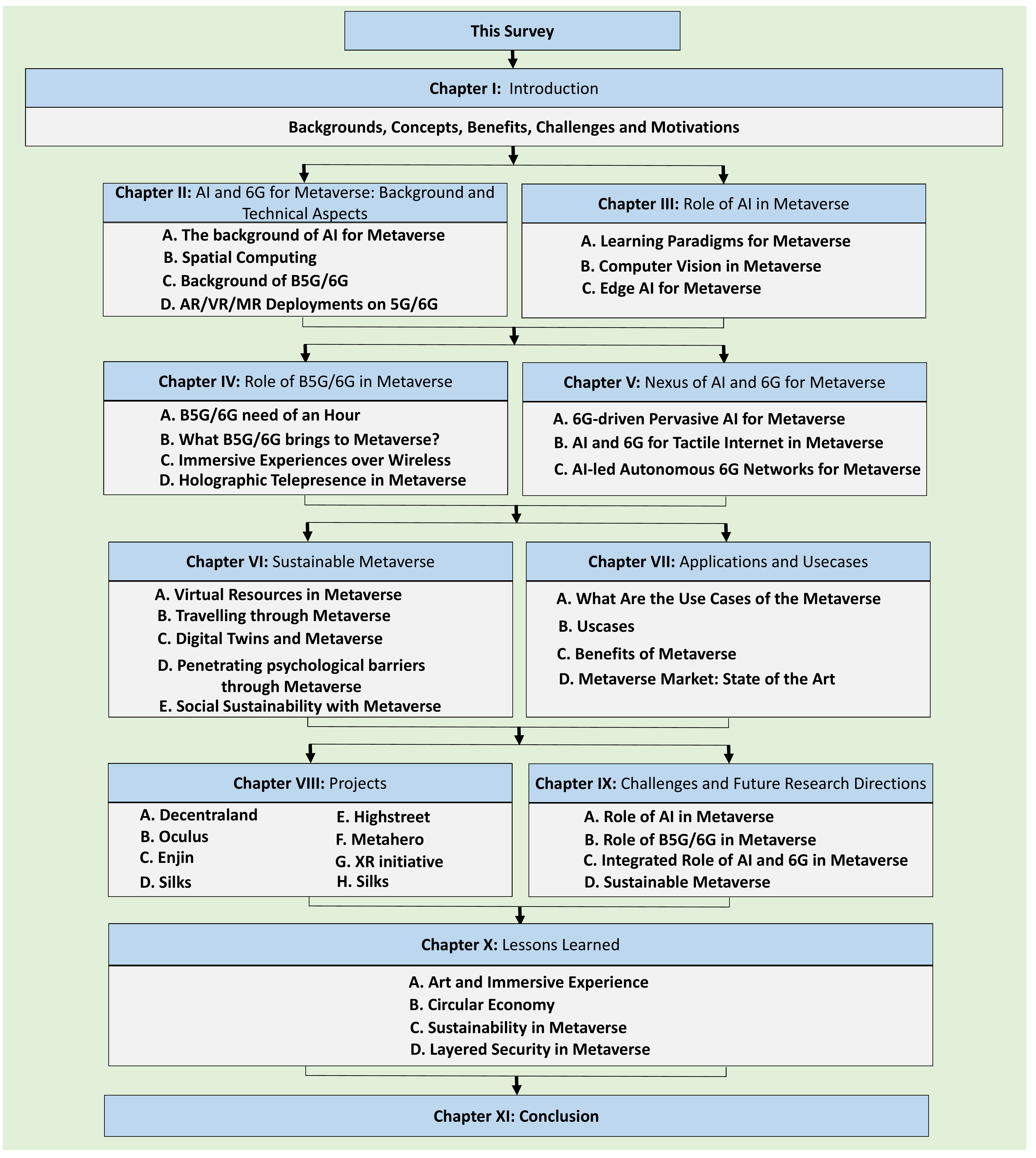}
\caption{Structure of this survey}
\centering\label{Fig1}
\end{figure*}

\begin{figure}
\includegraphics[width=0.5\textwidth]{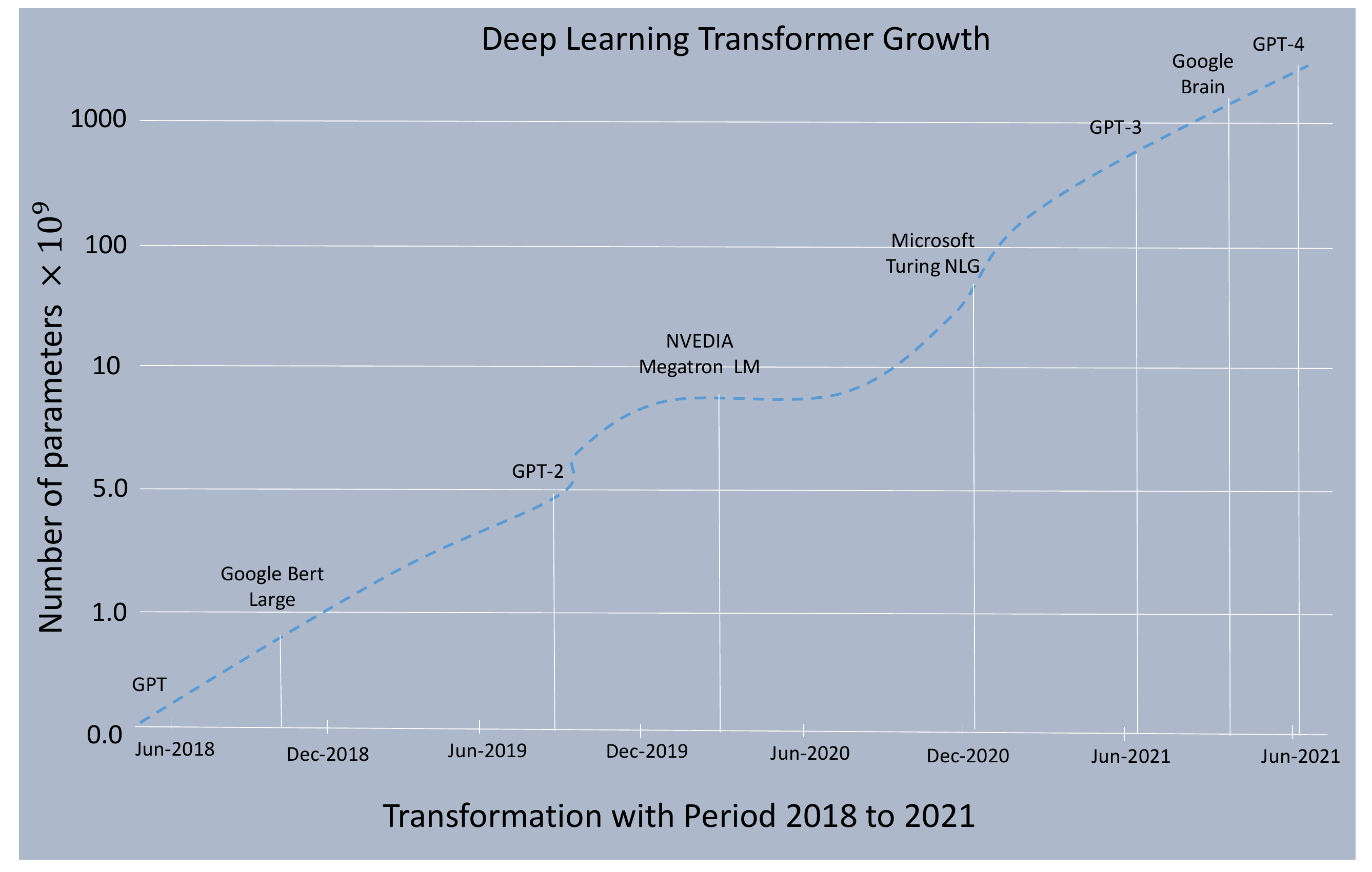}
\caption{Growth of Deep Learning Transformers from 2018 to 2021}
\centering\label{Fig2}
\end{figure}

Consider the following transformers: the Generative Pre-trained Transformer (GPT) dealt with 110 million parameters; the newest Google Brain transformer dealt with over 1 trillion parameters. GPT-4 will most likely have far more. This suggests that the volume of DNN networks will continue to grow in the future. With the advancement of neural networks, AI has already achieved a lot of success in voice recognition in Alexa, Tesla's autonomous driving system, Google image recognition, and a few algorithms that elicit reactions from social media and deep fakes; however, all of these applications are just basic in comparison to AI's future applications \cite{Jon1:2021}\cite{Jon3:2021}. The Metaverse is not in the future; it is a transformation from being to becoming.

The Metaverse is where the physical and digital worlds intersect; it contains the shared virtual environment, AR glasses, headgear devices, and Oculus VR headsets. These can be accessed over the Internet via smartphones and wristband technology. People will play, learn, create, shop, and communicate with friends in a virtual, online world. Today, the Metaverse is emerging as metaphysics, or, as we might say, the transformation of the universe. In the past two years, we have seen that with the covid epidemic, digital technologies shift their waves quickly, and it has been in the past as well. The 1980s saw the rise of computers, the 1990s saw the rise of the Internet, the 2000s saw the rise of mobile Internet, the 2010s saw the rise of AI/ML/DL/Data alaytics \cite{aggarwal2022has}, and the 2020s saw the rise of true AI/Trans-AI/Meta-AI. The Metaverse, transverse, or omniverse could be the next wave. The waves are now merging, such as the Internet on PCs, the Internet on mobile devices, and AI/ML on the mobile Internet. The Metaverse will demand the Internet, mobile devices, and REAL AI.
A Metaverse contains simulated AR, VR, MR \cite{lik2017interaction,lee2019hibey,huang20193d,d2020markerless, bajireanu2019mobile,nuzzi2020hands, wang2020avatarmeeting}, a 3D World Wide Web, and other forms of digital twins, including digital humans as intelligent avatars. Metaverse will function as a real-time digital environment containing all feasible entities, occurrences, activities, and interactions. Currently, trillions of dollars are being invested in Metaverse projects like AI-driven digital reality to offer a digital infrastructure that can accommodate all of its social media networks, e-commerce, and big tech oligopoly \cite{ponnusamy2022ai}. We risk creating a Matrix-like Metaverse of digitized humans and superintelligent agents as a virtual world where young people can transcend the meaningless reality of the real world in which they live. 
The Metaverse is a virtual reality that integrates all essential aspects of cyberspace or the world wide web, such as cloud and edge computing, social media, online gaming, augmented reality (AR), virtual reality (VR) \cite{lik2017interaction,lee2019hibey}, cryptocurrencies \cite{phillip2018new}, and AI/ML/DL platforms \cite{aggarwal2022has} and applications, to allow users to interact virtually. Everything we use anyway, such as video conferencing, games, email, mixed reality, e-commerce, social media, and Netflix/YT live-streaming, are all part of the Metaverse. For many MMOG gamers, the Metaverse, defined as the Internet filled with mixed, virtual, and augmented worlds, is more real than any physical reality. It is as real as its technological solutions, which include social media, virtual virtual world games, extended reality, simulation and modeling, human-computer interaction, digital twin, machine learning and reasoning, data analytics, computer vision, edge and cloud computing, and mobile networks, all of which are powered by true AI Metaverse technology.  Using machine learning and deep learning, we have demonstrated the detailed involvement of the metaverse in 6G technologies in Figure \ref{Figg}.

\begin{figure}
\includegraphics[width=0.5\textwidth]{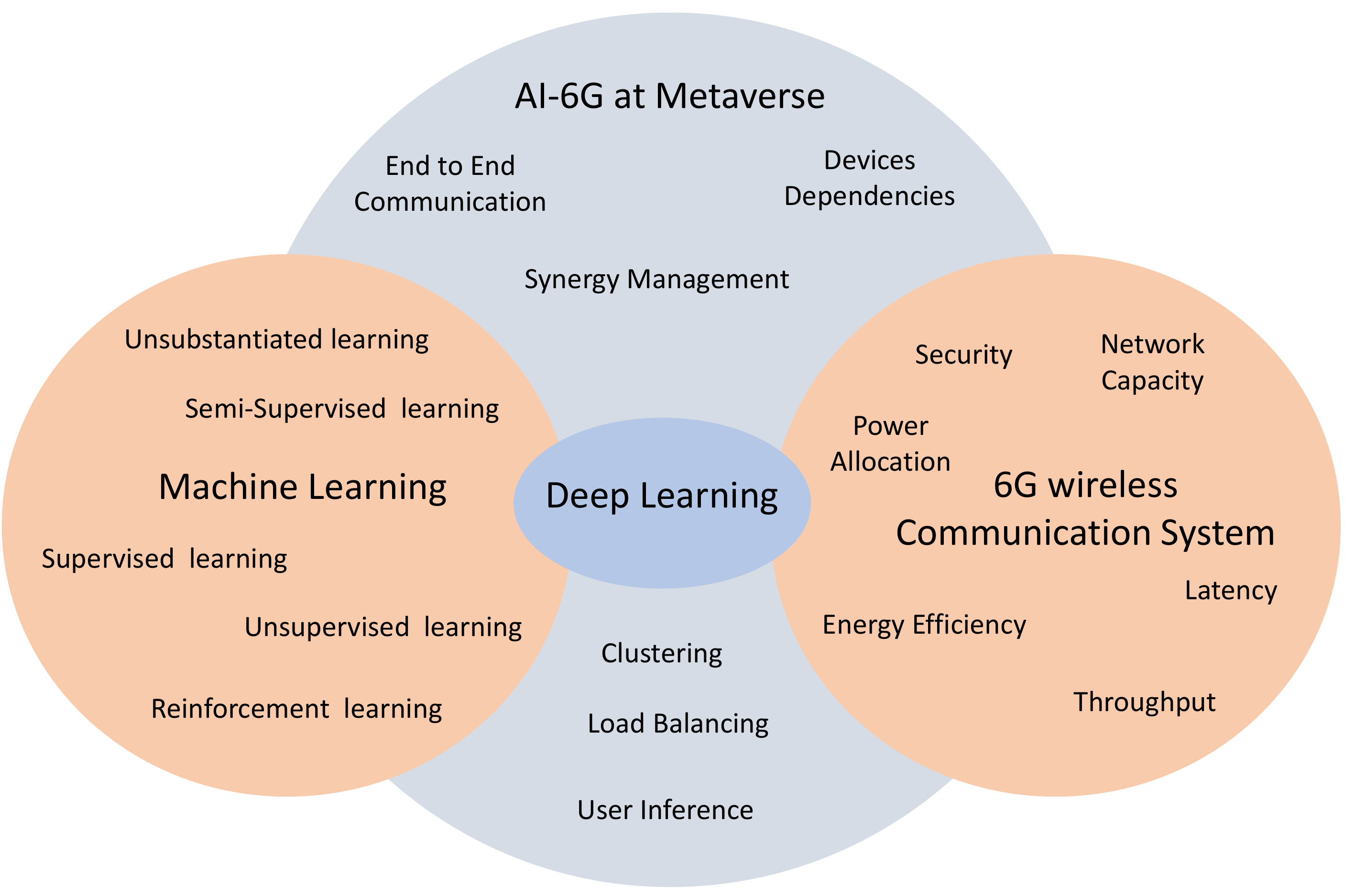}
\caption{Application and infrastructure-level machine learning and deep learning techniques for 6G.}
\centering\label{Figg}
\end{figure}

\textbf{Building the Metaverse as a Global AI Platform:} Based on the data presented above, it is evident that the Metaverse will prevail in the coming years due to its immense potential for success as a global AI computing system. Breakthroughs in the Metaverse are expected in the future, including several technological developments, protocols, companies, and scientific points of view. To advance in the Metaverse context, one must establish Metaverse infrastructures and then create a set of standard rules and protocols to load valuable intelligence information, evolve, and continue to perpetuate. With AI, DL, and ML, the Metaverse can achieve Real Super Intelligence (RSI), demonstrating that it can function as advanced man-machine intelligence. The Metaverse will be enabled, filled, and sustained by the most sophisticated man-machine intellect, the RSI, rather than the narrow and weak AI and ML \cite{zhu2022metaaid}. It will power all seven Metaverse technological layers, including spatial computing, creator scaffolding, and new and complex types of narrative. We envision the role of DL and ML in the Metaverse in the broader context.

\subsubsection{Deep learning in Metaverse}
Deep learning algorithms (DL) have gained significant popularity due to their accuracy and efficiency in a variety of computer vision applications, such as extracting 2D human pose information from RGB camera data \cite{dang2019deep, cao2017realtime,fang2017rmpe} or 3D human pose information from RGB-D sensor data \cite{mehta2017vnect,moeslund2001survey,hu2020fingertrak}. Among SoTA 2D pose tracking approaches in the Metaverse, OpenPose \cite{cao2017realtime} has been widely used by scientists to track user postures in various virtual contexts, such as VR \cite{huang20193d,d2020markerless}, AR \cite{bajireanu2019mobile,nuzzi2020hands, wang2020avatarmeeting}, and Metaverse \cite{shin2021non}. Furthermore, 3D pose tracking, FingerTrack \cite{hu2020fingertrak}, has a great potential for XR applications and the Metaverse in tracking and the hand position estimation approach. As we have seen, multi-person tracking is more complex than single-person tracking; yet, the tracking algorithm must count the number of users and their positions and classify them \cite{dang2019deep}. The Metaverse environment mandates using both single-person and multi-person body posture monitoring techniques. The algorithms designed must be resilient and efficient to ensure strong connections between the Metaverse, the physical world, and people.


\subsection{Spatial Computing}
Spatial computing is the digitalization of machine, human, and object activities and the locations in which they occur to facilitate and improve actions and interactions. Spatial computing is the seamless integration of three technologies: augmented reality (AR), virtual reality (VR), and mixed reality (MR) in a three-dimensional world \cite{Spatial1}\cite{Spatial2}. This technology can digitally revolutionize how industrial organizations optimize operations for front-line workers in factories, work areas, and logistics. Spatial computing technology is analogous to the commonwealth of the physical and digital worlds. This means that almost all people no longer interact with computers in the fashion that an impartial observer should; conversely, they choose to encounter what it is really like to be in the digital domain by associating with elements that only reside in it.
In contrast to traditional computing, which is predominantly two-dimensional, spatial computing allows users to think out of the box and communicate with gadgets from above the screen. The innumerable devices are already on the marketplace for spatial computing. Here are a few examples: VR Headsets, AR Glasses, and Hybrid Gear are all examples of cutting-edge technology \cite{shekhar2015spatial}. Whereas VR headsets enable visitors to explore being a part of a virtual universe, AR Glasses allow individuals to go greater depth into the digital space, and Hybrid Gear combines VR, AR, and MR technology. Since the user may fully integrate their senses, this program enables them to enjoy a cinematic experience. In the past, gamers adopted VR headsets to interact with video game items and avatars. Their application has recently expanded to encompass additional applications such as training and modeling. AR Glasses are gadgets that project data and visuals and are particularly beneficial in industrial workplaces. Google Glass and Microsoft HoloLens are two special AR glasses on the market right now. FellReal's multi-sensory mask is an excellent example of such technology. While such merchandise is still in the works, major tech corporations like Samsung, Google, Apple, and Microsoft invest in startups to create cutting-edge hybrid tools and equipment. By showing how the physical world will be transformed into the digital world via the rise of Metaverse, we demonstrate in Figure \ref{Fig3} how technologies such as enabling technologies will play a significant role. The ultimate advantage of spatial computing is the ability to interact with digital objects. It provides extra privileges that we could only dream of aeons ago but are now a  “spatial" reality \cite{zhang2016effect}.
\begin{figure*}[ht]
\includegraphics[width= 1.0\textwidth]{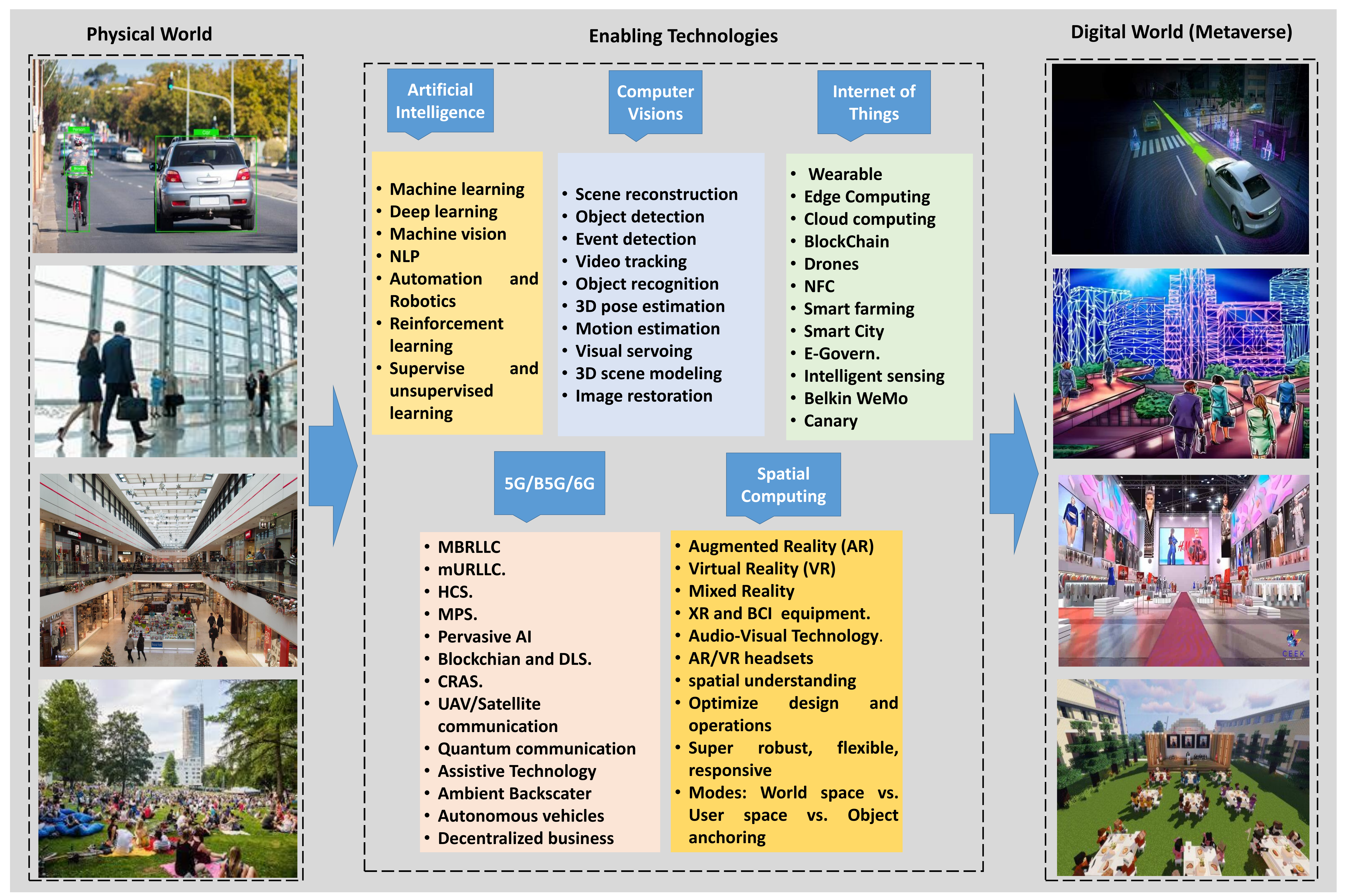}
\caption{Metaverse refers to tranformation of physical reality into the digital realm, so it could be a mix of physical reality and digital realm such as AR, VR, and MR.}
\centering\label{Fig3}
\end{figure*}

\begin{figure*}[ht]
\includegraphics[width= 1.0\textwidth]{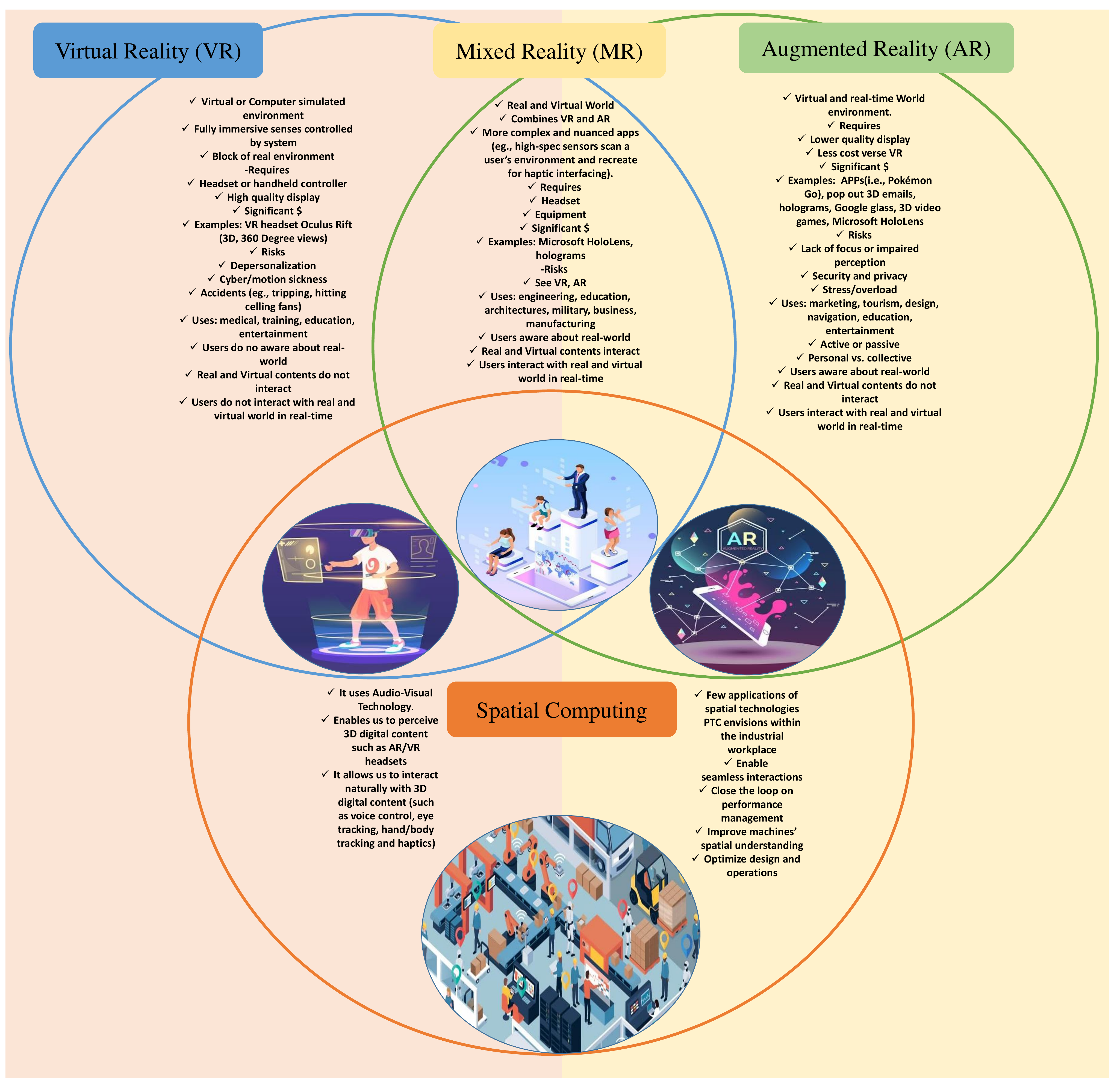}
\caption{Virtual Reality (VR), Augmented Reality (AR), and Mixed Reality (MR), explored in depth.}
\centering\label{Fig4}
\end{figure*}

\subsubsection{Virtual Reality (VR)}
Virtual reality is defined as computer technologies to construct space that can be explored in $360^\circ$ (VR). Unlike traditional technologies, virtual reality (VR) plunges the user in a virtual world to give an interactive experience \cite{Joe:2020}. It has been revealed that virtual reality has the most conspicuous characteristics of completely synthetic vistas. Commercial VR removes real-world vision and provides video to each eye, allowing for depth of view \cite{kelly2021virtual}. It is a virtual environment in which the user is fully immersed and is interacting with real items via interactive techniques via the Internet. This technology is augmented with head and body tracking to link the virtual environment to the user is viewing. Furthermore, VR is described as the “farthest end from reality in the Reality-Virtuality Horizon" \cite{milgram1995augmented}. In a VR environment, users can produce their own content, such as VR painting, to devote their complete attention to the virtual worlds, which are distinct from the real environments \cite{speicher2019mixed}. Interacting with virtual entities in a virtual space can assist in discovering user accessibility, such as changing the scene's shape and adding new tasks.
The collaboration between multiple users in virtual space can therefore occur in real-time, which aligns with well-defined requirements for virtual environments, including a sense of location, time, and availability, gestures, texts, and voice communication, and information sharing \cite{zyda1999networked}. Furthermore, Virtual Reality expands democratization and permits more humans to engage with digitization. For example, Google's Tilt Brush5 enables customers to create unique artworks using VR goggles. The link between new and classic arts technologies is not unidirectional; instead, they can live harmoniously. When contemplating the Metaverse, users should locate themselves in their own shared spaces while interacting with the physical analogue, such as AR and MR. The goal of the Metaverse is to create multiple shared spaces in which to undertake concurrent activities among existing objects, avatars, and their relationships, for example, Object-to-object, object-to-avatar, and avatar-to-avatar interactions. To represent the events of the virtual spaces, all users who participate in the virtual environment must synchronize \cite{liu2012survey}. However, synchronizing dynamic events at scale is a large problem because unlimited users can simultaneously together act on virtual objects and interact with one other without reasonable latency, which might be detrimental to users. VR in the Metaverse seems to be a more advancing technology that may be deployed for various purposes, such as classrooms, to help teach a subject or topic by allowing students to 'experience' the knowledge.

\subsubsection{Augmented Reality (AR)}
Digital visual elements, music, or other sensory stimuli are used in augmented reality (AR) to enhance the physical world. Commercial apps and mobile computing companies are embracing it. Augmented reality (AR) goes beyond virtual worlds to improve our physical surroundings through different experiences. Audio, images, taste, and touch are all possible perceptual input channels for computer-generated virtual elements \cite{narumi2011augmented,schmalstieg2016augmented, kruijff20043d}. Digital toppings are projected on top of our physical environments using the first generation of AR system frameworks. Users' participation with digital entities has been studied extensively in AR so far. It is worth noting that digital entities are superimposed on top of the user's physical surroundings, allowing users to perform many actions at once, similar to VR. Interactions with such digital entities in AR are difficult and intended to connect people between the physical world and the Metaverse \cite{kruijff20043d}. The majority of science fiction films use urban and peri-urban concepts to present intuitive AR user interfaces \cite{lik2017interaction}. Like, Voodoo Dolls offered a freehand technique in which the user can select and work with virtual content using squeeze motions. Later, HOMER proposed a method \cite{pierce2002comparing} that provides ray-casting from the user's virtual hand, displays item selections, and is finally deceived. 
Furthermore, AR will play an important part in our daily lives, such as gaming, annotating directions in new regions, and locating objects based on user surroundings \cite{lee2021towards}. As a result, we can anticipate that the AR in Metaverse will integrate urban environments, and digital entities will appear in visible and tangible ways on top of various physical things in metropolitan places. We can conclude that AR will facilitate communication between the physical and virtual worlds in the Metaverse. However, mapping virtual objects with regard to their corresponding position in the real environment needs a significant amount of effort for simple actions such as detection and tracking \cite{langlotz2012sketching,langlotz2011robust,macintyre2002estimating }. As the first research prototype of augmented reality outdoors, Traveling Machine is considered groundbreaking. A head-mounted display containing map navigation information and a GPS device are included in the prototype. Using a pointer and touch surface, users interact with AR maps \cite{feiner1997touring}.
A recent AR headset, however, has shown great advancements, particularly in mobility.  AR headsets can receive video and auditory feedback, but other senses remain untapped, including smell and haptics \cite{lee2018interaction}.  Interestingly, AR headsets are not the only way to view Metaverse content. AR patches, as well as digital entities from the Metaverse, can be supplied via a variety of devices, including but not limited to AR headsets \cite{lik2017interaction,lee2019hibey}, hand-held tablet devices \cite{wacker2020heatmaps}, overhead projectors \cite{xie2016large}, and tabletops \cite{roo2017inner}, Pico (portable) projectors \cite{hartmann2020aar}, and so on. Matter of fact, in terms of switching user attention and occupying users' hands, AR headsets have an advantage over other tactics. The finest aspect of this technique is that users must shift their focus between physical environments and digital material on other sorts of AR gadgets. AR headsets also allow users to visualize \cite{chaturvedi2019peripheral,zhang2016effect}, and the user's hands will not be burdened by tangible items functioning as processing elements. These advantages make AR headsets more immersive in the Metaverse via AR lenses. Augmented reality continues to evolve and spread across a wide range of applications.

\subsubsection{Mixed Reality (MR)}
A mixed reality (MR) environment is one in which physical and digital items coexist and interact in real time in a physical and virtual world. In other words, we can say that MR is a hybrid of AR and VR technologies \cite{milgram1994taxonomy}, and we will go over the role of MR in the Metaverse further. There is no formal definition of MR, but it is imperative to have a name that describes the alternate reality that exists between the two poles of AR and VR. While reading the previous literature, we came across several descriptions of MR, including a “traditional" notion of MR in Continuum \cite{milgram1994taxonomy}, MR as a generic term for AR \cite{lopes2018adding}, MR as a type of liaison \cite{reilly2015mapping}, MR as a combo of AR and VR \cite{ohta2015mixed}, MR as a synchronisation of environments \cite{roo2017inner}, and a “greater" version of AR \cite{yue2017scenectrl}. MR is a combination of AR and VR that allows users to interact with virtual elements in physical environments, according to the scientific community. Nowadays, mobile AR is the most popular mixed reality service on social media. Mix reality events are created by AR filters on Instagram. Microsoft Windows Mixed Reality combines all user experiences to create holographic portrayals of people, 3D models, and their real environment. In comparison, as seen in existing applications \cite{lee2018interaction}, AR typically shows information layered on physical locations without regard for compatibility. Due to the massive aforementioned characteristics of MR, it is regarded a more powerful version of AR, with collective linkages to physical spatial, user interaction, and virtual entities \cite{lee2018interaction, lee2020towards, malinverni2017world, gardony2020eye}. Humans, computers, and the surroundings can interact naturally and immersively with mixed reality. The world of computer vision, graphics, display technologies, input methods, and cloud computing has changed dramatically. Sensor and processing power breakthroughs are resulting in novel computer perceptions of surroundings based on innovative input modalities. A person's body position in the physical world, objects, surfaces, and boundaries, environmental lighting and sound, object classification, and geographical locations can all be captured via environmental inputs.

\subsection{Background of B5G/6G}
To fully utilize the Metaverse, we must have smooth outdoor mobility enabled by cellular networks such as the 5G and 6G networks. With the advent of 6G technology, 5G has been ruled out, but it is still being rolled out to replace the aging 4G standard \cite{gustavsson2021implementation}. As shown in the Table \ref{table:softarchitectures} below, we have laid out different ways in which enabling technologies play a key role from 1G to 6G. 6G networks are being developed and are projected to be speedier, with increased capacity and fewer latency \cite{giordani2020toward}. It can be seen that 6G networks will encompass a wide range of capabilities, as well as current mobile applications such as VR/AR, AI, and the IoT. Mobile network operators are also predicted to use flexible distributed approaches for 6G, including local spectrum licensing, bandwidth allocation, and connectivity sharing. Mobile edge computing, short-packet communication, and blockchain technology would be responsible for all of this.
As evidence suggests that new wireless technologies are released regularly after every ten years, it may be anticipated that 6 will be in use by 2030. In the future, more designs and protocols will be launched in 6G. Users will continue to use more IoT, mobile devices, and similarly use a lot of data at increasing rates as 6G evolves, indicating the more theoretical aspects of 6G are predicted to emerge \cite{tang2022roadmap, chang20226g}.
\subsubsection{5G evolution}
Although 6G is replacing 5G \cite{rao2018impact}, however, 5G technologies are still required. Wireless telecommunication industries continue to rely on 4G for consumer use, and 5G is attempting to replace the outdated technology, albeit gradually, with most deployments taking place in tightly populated regions. In 2009, the 4G/LTE \cite{glisic2006advanced} standard was introduced, which proved to be a game-changer for mobile devices by increasing data speed and letting users watch HD movies, play online games, and exchange enormous amounts of data at speeds of up to 33 Mbps.
5G \cite{rao2018impact} outperforms 4G by using microwave and mmWave technology to boost its speed to roughly 900 Mbps or more. Faster speeds and capacity are comparable to commercial broadband providers, offering more uses beyond streaming media. IoT and edge computing will have real-time sensing capabilities, which means they will obtain data instantaneously by leveraging the cloud. In the healthcare profession, such as medical services, instantaneous knowledge from patients is required for diagnosis and treatment, which can only be accomplished with faster computations; therefore, 5G has applications in retail and industry. The list is virtually endless, but we will not see 5G's entire potential until the advanced technology is implemented globally.

\definecolor{Gray}{gray}{1}
\definecolor{Gray1}{gray}{0.95}
\bgroup
{\renewcommand{\arraystretch}{1.1}
\begin{table*}[t!]
\centering
	\caption{Role of Key Enabling Technologies from Pre-4G to 6G } 
	\scalebox{0.75}{
	\setlength\tabcolsep{2pt}
\begin{tabular}{ccccc}
\toprule
\rowcolor{Gray}
\textbf{Generations} & \textbf{Pre 4G} &
  \textbf{4G} &
  \textbf{5G} &
  \textbf{6G}\\ 
\toprule

\rowcolor{Gray1}
Advantages & \begin{tabular}[c]{@{}c@{}} \tabitem Improve voice clarity \\ \tabitem The network uses the analog signal (1G) \\ \tabitem Consume less battery power \\ \tabitem Data and voice signals are \\digitally encrypted \\ \tabitem Fixed and variable data rates
\\ \tabitem Asymmetric data rates (3G)
\end{tabular} & \begin{tabular}[c]{@{}c@{}} \tabitem High speed \\ \tabitem Tight network security
\\ \tabitem High usability: anytime, anywhere\\ and any with technology
\\ \tabitem Support for multimedia \\ services low transmission cost
\\ \tabitem Low cost per bit
 \end{tabular}  & \begin{tabular}[c]{@{}c@{}}  \tabitem Greater speed in the transmissions \\ \tabitem A lower latency \\ \tabitem Greater capacity of remote execution \\ \tabitem A greater number of connected devices
 \\ \tabitem Possibility of implementing virtual networks
  \\ \tabitem Your PCs can be controlled by handsets

 \end{tabular}
 &  \begin{tabular}[c]{@{}c@{}} \tabitem AI-assisted intelligent\\ connectivity \tabitem Depth connection
\\ \tabitem The term “holographic connection"\\ refers to the use of AR/VR to \\provide continuous coverage\\ anywhere
\\ \tabitem A ubiquitous connection that \\spans space, air, ground, and sea
\\ \tabitem By removing time, space and\\ workflow barriers,  6G will transform \\the health-care industry
 \end{tabular}
  \\ \midrule
 
\rowcolor{Gray}
Disadvantages & \begin{tabular}[c]{@{}c@{}} \tabitem Poor voice quality \\ \tabitem Large phone size \\ \tabitem Poor battery life \\ \tabitem  Complex data, such as videos, \\ could not be handled by the system \\ \tabitem Different handsets are required
\\ \tabitem Too little bandwidth
\\ \tabitem Consumption of energy is high
\\ \tabitem Spectrum license cost
\\ \tabitem High expenses of 3G phones
\\ \tabitem 3G compatible handset
\\ \tabitem Connection rate
\end{tabular} & \begin{tabular}[c]{@{}c@{}} \tabitem The battery uses is more \\ \tabitem Hard to implement
\\ \tabitem Need complected hardware
\\ \tabitem It required the use of 4G technology
\\ \tabitem It is still costly to build a\\ next-generation network
\\ \tabitem The network has more problems \\has security issues
 \end{tabular}  & \begin{tabular}[c]{@{}c@{}}  \tabitem  Old devices are unable to handle 5G, \\so they must be replaced by new ones \\ \tabitem  A costly process \\ \tabitem Developing infrastructure\\ requires a high price tag
 \\ \tabitem There is yet to be a resolution to the\\ privacy and security issues

 \end{tabular}
 &  \begin{tabular}[c]{@{}c@{}} \tabitem Multi-connection architecture\\ and cell-less architecture\\ are used in 6G. In a cell-less\\ deployment, UEs are not \\connected to a single cell,\\ but to the RAN. Network \\architecture needs to be redesigned \\ \tabitem As part of its \\communication, 6G uses visible \\light frequencies; therefore, \\its drawbacks could be considered\\ to be those of 6G wireless technology
\\ \tabitem It is challenging to design communication\\ protocol stacks for network and \\terminal types of equipment
 \end{tabular}
  \\ \midrule

\rowcolor{Gray1}
Applications & \begin{tabular}[c]{@{}c@{}} \tabitem Voice calling \\ \tabitem SMS
\end{tabular} & \begin{tabular}[c]{@{}c@{}} \tabitem HD movies \\ \tabitem Video conferences
\\ \tabitem Telephony
\\ \tabitem IoT, Wearable, Smart manufacturing 
\\ \tabitem Edge computing, Smart farming, AI
 \end{tabular}  & \begin{tabular}[c]{@{}c@{}}  \tabitem Assistive Technology \\ \tabitem Immersive AR/VR, Advanced AI \\ \tabitem Autonomous vehicles,\\ Decentralized business \\ \tabitem A greater number of connected devices
 \\ \tabitem 6G has a high data rate (Tb/sec) \\and a short latency (sub-ms).

 \end{tabular}
 &  \begin{tabular}[c]{@{}c@{}} \tabitem AI-assisted intelligent connectivity \\ \tabitem Depth connection
\\ \tabitem The term “holographic connection"\\ refers to the use of AR/VR to \\provide continuous coverage anywhere
\\ \tabitem A ubiquitous connection that spans\\ space, air, ground, and sea
\\ \tabitem By removing time, space and\\ workflow barriers,  6G will transform \\the health-care industry
 \end{tabular}
  \\ \midrule

\rowcolor{Gray}
Application \\ Types & \begin{tabular}[c]{@{}c@{}} \tabitem Text messages,Picture messages and MMS \\ \tabitem Wireless voice telephony,\\ Mobile Internet access, Fixed wireless \\ \tabitem Internet access, Video calls, Mobile TV
\end{tabular} & \begin{tabular}[c]{@{}c@{}} \tabitem Amended mobile web access, \\IP telephony, Gaming Services \\ \tabitem Video conferences
\\ \tabitem High definition mobile TV, \\Video conferencing, 3D television
 \end{tabular}  & \begin{tabular}[c]{@{}c@{}}  \tabitem eMBB \\ \tabitem uRLLC \\ \tabitem mMTC 

 \end{tabular}
 &  \begin{tabular}[c]{@{}c@{}} \tabitem MBRLLC  \\ \tabitem mURLLC
\\ \tabitem HCS
\\ \tabitem MPS
 \end{tabular}
  \\ \midrule
  
  \rowcolor{Gray1}
Application Types & \begin{tabular}[c]{@{}c@{}} \tabitem Text messages, Picture messages and MMS \\ \tabitem Wireless voice telephony,\\ Mobile Internet access, \\Fixed wireless \\ \tabitem Internet access, Video calls, Mobile TV
\end{tabular} &  \begin{tabular}[c]{@{}c@{}} \tabitem Amended mobile web access, \\IP telephony, \\Gaming Services \\ \tabitem Video conferences
\\ \tabitem High definition mobile TV, \\Video conferencing, 3D television
 \end{tabular}  & \begin{tabular}[c]{@{}c@{}}  \tabitem eMBB \\ \tabitem uRLLC \\ \tabitem mMTC 

 \end{tabular}
 &  \begin{tabular}[c]{@{}c@{}} \tabitem MBRLLC  \\ \tabitem mURLLC
\\ \tabitem HCS
\\ \tabitem MPS
 \end{tabular}
  \\ \midrule
  
  \rowcolor{Gray}
Device Types & \begin{tabular}[c]{@{}c@{}} \tabitem FDMA \\ \tabitem TDMA \\ \tabitem CDMA \\ \tabitem W-CDMA \\
\tabitem GSM, EDGE, UMTS, DECT \\ \tabitem WiMax, CDMA 2000
\end{tabular} & \begin{tabular}[c]{@{}c@{}} \tabitem ADVANCED LTE, WirelessMAN-Advanced, \\IEEE 802.16m, Three-GPP Long Term Evolutions \\ \tabitem WiMAX (IEEE 802.16e) for mobile devices
\\ \tabitem China's TD-LTE network
\\ \tabitem The UMB (formerly the EV-DO Rev. C)
\\ \tabitem Flash-OFDM
 \end{tabular}  & \begin{tabular}[c]{@{}c@{}}  \tabitem Smartphones \\ \tabitem Sensors \\ \tabitem Drones 

 \end{tabular}
 &  \begin{tabular}[c]{@{}c@{}} \tabitem MBRLLC  \\ \tabitem Sensors and \\ DLT devices
CRAS
\\ \tabitem XR and BCI equipment
\\ \tabitem Smart implants
 \end{tabular}
  \\ \midrule
  
  \rowcolor{Gray1}
 Rate \\Requirements & 2.4 Kbps to 2 Mbps & 33.88 Mbps  & 40 to 1100 Mbps & Up to 1 Tbps

  \\ \midrule
  
    \rowcolor{Gray}
 Rate \\End-to-End \\ Delay Requirement &  20 to 100 ms &  10 ms  & 5 ms &  $<$ 1 ms
 \\ \midrule
 
 \rowcolor{Gray1}
Rate Processing\\ Delay & 100-1000 ms & 20-30 ms  & 100 ns & 10 ns

  \\ \midrule
  \rowcolor{Gray}
Architecture & Wide Area Network & Hybrid Network  & \begin{tabular}[c]{@{}c@{}} \tabitem Denser sub 6 GHz small cells \\with umbrella macro base stations \\ \tabitem $<$ 100 m tiny and dense mmWave cells
 \end{tabular}  & \begin{tabular}[c]{@{}c@{}}  \tabitem High-frequency cell-free smart\\ surfaces powered by mmWave\\ tiny cells that can be accessed\\ by mobile and fixed devices
 \end{tabular}
  \\ \midrule

\label{table:softarchitectures}
\vspace{-10pt}
\end{tabular}}
\end{table*}
}
\egroup

\subsubsection{6G and Internet of Everything}
 As already mentioned, 6G \cite{giordani2020toward} is still in development and is too early to have accurate information on its technology or functionality. We anticipate that the telecommunications industry will use dynamic, decentralized marketing strategies with local spectrum licensing/sharing and infrastructure sharing in 5G. 6G is fully integrated with Internet-based systems that enable speedy communication between users, gadgets, automobiles and the surrounding environment \cite{saad2019vision}. As a result, it is possible to estimate when we will transition from the Internet of Things (IoT) to the Internet of Everything (IoET), which 6G could theoretically deliver.
The existing fibers can provide up to 1 Gbps; however, the researchers believe that a 6G network could provide thousands of times quicker than optical fibres. The Federal Communications Commission (FCC) has opened the door to 6G speeds in 2019, enabling firms to experiment with terahertz waves in the 95-GHz to 3 THz spectrum. 5G, on the other hand, uses a low band as low as 24 GHz to 40 GHz, such as mmWave and microwave technology. Although terahertz waves could enhance 6G speeds to 1 Tbps, they would be subject to the same restrictions as 5G.
Using N-Polar gallium nitride high-electron-mobility transistors (HEMTs), researchers from the University of California, Santa Barbara have created a device to speed up the 6G development process. These HEMETs feature a junction between two materials with differing bandgaps that work as a channel instead of the common doped region found in MOSFETS, allowing the device to operate at significantly higher frequencies (140 to 230 THz), as required by 6G\cite{shrestha2020high}. As time passes, new advancements in 6G emerge. Last year, researchers from Singapore's Nanyang Technological University and Japan's Osaka University produced a device for terahertz waves that could be used for 6G \cite{Andy2021}. Engineers designed millimeter wave products for G-bands that function in terahertz waves this year. A few tools can help 6G become a reality and progress beyond the research phase. It may take the time that 6G becomes increasingly widespread in real-time applications to roll out 5G progressively.

\subsection{AR/VR/MR Deployments on 5G/6G Networks}

Although AR/VR technology has existed for a couple of years, its scale adoption depends on 5G/6G technology and edge computing. In order to enable the use cases, 5G/6G must bring ultra-low latency and high bandwidth.
As a result of private 5G/6G solutions, many industrial and enterprise customers can carry out mission-critical processes with the capabilities necessary to accomplish this. While public 5G/6G networks do not extend adequate coverage, do not deliver a particular ability to the required level, or are deemed to be unreliable, private 5G/6G networks provide the capabilities required for mission-critical applications.

\begin{table*}[]
\centering
\caption{State-of-the-art on AR/VR/MR implementations in various industry verticals envisioned by 5G and 6G technology.}
\label{5g/6g}
\resizebox{\textwidth}{!}{\begin{tabular}{|c|c|c|c|c|c|}
\hline
\rowcolor{Gray1}{\color[HTML]{0E101A} \textbf{Reference}} &
  {\color[HTML]{0E101A} \textbf{Years}} &
  \textbf{Aims and Objectives} &
  \textbf{Contribution}&
  \textbf{Limitation} &
  \textbf{Taxonomy Parameters}\\ \hline
 \cite{fernandez2019review} &
  {\color[HTML]{0E101A}} 2019 & {\color[HTML]{0E101A} \begin{tabular}[c]{@{}c@{}}IoT and AR/VR are being used in modern industries \\ to demonstrate how they can survive\end{tabular}} &
  \begin{tabular}[c]{@{}c@{}} Analyze the benefits and challenges\\ of integrating AR with blockchain. \end{tabular} &
  \begin{tabular}[c]{@{}c@{}} A hypothetical scenario is considered \\ rather than a reality scenario \end{tabular} & BC based AR/VR/MR \\ \hline
\rowcolor{Gray1}{\color[HTML]{0E101A}} \cite{cannavo2020blockchain} &
  2020 & \begin{tabular}[c]{@{}c@{}} Observing how VR, AR, and AR have been \\integrated to benefit companies and industries.  \end{tabular} &
  \begin{tabular}[c]{@{}c@{}} The system has the potential to eliminate  \\the need for third parties. \end{tabular} &
 There are limited use cases in the scenario & BC based AR/VR/MR \\ \hline
\cite{french2020interaction} &
  {\color[HTML]{0E101A} 2020} &
   \begin{tabular}[c]{@{}c@{}} To see how blockchain can be \\ integrated with mobile devices and AR/VR \end{tabular} &
 \begin{tabular}[c]{@{}c@{}} The efficiency of VR combined\\ with blockchain technology and augmented reality  \end{tabular} &
   AR is not taken into account & BC based AR/VR/MR\\ \hline \rowcolor{Gray1}{\color[HTML]{0E101A}}
  \cite{nguyen2020blockchain} &
  2020 & 
\begin{tabular}[c]{@{}c@{}}An observation of blockchain's convergence with 5G and 6G.  \\  networks for the development of immutable, distributed and smart applications\end{tabular} &
  Practical overview and solution of AR/VR &
  Does not address existing problems with 5G networks & 5G/6G enabled AR/VR/MR
  \\ \hline
 \cite{el2020block5gintell}  &
  2020 &
  5G network problems countered by AI  &
  Taxonomy for 5G networks at a high level &
AI and 5G problems are not discussed & 5G/6G enabled AR/VR/MR \\ \hline
 \rowcolor{Gray1}{\color[HTML]{0E101A}} \cite{haddad2020blockchain} &
  2020 &
  Demonstration of authentication and key protocols for 5G and 6G networks &
  \begin{tabular}[c]{@{}c@{}}In blochain applications, various stats \\ are observed in various situations\end{tabular}
  &
 Threats to security are not discussed & 5G/6G enabled AR/VR/MR \\ \hline
   \cite{chang20226g} &
  2022 & A metaverse system based on 6G edge intelligence was investigated & \begin{tabular}[c]{@{}c@{}} The organic integration of edge AI \\ and metaverse enabled by 6G\end{tabular}   & Edge node computing power distribution is imbalanced.&5G/6G enabled AR/VR/MR 
  \\ \hline
  \rowcolor{Gray1}{\color[HTML]{0E101A}} [1] &
  2022 & A metaverse vision towards enabling 6G wireless development is presented & \begin{tabular}[c]{@{}c@{}} Identifying key enablers, such as interactive experience\\ technologies, avatars, and digital twins and their potential\\ benefits for wireless systems and mobile services\end{tabular}   &Wireless systems that self-configure only & 6G enabled AR/VR/MR
  
  \\ \hline
  
\end{tabular}}
\end{table*}

The metaverse proves to have huge potential when it comes to securing the AR/VR/MR space in the future. It is designed to allow its users to build a full-featured virtual world governed by a set of rules without the involvement of platform developers or the risk of being exposed to their risks. The second benefit of BC is that it allows copyright protection to be applied to the content as a user may record it. Finally, BC can enhance the popularity of VR by merging the market for cryptocurrencies with VR, resulting in increased profitability due to the merger. There are several aspects of AR/VR/MR that could benefit from the use of 6G that are outlined below. 

\begin{itemize}
    \item With the advent of 6G networks, it is possible to communicate at THz in the submillimetre range with extremely low latency. As part of 6G, virtualized service sets simplify holographic communication over physical boundaries and enhance management capabilities. As a result, 3D imagery can be experienced in real-time (XR) in conjunction with autonomous driving, etc.

\item For data security in AR/VR applications, it is essential to ensure trusted decentralization among multiple nodes to ensure massive data-sharing. 
    \item The fast display and processing of images and videos required by AR/VR assets requires that the data be stored and exchanged through a local central server due to the amount of memory and processing required limitations. 
    \item Since AR/VR-oriented applications
     AR \cite{lopes2018adding,ohta2015mixed, yue2017scenectrl} consume a great deal of data and bandwidth; the central server could become overloaded due to multiple asynchronous communications among the AR/VR devices. 
    \item For military and battlefield applications that use AR/VR devices, Blockchain improves cybersecurity through sharing sensitive data, law enforcement, and public safety. 
    \item By creating a user-defined marketplace for storing and uploading AR content, Metaverse can assist in the commercialization of AR devices.
\item In addition to assisting users and developers in downloading and uploading content, communication can also facilitate the development of new applications and set up marketplaces and storefronts for AR/VR/MR devices to make them more commercially viable.

\end{itemize}

A summary of recent papers incorporating AR/VR/MR \cite{lopes2018adding,ohta2015mixed, yue2017scenectrl} into 5G/6G networks can be found in the Table \ref{5g/6g}.The adoption of AR and VR at scale has yet to become mainstream. 5G and edge computing can now overcome some of these challenges. We explore specific use cases and real-life applications in sections VII.

\section{Role of AI in Metaverse} 
In this section, we present the AI pyramid for Metaverse technology, followed by methods related to different learning paradigms, computer vision, and Edge AI. This subsection also serves as a basis for discussing sustainable metaverse in Section VI. Post Facebook's name change to Meta, the rejuvenation of the Metaverse field has been the talk of the research community. Another reason for inclined interest towards the Metaverse is its impact on socioeconomic and technological aspects, such as combining 3D animation, virtual reality, blockchain, non-fungible tokens, digital economy, and many others. The non-fungible tokens (NFTs) created a lot of hype that explained the role of blockchain in the context of the Metaverse, however, this section will mainly focus on the role of AI in the said technology. Recently, an article was surfaced, which explained the concept of the Metaverse in the context of value chain \footnote{https://medium.com/building-the-Metaverse/the-Metaverse-and-artificial-intelligence-ai-577343895411}. The article summarized the Metaverse as the (real-time activity-based Internet) and proposed an abstract layered architecture for the same. We use the seven-layered Metaverse architecture as a reference to explain the role of AI, as shown in Figure \ref{FigS-A}. We briefly explain the areas and define the equivalent AI mapped role in the layered schema to understand the diversity and advancements that AI brings to the field of Metaverse. This correspondent has already been performed by David Pereira \footnote{https://towardsdatascience.com/how-ai-will-shape-the-Metaverse-4ea7ae20c99}, but this study further elaborates on the use of AI in each layer, accordingly. We explain each of the corresponding layers in a bottom-up approach.
\begin{figure*}[ht]
\includegraphics[width= \linewidth]{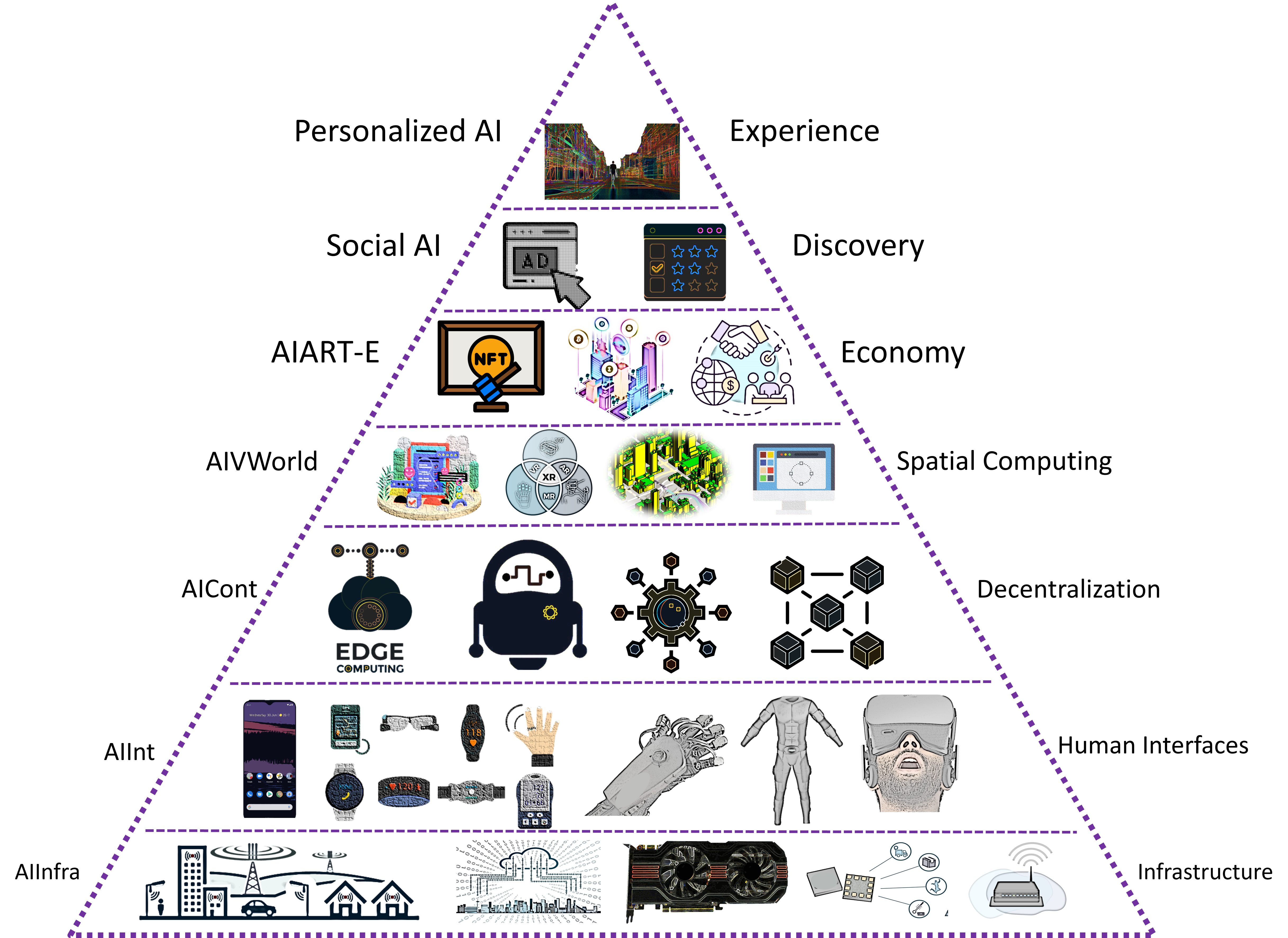}
\caption{Layered Architecture of Metaverse by Jon Radoff. The labels on the right side are original layers defined in "Building the Metaverse", while the labels on the left side exhibit equivalent mapping in the context of AI.}
\centering\label{FigS-A}
\end{figure*}
\begin{figure*}[ht]
\includegraphics[width= \linewidth]{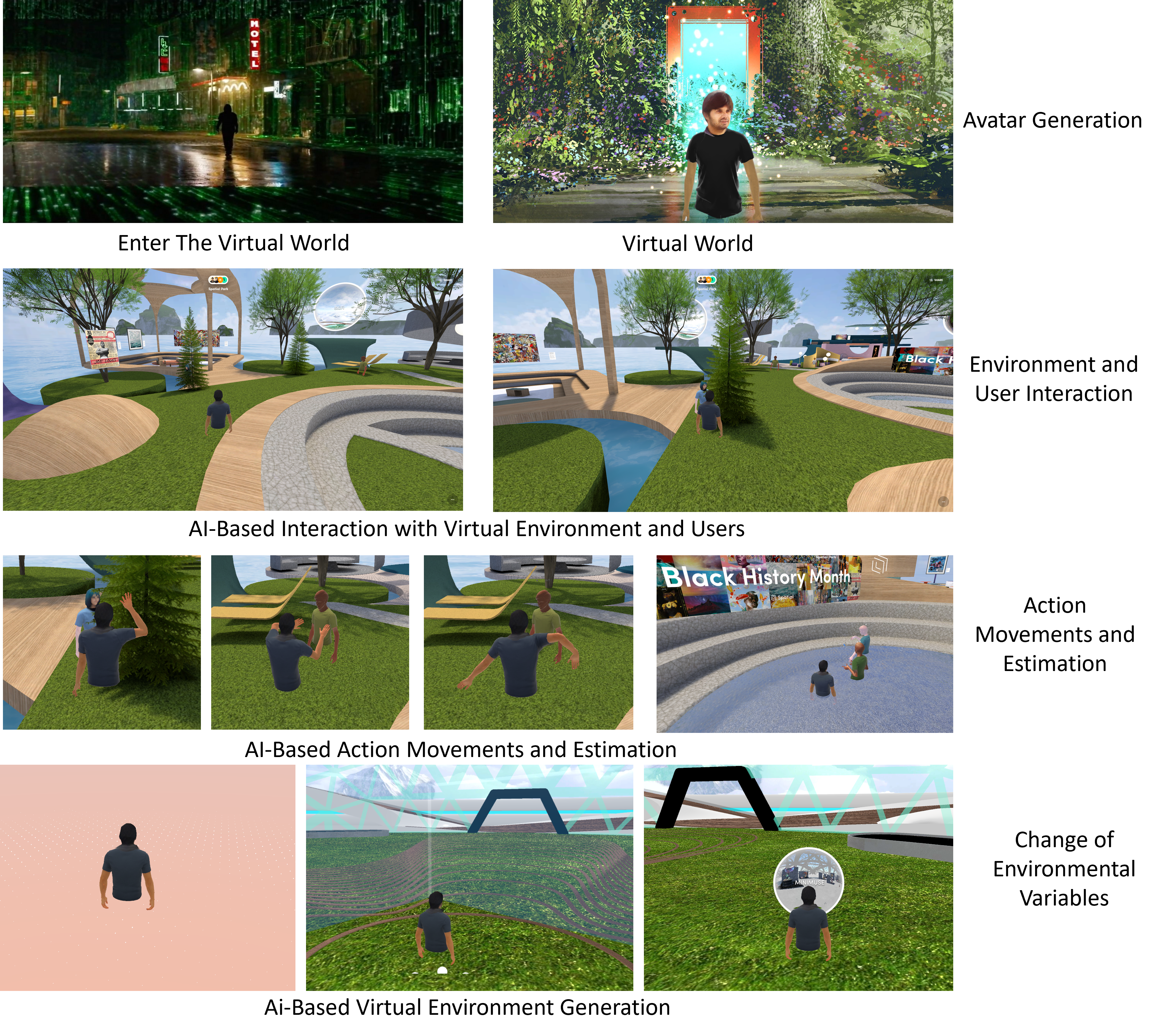}
\caption{AI in Avatar and Environment Design for Metaverse: Image courtesy: https://spatial.io/}
\centering\label{FigS-B}
\end{figure*}

\textbf{AIInfra}: It is apparent that with increasing popularity, a large number of users will opt to use the services associated with Metaverse. In this regard, the infrastructure should be capable of accommodating massive machine-type communication (mMTC) devices along with the support for executing AI-based models. Furthermore, it should be noted that the infrastructure is the base for the services being offered at stacked layers, therefore, the support for temporary storage of data or support for cloud connectivity is also necessary. For instance, 5G/6G networks already allow support for AI devices and operations. Similarly, Microelectromechanical (MEMs) devices enable the wearables to perform automatic sensing and data collection, and Graphic Processing Units (GPUs) enable the device to perform faster computations. Together, the AI-based devices (AIDev) provides the necessary infrastructure to perform decision-level operations and allow the Metaverse services to operate in a seamless manner, hence the name, AI-based Infrastructure (AIInfra). 

\textbf{AIInt}: Metaverse services heavily rely on user experiences and, to make them satisfactory, design of user interface is the initial step. Although the design of interactive and intuitive user interfaces helps improve social interactions, it can be a hindrance to some of the users that include disabled, introvert, and specially challenged people. To make the interface accessible in general, AI methods can be extensively used (AIInt). For instance, usage of computer vision approaches for visually challenged users, or generating automated voice from text for visually impaired ones, design of social avatars for friendly interaction in virtual world, brain computer interface for specially challenged users, and so forth. AI-based methods can not only be used for enabling the user-experience in terms of interface design for some users, but also can be used to improve the accessibility and interaction with the interfaces, in general.

\textbf{AICont}: Decentralization has been extensively achieved through smart contracts and blockchain based methods, however, Metaverse emphasizes on democratization along with the decentralization. Decentralization allows users and creators to protect ownership and exchange entitlements and assets easily in the digital space; however, it does not ensure disintermediation from corporations, financial institutions, and investors. For instance, let us consider the Adidas NFT case, which was dropped in December 2021 with a constraint that purchases were limited to 2 persons. The NFT was sold out in less than 1 second because a single user purchased 330 of them in a single transaction. This was accomplished by modifying smart contracts, which proves that the democratization cannot be achieved by just using conventional contracts. Democratization refers to the equal opportunity for all users, thus snatching away the power from big corporations. In this regard, AI can be used to detect modifications and anomalies concerning smart contracts, hence the name AI-based smart contracts (AICont). Furthermore, AICont can leverage the information acquired from AIInt and infrastructure to detect non-democratic activities for preventing such kind of incidents in blockchain transactions. 

\textbf{AIVWorld}: The creation of virtual environments is of vital importance in Metaverse, and making it believably realistic is one of the research challenges to solve. Some examples of digital worlds are NVIDIA's Omniverse\footnote{https://developer.nvidia.com/nvidia-omniverse-platform} and Spatial IO\footnote{https://spatial.io}. These platforms provide a set of components to simulate the real-world objects and create unique digital worlds in quite an impressive manner. The use of AI is extensive in creating this set of components as the process needs to be autonomous, realistic, and visually pleasing (AIVWorld). These virtual worlds can be used for variety of applications such as building massive worlds, creating simulation environments for robots and autonomous vehicles, simulating voice-based commands in the digital world, and so forth. AI will be responsible for not only creating realistic digital worlds, but also testing automation within the Metaverse. Some snapshots from Spatial IO virtual environment are shown in Figure \ref{FigS-B}\footnote{https://spatial.io/}.

\textbf{AIART-E}: Non-fungible tokens (NFTs) have proven to be a stepping stone in Metaverse from an economic perspective. Although, NFTs are generated by the artists, but AI can be used to generate unique art as well. Considering the expansion of AI usage among noncoders (thanks to GPT-3 and similar models), the users will be able to generate their own NFTs. The advances in natural language processing (NLP) have allowed AI to generate art for economic enrichment (AIART-E) in the digital space. Unique stories can be generated using models such as GPT-3 \cite{Floridi2020, GPT3}, while realistic unique art can be generated from a single line of text using DALL-E \cite{DALLE1, DALLE2} and GauGAN2 \cite{GauGAN}. There are also community-driven platforms governed by autonomous artists that build algorithms to create arts. One of such community-driven platforms is Botto\footnote{https://botto.com/}. Some creations from Botto, GauGAN2, and DALL-E are shown in Figure \ref{FigS-C}\footnote{https://botto.com/} \cite{DALLE1, GauGAN}. 
\begin{figure*}[ht]
\includegraphics[width= \linewidth]{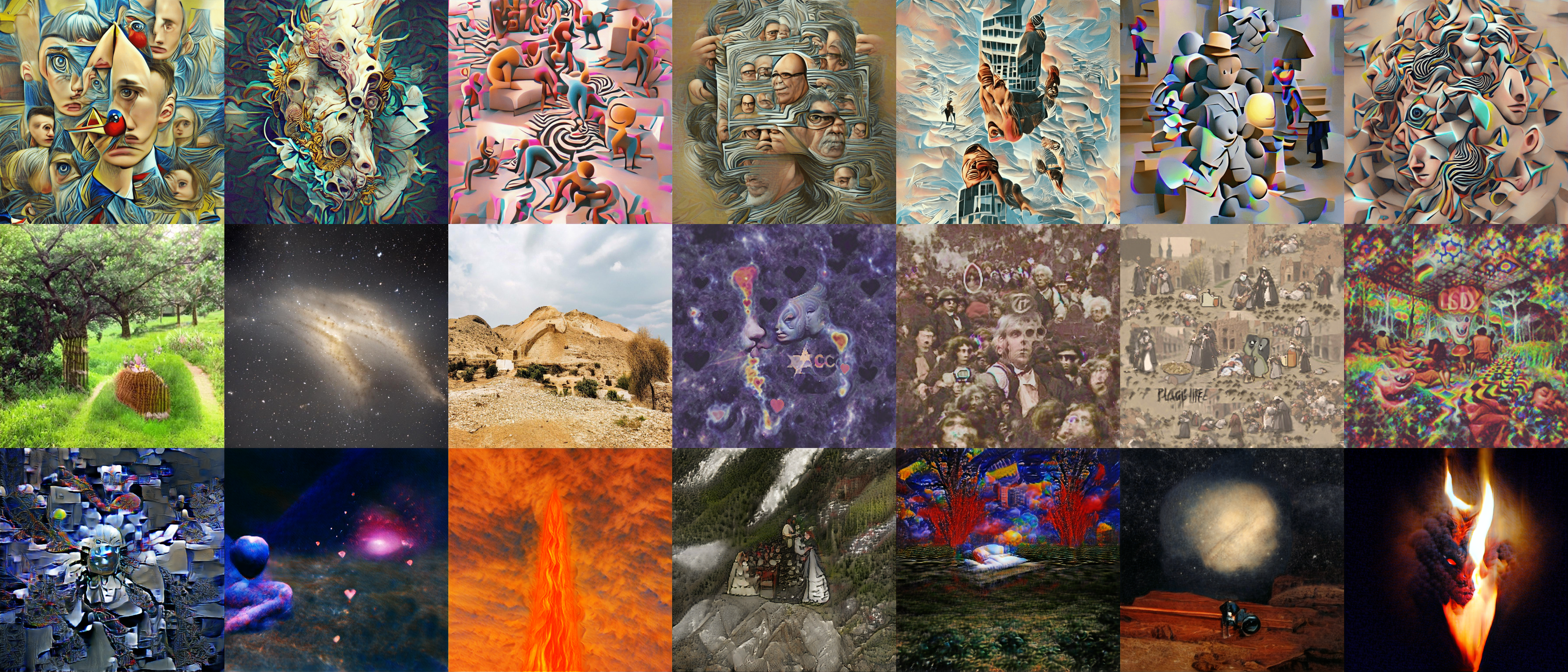}
\caption{Images generated with Botto, GauGAN2 and DALL-E}
\centering\label{FigS-C}
\end{figure*}

\textbf{SocialAI}: The core motivation behind Metaverse is to create an interactive environment that improves the social network experience. However, with current social networking platforms, there is a lot of abuse, hate, and serious concerns in terms of safety and inclusion for children and minorities. Such constraints can contribute to the depowering of digital personalization and self-knowledge, which are the things that Metaverse stands for. The use of artificial intelligence can be leveraged to improve the experience of social networks (social AI). The techniques can be used for the machine learning observability, focus more on Explainable AI that can avoid bias, compute content relevance with respect to the minority groups belonging to different ethnicity, culture, geography and language, and detect hate and abusive speech. 

\textbf{PersonalizedAI}: The final layer in Metaverse pyramid is the hyper-personalization in terms of experience. In terms of AI usage, the former layers leverage the techniques to achieve goals that ultimately contribute to enhance the user's performance. However, the experience enhancement was applied in a general manner. In this layer, the goal is to provide the user with a personalized experience that is specific to each user in the Metaverse. For instance, adjust the gaming models with respect to users' emotions or mental well-being, or apply personalization by considering the users' disabilities. Personalized AI can be explored in different ways for a variety of applications such as education, healthcare, gaming, socialization, sports, and others using real-time analytics. 

\subsection{Learning Paradigms for Metaverse}
Over the years, applications in Metaverse such as robotics, embodiement, virtual reality, augmented reality, digital rendering, and more saw remarkable progress through supervised learning techniques \cite{bishop_2016}. However, the supervised learning strategy tends to make the Metaverse application task dependent, as it is highly based on human supervision. Although, diversified learning strategies such as reinforcement learning \cite{francois_2018}, self-supervised learning \cite{Liu_2021}, and more have been proposed to make the Metaverse application task independent, its still a work in progress. The Meta AI research group is exploring the self-supervised paradigm, specifically in the domain of text and speech, to make advances in task-independent learning networks. Recently, Meta AI research released the data2vec model which is presented as a unified solution for the speech, vision, and natural language processing task while leveraging multiple modalities \cite{data2vec_2022}. Currently, Meta AI is working towards more unified models that could leverage multiple modalities while helping in reading lips and understanding the speech semantics to make leaps of progress in digital reality. While Meta is exploring the self-supervised paradigm, its rival Google's DeepMind has stuck with supervised learning and uses massive amounts of data from VRChat to train a digital agent on how to interact with humans in a simplified environment. In 2018 alone, 16 million hours of VR data was generated. Google DeepMind used 20,000 hours of VRChat data to train a digital agent to learn ground language while interacting with humans\footnote{https://www.youtube.com/watch?v=b-fvsi9YIP4}. The study showed that the digital agent becomes curious and autonomous with such a training strategy. According to the experiments, the DeepMind suggested that it could pave the way for understanding how the Metaverse could be equipped with life like intelligence. 

\subsection{Computer vision for Metaverse}
In the field of computer vision, Metaverse has gained the most progress, as researchers can replicate human-like avatars or hologoraphic images of loved ones in the digital world. William Shatner who played Captain James T. Kirk of the starship Enterprise in vintage star trek television series is one of the most beloved actors. An AI startup StoryFile\footnote{https://storyfile.com} acquired hours of William Shatner's footage while answering questions using volumetric cameras. The footage was acquired with a green screen so that William Shatner's image can be easily segmented, and a proprietary model Conversa was used to associate answers with the asked questions. Currently, the startup built a lifelike videobot of the aforementioned actor. Although, the startup used green screen, there are many deep learning based semantic segmentation methods that does not need a green screen to get a person segmented in even a complex scene, such as FCN \cite{FCN2015}, UPerNet \cite{UPPSU2018}, BiSeNet \cite{BiSeNet2018}, FPN \cite{FPN2019}, SFNet \cite{SFNet2019}, SegFormer \cite{SegFormer2021}, FaPN \cite{FaPN2021}, CondNet \cite{CondNet2021}, and Lawin \cite{Lawin2022}.

There are further many projects in development that can be considered as a technological leap towards Metaverse. For instance, companies are working on the hologram of holocaust survivors. It is an innovative medium to keep stories from the past alive using artificial intelligence\footnote{https://www.cbsnews.com/news/artificial-intelligence-holocaust-remembrance-60-minutes-2020-04-03/}. A socially distant santaclaus \footnote{https://edition.cnn.com/2020/12/15/health/children-social-distance-with-santa-wellness/index.html} and avatars of the loved ones. Recently, Microsoft \cite{abramson_johnson_2020} registered a patent for an AI system that could help an individual interact with a dead loved one through its 3D digital version. Similarly, Replika AI\footnote{https://replika.ai} is also trying to address mental well-being and loneliness through a digital avatar. Apart from people, researchers are also trying to build a digital system for animals. Researchers from Ubisoft china have proposed a rendering keyframe animation method, i.e. ZooBuilder \cite{ZooBuilder2020} for animals by using OpenPose method \cite{OpenPose2021} for extracting 2D joint coordinates and 2D-3D human pose estimator \cite{PoseEstimator2018} to track the temporal relationships between connected joints. With the emergence of Metaverse, many Internet-based tools now use computer vision techniques to provide animated 3D human models or realistic animated faces, some of them include Kinetix\footnote{https://www.kinetix.tech/}, Mixamo\footnote{https://www.mixamo.com/}, and StyleGAN2 \cite{StyleGAN}.

\subsection{Edge AI for Metaverse}
The use of Edge AI would be quite crucial for realizing Metaverse, as it is highly dependent on low latency and high throughput to meet the needs of user experience. The fifth generation (5G) networks meet the user experience demands to some extended, i.e., maximum data rate of 10 Gbps and delay of less than 10 milliseconds, but as per the estimations and forecasts, the delay needs to be lower and the peak data rate needs to be higher in order to sustain user experiences in Metaverse. Thanks to the use of AI techniques in combination with edge computing, the foundation to develop emerging mission-critical applications is laid \cite{She2021}. Several deep learning, soft computing, and machine learning techniques have been proposed to achieve uRLLC, be it active user detection \cite{Dev2022}, resource allocation \cite{She2020}, scheduling problems \cite{Alsenwi2021}, and power-management problems \cite{Gu2020}, accordingly. Recently, AI has also been exploited for mobility prediction, traffic estimation, channel prediction, and spectrum management for maintaining uRLLC. For instance, SCGNet \cite{Tunze2020} and MCNet \cite{Huynh2020} were introduced to enhance the spectrum utilization efficiency while demodulating the receiver's signal accurately. In \cite{Luo2020} a combination of CNN and LSTM was introduced to predict the channel state information for improving robustness in practical 5G systems. The study \cite{Guo2019} proposed a 3D convolutional networks for forecasting cellular traffic by leverage long and short term spatial patterns from the traffic data. In summary, Edge AI techniques are very crucial in achieving low latency with high throughput that acts like a support system for high class integrated services in Metaverse. 

\subsection{6G: A requirement for AI-based Metaverse}
The services associated with Metaverse rendered by AI require the infrastructure to be scalable, services to be lightning fast which require low latency, reliability through stable connection, and security. Although the current generation of communication system, i.e. LTE and 5G, provide some of the basis to realize Metaverse its still far away from large-scale implementation due to the aforementioned characteristics. Since the metaverse needs to support atleast thousands if not millions of remote devices that need to process and transmit huge amounts of data, the latency will be the core issue for current network environment as it is considered to be the bottleneck for adaptation of wide array of applications \cite{wang2020}. Some haptic and spatial AI applications require latency of less than 1ms while the transfer rate of more than 1 Mbps \cite{Dev2022}. The fifth generation communication system alleviate this problem to some extent but as the Metaverse grows, which is assumed to be the case, a stable and faster new generation of mobile communication will be definitely required \cite{IIFNet}.\\
As discussed in previous subsections, large language models will play a key role in understanding user preferences and generating customized and unique environments. In this regard, the training of such models require computing power and a stable connection \cite{DBFL}. However, researchers have shown that the current iteration of network environment is not suitable for training large language models, rather it works better with small-scale ones, and one of the reasons for such failure is the lack of stable connection \cite{TeraPipe}. Therefore, in order to realize the metaverse with its full potential, the communication network needs to play an effective part along with the AI methods, respectively. 

We analyzed several studies \cite{IIFNet, HioTSP, FinTech, Req6G1, Req6G2}for requirement gathering concerning delays and bandwidth with respect to each layer in Metaverse pyramid shown in \ref{FigS-A} and report the figures in \ref{Tab-S-A}. The requirement is gathered with respect to delay, average inference time (AIT), average inference time with enhanced user experience (AIT-EUE), transmission bit rate (TBR), and transmission bit rate with compression (TBR-C), respectively. The AIT-EUE refers to the data enhancement techniques for enhancing user experience, for instance, image denoising, image super resolution, and more. The AIT-EUE mainly depends upon the image resolution (assuming that Metaverse mainly deals with imaging technologies). The relationship with the inference time is simple, higher the resolution, higher the inference time. The reported values are in terms of average and also hypothetical, as Metaverse and 6G are not realized to a larger extent. For AICONT, we considered the average transaction speed of cryptocurrencies that are fastest, such as Algorand and EOS. The values for average transaction speed for the cryptocurrencies and related information are obtained using Statista\footnote{https://www.statista.com/statistics/944355/cryptocurrency-transaction-speed/} and Algorand Blog\footnote{https://www.algorand.com/resources/blog/role-of-transaction-finality-speed-in-nft-minting}, while the information regarding NFTs are obtained using Chainalysis\footnote{https://blog.chainalysis.com/reports/chainalysis-web3-report-preview-nfts/}. The requirement suggests that the current iteration of communication system (5G) will not be able to meet most of the requirements, thus, the realization of Metaverse is closely tied with the emergence of sixth generation communication system which promises to have lesser delays, latency, and higher transmission bit rates, accordingly.

\begin{table*}[]
\centering
\caption{Delays and Bandwidth requirement for each of the layer in Metaverse Pyramid}
\label{Tab-S-A}
\begin{tabular}{|c|c|c|c|c|c|}
\hline
Layer          & Delay (ms) & AIT (ms) & AIT-EUE (ms) & TBR         & TBR-C    \\ \hline
AIINFRA        & 1.167      & 2.86     & -            & -           & -        \\ \hline
AIINT          & 3.5        & 3500     & 5800         & 1062 Gbps   & 530 Mbps \\ \hline
AICONT         & 500        & 4500     & -            & 500 Kbps    & 500 Kbps \\ \hline
AIVWORLD       & 3.5        & 8000     & 19300        & 63.70 Gbps  & 210 Mbps \\ \hline
AIART-E        & 3.5        & 10000    & 24400        & 10.62 Gbps  & 35 Mbps  \\ \hline
SocialAI &
  1.167 &
  \begin{tabular}[c]{@{}c@{}}1.25 (Text) /\\ 2000 (Image)\end{tabular} &
  \begin{tabular}[c]{@{}c@{}}1.37 (Text) /\\ 3400 (Image)\end{tabular} &
  500 Kbps &
  128 Kbps \\ \hline
PersonalizedAI & 150        & 19200    & 48800        & 238.89 Gbps & 796 Mbps \\ \hline
\end{tabular}
\end{table*}

\section{Role of B5G/6G in Metaverse} 
This section outlines the importance of B5G/6G services towards Metaverse in subsections IV-A and IV-B, followed by a comprehensive review of the state-of-the-art wireless communication technologies for immersive experiences in subsections IV-C and IV-D. 
\subsection{Is B5G/6G Need of an Hour?}
It is anticipated that Metaverse will be grounded on large-scale wireless cellular network technologies for the purpose of telecomputing, telecommuting, teleporting, teleoperating, telepresence, avatar interactions, and data mobility. The interactive and immersive experiences inside Metaverse will require very high data rates and low latency while transmitting high-resolution content. 5G networks could deliver data rates of up to $<$10GBs by exploiting the mmWave (30-300 GHz) spectrum \cite{ana2021study}. This frequency and data rate remained successful in fulfilling the uplink and downlink throughput requirements of several multimedia applications ranging from live streaming to connected and autonomous vehicles. However, Metaverse is likely to interconnect a wide range of such services potentially involving holographic communication and real-time haptics, requiring milliseconds of motion-to-photon latency. Consequently, the bandwidth and throughput requirements for delivering such services will be $10^3$x more compared to the existing 5G-based cellular systems \cite{lin2021wireless}. Therefore, 5G and pre-5G networks will not support the Metaverse for the following key reasons:
\begin{enumerate}
\item \textbf{Proliferation of devices:} Metaverse is expected to drive the growth of mobile and other IoT devices significantly. Thus, the connection density of 5G and pre-5G networks will not be able to accommodate the increasing number of devices.
\item \textbf{Multi-sensory communications:} Apparently, Metaverse will bring immersive services with multisensory communications that will pose stringent challenges to existing 5G QoS and QoE classes. Thus, 5G and pre-5G wireless systems will fail to provide a seamless experience to multiple immersive services such as avatar interactions or teleoperations in parallel.
\item \textbf{Complex and ultra-massive connectivity:} Metaverse is envisioned to build on prolific heterogeneous devices distributed in a highly dynamic, complex and ultra-massive network. In such a scenario, the existing 5G and pre-5G services catalog will not offer a reliable solution since those services are often designed for static and predefined optimization challenges.
\item \textbf{Decentralized services:} Decentralized intelligence will be a cornerstone in Metaverse driven heterogeneous multi-machine, multi-technology, multi-user, multi-application, and multi-device environments. Although 5G architecture can support the AI services on the network's decentralized edge, the edge intelligence required by Metaverse is beyond the original goals of current 5G networks. The transmission rate offered by 5G can not guarantee the resilient edge AI because AI models will need to be subtle and highly dynamic while serving on the edge or other devices in Metaverse.
\end{enumerate}

Pre-5G networks successfully supported the deployment of AR among retail applications, educational services, the gaming arena, and Google Maps \cite{siriwardhana2021survey}. VR/AR devices such as Google Glass\footnote{https://www.google.com/glass/start/}, Toshiba dynaEdge\footnote{https://us.dynabook.com/smartglasses/products/index.html}, Microsoft hololens 2\footnote{https://www.microsoft.com/en-us/hololens/}, etc., and mobile applications such as IKEA Place\footnote{https://www.ikea.com/au/en/customer-service/mobile-apps/say-hej-to-ikea-place-pub1f8af050}, Pokemon Go\footnote{https://pokemongolive.com/en/}, etc., have widely benefited from pre-5G networks. These apps and devices were not only self-sufficient but also leveraged Internet access via pre-5G access methods. Studies show that VR/AR users gained adequate experience via 4G LTE networks \cite{sukhmani2018edge}. However, such networks are insufficient to meet Metaverse users' expectations. Metaverse will bring in next-generation multimedia such as real-time haptic feedback, high dynamic range (HDR), and stereoscopic video formats. The estimated data rate required to render a stereoscopic HDR $360^\circ$ video of 8K resolution with 90 frames per second (FPS) is 200Mbps\cite{siriwardhana2021survey}. In addition, the holographic avatars will be rendered as a physical presence for remote users in Metaverse. The rendering of holographic avatars will be witnessed in situations where a doctor will be required to perform remote surgery, an engineer to troubleshoot in a remote factory, and a farmer to look after the grass and animals in remote geographic locations. Apart from typical graphical properties (e.g., resolution, depth, texture, or color), the holographs will be organized from various postures to ensure consistency in angles, frames, and slopes relative to the avatar. If an avatar is rendered in tiles of 4" x 4" dimensions, then an avatar with a 6' x 20" human size will require a transmission rate of 4.32 Tb/s \cite{li2018network}. According to a Third
Generation Partnership Project (3GPP) analysis on the communication aspect of healthcare services, a remote AR-based surgery can operate on approximately  $<$1ms of E2E latency and 12 Gbps of transmission rate considering an HDR 10bits stream of 4K resolution with 120fps\cite{gotsch2018telehuman2}. Metaverse is envisioned to deal with longer durations of remote surgeries, operations, and other clinical trials. To offer immersive experiences in Metaverse, additional synchronization will be required apart from high data rates to ensure seamless transmission of multisensory content delivery and user experience. Such requirements are beyond the offerings of pre-B5G/6G systems, and non-3GPP access networks, including Wi-Fi, will be impractical due to the performance benchmarks and handover issues occurring from mobility\cite{yang2018digital}. Thereby, B5G/6G plays its role in recognizing the demands mentioned above. In the subsequent section, we will entail the enabling technologies offered by B5G/6G. 

\begin{figure*}[ht]
\includegraphics[width=\linewidth]{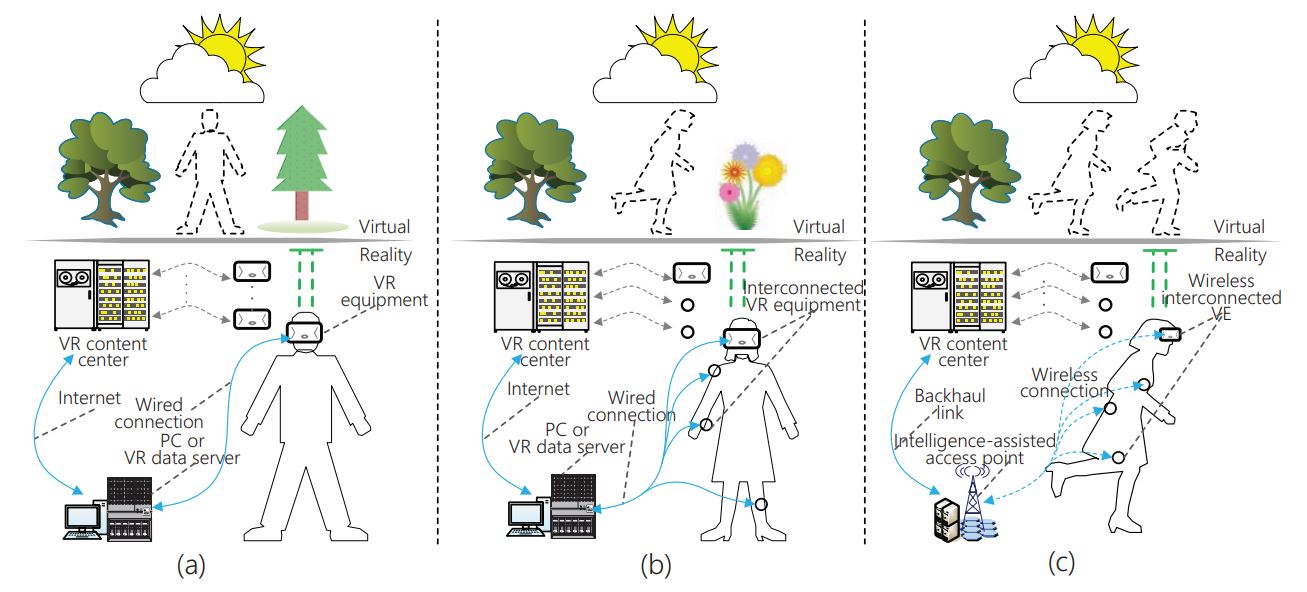}
\caption{ Evaluation of VR experiences: a) VR over wired connectivity, b) VR over interconnected network, and c) VR over fully wireless connectivity \cite{lin2021wireless}}
\centering\label{VRevo}
\end{figure*}

\subsection{What B5G/6G Brings to Metaverse?}
Since Metaverse is coined to be the next frontier in multimedia after VR/AR, it will encompass multisensory network services such as haptic and holographic communications. Soon users will undermine the QoE delivered by VR/AR and demand new lifelike multimedia experiences, unlike today’s not\textminus real\textminus enough VR/AR experiences. Metaverse applications involving avatar interactions aided with novel communication modalities such as avatar\textminus to\textminus avatar (A2A), avatar\textminus to\textminus human (A2H), and human\textminus to\textminus avatar (H2A) will drive the majority of the Internet traffic by 2030\cite{yastrebova2018future}. Thereby, the services provided by B5G/6G have a crucial role in realizing the lifelike experiences in Metaverse.

\subsubsection{\textbf{5G Services Towards Metaverse}}
5G networks offer three prime categories of services listed by the International Telecommunication Union (ITU)\footnote{https://www.itu.int}: uRLLC, mMTC, and eMBB to handle the requirements of extremely low latency, enormous content transmission and high computations. eMBB traffic requires a gigabit/s level of the transmission rate. At the same time, uRLLC expects high reliability of 99.999\% with ultra-low-latency of 0.25 $\sim$ 0.30 ms/packet, and mMTC requires highly dense connectivity with significant energy efficiency. However, uRLLC can only provide a data rate of up to 10 Mbps which might fit in a few applications of VR/AR. In contrast, eMBB can be suitable for delivering ultra-high throughput in order to meet the requirements of Metaverse users accessing multisensory content ranging from teleoperations to holoconferences to telesurgeries with 3D 4K resolutions\footnote{https://www.5g-eve.eu/}. In addition, it can offer the minimum guaranteed delivery of 100 Mbps to users. Therefore, eMBB can play a crucial role in Metaverse's gaming and entertainment paradigm by providing mobile users with 100-200 Mbps downlink with a peak threshold of 250 Mbps and latency of $<$100ms \cite{erel2019road}. 

An evolution of VR technologies can be seen in Figure \ref{VRevo}. From basic VR video streaming to interconnected VR to a fully wireless interconnected VR, a series of stages can be seen \cite{lin2021wireless}. However, unlike conventional video streaming, wireless streaming in Metaverse may require transmitting enormous omnidirectional visual content with a latency of $<$20ms. The success of such an immersive virtual experience is highly reliable on a smooth transmission mechanism of ultra-high-resolution visual content along with a highly subtle tactile feedback in downlink networks. As discussed in section IV-A, it is obvious to rely on the B5G wireless technologies to support the immersive virtual applications. Therefore, mmWave has progressed towards standardization by 3GPP in B5G wireless systems\cite{roh2014millimeter}. Simultaneously, researchers have also explored the potential of mmWave mobile networks in realizing the wireless bandwidth requirements of VR applications. According to \cite{lin2021wireless}, mmWave is able to successfully deliver high data rates with low-latency transmission for several VR applications. Similarly, the mMTC service can support dense mobile connectivity of up to 1 million devices per square kilometer which is approximately 10x greater than 4G LTE networks. A study by the European 5G Observatory\footnote{http://5gobservatory.eu/5g-trial/major-european-5g-trials-and-pilots/} reported the data obtained from 180 trials to assess the capacity of 5G in a dense urban area of Finland. The mMTC service was able to deliver a realistic data rate of 700 Mbps to 1 Gbps by accommodating a reasonable user density. Based on the overall analysis, 5G service equipped users’ devices with average data rates of 1-4.5 Gbps with a delay of $<$5ms \cite{zhang2019will}.

Since a few Metaverse applications will require a data rate ranging from 100 Mbps to a few Gbps with 99\% reliability and a delay of $<$5ms, 5G networks may fulfill these requirements efficiently. Apart from that, Metaverse users will utilize the diverse features of mobile systems ranging from display and sensing capabilities to near-user computing resources \cite{bajireanu2019mobile}. In addition, the recent innovations in vision and tracking technologies such as depth, $360^\circ$ resolution and precision have supported the idea of user’s interactions with the surroundings. These capabilities are not limited to smartphones; many devices such as fitness watches, smart TVs, game consoles, and autonomous vehicles also possess these features. These devices can leverage the services of 5G with single-digit latency, high throughputs (i.e., 10x $>$ 4G), five-nines of network reliability and dense network coverage (i.e.,$>$ 10-100x) \cite{milovanovic20215g}. In 2021, 5G smartphones entered the commercialization phase and were expected to spread in the global market within the next few years \cite{chen2022standardization}. The growth of these devices with seamless connectivity will certainly form the basis of several immersive services such as tele-classrooms, tele-surgeries, and teleoperations. Nevertheless, to deliver such services with a sufficient QoE requires very high data rates within a range of 1 Gb/s to 1 Tb/s which can be supported by B5G cellular and underlying transmission networks \cite{yeh2022perspectives}. 

\subsubsection{\textbf{6G Services Towards Metaverse}}
6G networks are envisaged to provide ubiquitous intelligence with abundant compute power and high-speed wireless data fidelity over the air, space, and sea as shown in Figure \ref{Fig9}. 6G services have been defined in terms of the real-time mesh of physical and cyber worlds, with a fine-grained focus on the interactions among physical, digital, and biological (human-centered) worlds \cite{viswanathan2020communications,docomo2021white}. An immersive merger of cyber and physical space is likely to bring new platforms in the shape of embodied Internet, providing 6G based Metaverse applications and other cyber-physical systems (CPS). 6G will exquisitely enhance user experience in the Metaverse since its services are primarily human-centric instead of data-centric. Consequently, Metaverse users and the associated processes, user equipment, and the network will be holistically integrated to provide a plethora of immersive applications.

Moreover, 6G networks will empower edge intelligence through on-device machine/deep learning and distributed AI that will transform the “connectedness with intelligence” for not only human-centric but also machine-centric applications \cite{zhang20196g}. In Metaverse, 6G will enable neural and haptic sensory communications with an integrated holographic reinforcement \cite{chang20226g}. 6G will also fulfill the requirement of Internet-of-everything by featuring the network-in-box management and virtualized services that will motivate the massive multiple access points to shape a distributed (cell-less) multiple-input and multiple-output (MIMO) system \cite{yang2018communication}. In contrast to 5G systems, 6G supports terahertz frequency bands and is liable to provide 1 Tbps of data rate with an E2E delay of 0.1ms \cite{chaccour2022seven}. It also supports extremely reliable low latency communications (eRLLC)-based services along with a reliability rate of 99.9999999\%. In principle, Metaverse users can achieve high QoE by leveraging the following 6G services: support for multi-dimensional holograms, ultra-massive data-transmission rates to support HDR $360^\circ$ multimedia with 4K/8K resolution, extreme low-latency and high precision for haptics required in 3D-printing, telepresence, immersive multi-user gaming, and responsive digital twins in industry 5.0. Furthermore, these 6G services will be underpinned by AI, such as federated learning, split computing, and distributed deep reinforcement learning to reduce network congestion and improve the user QoE \cite{adhikari20226g}. 

In summary, 6G will surpass the previous generations by catalyzing the wireless revolution from “connectivity” to “connectivity with intelligence” with the following key emerging services and utilities:
\begin{itemize}
\item \textbf{High data rates:} The usage scenarios under ultra mobile broadband (uMBB) will exploit significantly higher data rates as compared to the ones in 5G eMBB services. 
\item \textbf{Dense connectivity:} The ultra-massive machine-type communication (uMTC) scenario will empower use cases where a relatively higher and massive number of connections in a certain space will be required simultaneously as compared to the ones provided in 5G-mMTC services.
\item \textbf{High reliability and precision:} The services under ultra-high precision communication (uHPC) require a very high degree of precision, reliability and accurate positioning, making them suitable for mission-critical applications as opposed to 5G-URLLC services. These services ensure stringent service level guarantees can be enforced independently or in part with each other. 
\item \textbf{Air-to-ground connectivity:} The services under extended 3-dimensional coverage (e3DC) will be able to exploit the in-space non-terrestrial satellite communications by utilizing the ground-based terrestrial mobile networks. The realm of in-space networks includes drones, high-altitude platform stations, and air-borne to provide multi-dimensional coverage across the globe.
\item \textbf{Communication meets computation:} The proliferation of emerging smart devices will urge the need for an autonomous and distributed computing framework to facilitate key technologies such as split computing and federated learning. 6G will facilitate the computation-centric communication that will ensure the required QoS provisioning by trading off the communication resources in order to achieve sufficient computational accuracy. 
\item \textbf{Context-driven communications:}  The 6G network context, such as network architecture and bandwidth, will navigate the provisioning of 6G uMBB services to the users. It will significantly benefit the Metaverse applications since the physical context, such as the user’s location, mobility and social connections, will be highly dynamic, eventually leveraging the agile and adaptive service provisioning of 6G networks.
\item\textbf{Event-driven communications:} 6G networks will ensure the orderly provisioning of uHPC services based on certain characteristics of emergency or disaster events. The highly dynamic spatio-temporal attributes of users and devices, network traffic and infrastructure will play a key role in resource allocation in 6G-uHPC application scenarios. 
\end{itemize}

Apart from the above services and their usage scenarios, 6G can also offer a hybrid of these services to meet the extreme demand for new performance metrics in the Metaverse that are unlikely to be achieved by 5G and pre-5G networks. For instance, an application can leverage the combination of uMBB and uHPC, subject to the fulfillment of requirements of both services simultaneously. Similarly, uMTC and uHPC can be combined for a certain application if the needs of both services can be met simultaneously \cite{saad2019vision}. 

\subsection{Immersive Experiences over Wireless}
Immersive experiences are facilitated through virtual worlds, where the users experience a fully immersive environment, thus creating a meta-world that projects the real-world scenarios. State-of-the-art VR applications are embedded with a dynamic streaming framework that automatically adapts the bitrate of the XR content according to the network bandwidth and user’s geolocation \cite{petrov2022extended}. Although the 5G legacy has remained capable of providing rudimentary XR support, scaling the adoption of XR devices and applications over wireless networks requires new research and engineering directions. Recently, 5G and beyond has paced further in boosting the XR performance by introducing various heterogeneous improvements in not only XR-specific entities but also service-related advancements. Table \ref{bcrtable} shows several studies that investigated the role of state-of-the-art B5G/6G technologies for immersive experiences. 

\textbf{mmWave for Wirless VR:}
Towards this end, the 3GPP is involved in standardization activities for mmWave communications in B5G/6G systems\cite{zhang20196g}. In the meantime, academia is also exploring the potential of mmWave mobile networks to support wireless VR streaming over high bandwidth\cite{maier2022art}. The use of mmWave communications enables high transmission rates and low delay transmissions so that wireless VR applications can be seamlessly supported. In addition, the abundance of wireless bandwidth can further alleviate the scarcity of transmission rates through mmWave communications. Researchers generally consider 60 GHz mmWave wireless technologies since it is the only standardized mmWave wireless technology that has been identified so far based on the IEEE 802.11ad \cite{kim2016performance}. For example, authors in \cite{kim2017strategic} explore the potential of a 60GHz wireless channel to simultaneously process the large-scale multimedia content delivery in distributed VR platforms.

\textbf{MEC for Wireless VR:}
Another challenge in supporting VR experiences is that the videos must also remap pixels from a viewing sphere to a 2D viewport in order to distinguish their content form from traditional videos. The rendering process involves complex matrix mult-add operations and is extremely computationally intensive. Hence, the power consumption of virtual reality videos is significantly higher than other forms of video\cite{leng2019energy}. As a result, this will bring great challenges to the ubiquitous applications of wireless VR, as the continuous rendering of viewports will lead to exhaustion of battery and will eventually shorten the battery life. To address this challenge, various scenarios of MEC and caching have emerged as promising solutions to support wireless VR \cite{yeh2022perspectives}. The deployment of MEC and caching paradigms at the mobile network edge can enable some computation tasks to be handled by in-network computing capacity.
Additionally, some popular VR content and computational outputs can be cached to reduce the amount of repetitive computation and transmission. For example, in the article \cite{yang2019joint}, the theoretical architecture of a hybrid mobile computing system is presented concurrently that combines the processing abilities of cloud servers, MEC servers and end devices along with the caching capabilities of MEC servers. Similarly, according to \cite{yang2018communication}, VR tasks can be divided into several sub-components, and part of the sub-component can be cached using the mobile device's caching mechanism. 
\begin{figure}[t!]
\includegraphics[width= 0.5\textwidth]{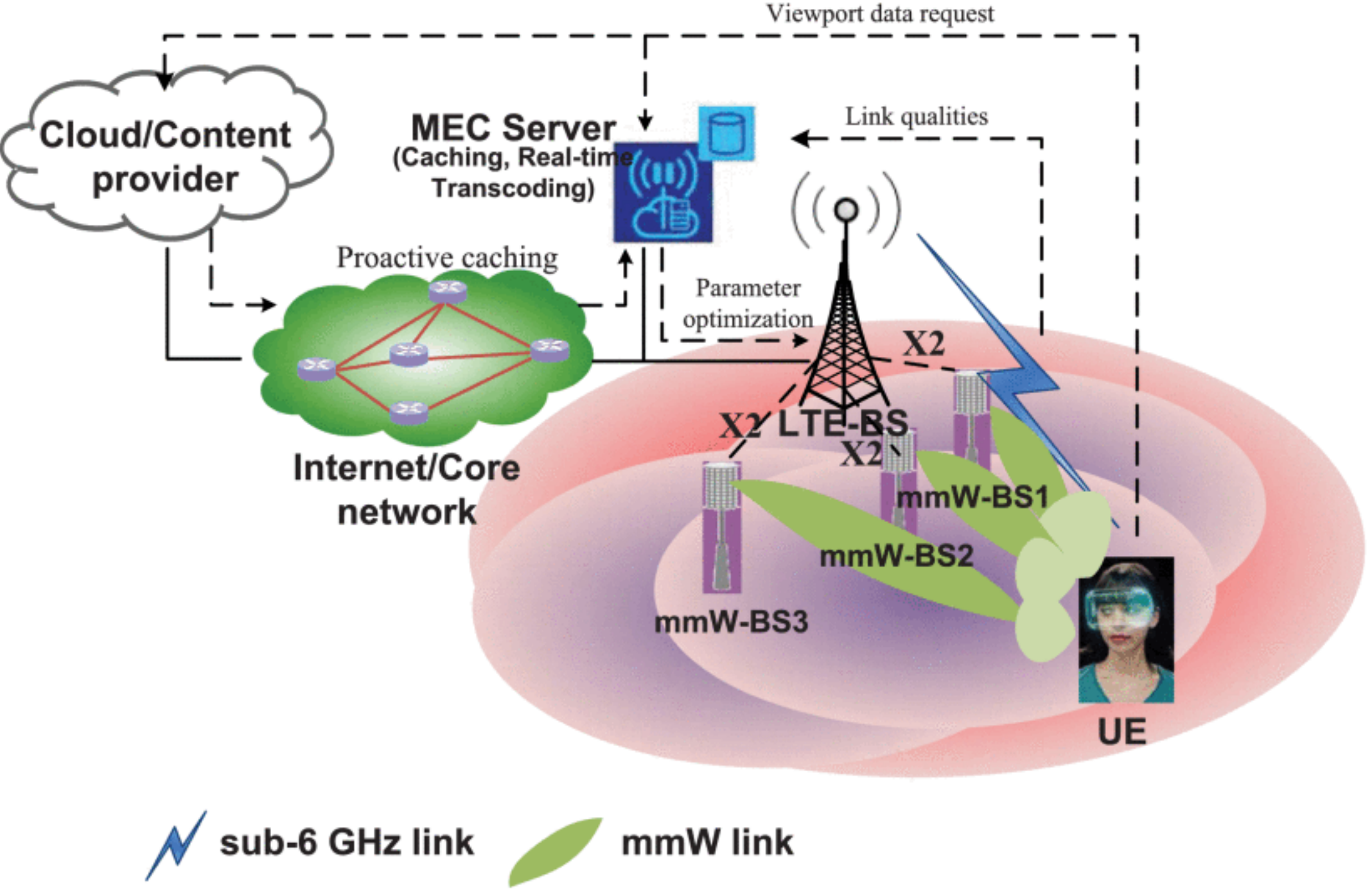}
\caption{ Transmission of a panoramic virtual reality video (PVRV) stream using mmWave and sub-6GHz link on a MEC framework\cite{liu2018mec}.}
\centering\label{fig8}
\end{figure}

\textbf{mmWave meets MEC for Transmission Efficiency in Wireless VR:}
Furthermore, several studies have indicated that introducing FOV into $360^\circ$ video will reduce up to 80\% of the bandwidth requirements compared to delivering $360^\circ$ video, hence lowering the overall necessary transmission data rate \cite{qian2016optimizing,mangiante2017vr}. For example, authors in \cite{sun2019communications} analyze the tradeoff between homogeneous and heterogeneous FOVs for a MEC-based mobile VR delivery model regarding computations and caching tasks. In addition, several methods have been proposed that take advantage of the FOV of users, for instance, introducing FOV prediction \cite{zhao2020optimal} and segregating different types of VR frames and multicasting them \cite{long2020optimal}. Based on the above discussion, it is apparent that more schemes can be developed if mmWave communication is combined with MEC-based FOV mechanisms. For instance, Liu et al. use mmWave and sub-6GHz link to transmit a panoramic virtual reality video (PVRV) stream by encoding the viewport and rendering the standby tiles according to different quality levels \cite{liu2018mec}, as shown in Figure \ref{fig8}. Similarly, Gputa et al. present A multiple-layer streaming scheme in which the base layer is encoded and transmitted via Wi-Fi \cite{gupta2019millimeter}. In contrast, a number of enhancement layers are transmitted via the mmWave access point. Therefore, a fusion of mmWave and MEC-based FOV schemes can reduce bandwidth requirements while reducing the transmission delay when compared with $360^\circ$ videos.

\textbf{Immersive fidelity with THz communication:}
Metaverse will use $360^\circ$ videos and MAR applications to bring immersiveness, as well as AI-driven applications with ubiquitous integration. Due to this, vast amounts of data will be exchanged; hence the demand for radio frequency bandwidth spectrum will surge rapidly. The existing mmWave technology cannot satisfy this demand. Consequently, switching to the terahertz and sub-terahertz bands would be imperative. THz communications are able to deliver ultra-high throughput and low latency by supporting data-intensive applications such as VR/XR over wireless personal area networks \cite{fantacci2021edge}. Several studies have leveraged the unique benefits of THz to be implemented with 6G for immersive experiences. In a study by Du et al., authors examined the problem of providing a high-quality immersive VR video service \cite{du2020mec}. In particular, the authors exploit the THz channel with MEC integration to optimize the immersive VR rendering, power management, and offloading viewports while satisfying the QoE constraints. Similarly, Chen et al., \cite{chen2019liquid} investigated an immersive application involving VR users transmitting $360^\circ$ images to a backend server. They formulated a problem to maximize the users' successful transmission probability by optimizing the downlink image transmission and the rotation of the surrounding images. Their proposed optimization problem involves a transfer learning approach consisting of liquid state machines to enhance the convergence speed by transferring the learned transmission into a new one. Despite the promising QoS delivery by the THz communication system, it still has significant propagation losses and water-molecule absorption losses due to high transmission frequency \cite{han2018propagation}. For this reason, THz has remained effective for indoor service delivery. For instance, Liu et al. propose a deep reinforcement learning (DRL)-based strategy to efficiently maintain the long-term QoE in THz-based immersive systems \cite{liu2021learning}. Specifically, authors have studied a joint use of THz and RIS to create a VR network in an indoor scenario with an aim to enhance downlink transmission. To achieve the aim, the authors utilize the current and historical viewpoints of VR users obtained from real VR datasets to train a gated recurrent unit (GRU) model for predicting the dynamic viewpoint preference of VR users. Table \ref{bcrtable} shows several studies that investigated the THz channels for immersive experiences.

\subsection{Holographic Telepresence in Metaverse}
Metaverse users would be able to see high-quality, three-dimensional digital representations of people using Holograms and without wearing HMD. Future networks (i.e., B5G/6G) will be able to support extremely low latencies and ultra-high data rates with high reliability in order to underpin the holographic communication that will facilitate Metaverse users to fully immerse in gadget-free communication\cite{maier2022art}. For example, a 6G network with 0.1ms of latency supported by an ultra-high bandwidth of Tbps can be leveraged for HT \cite{chowdhury20206g}. In order to evaluate HT applications, there should be seamless and quality connections between users, which requires an extremely low-latency data transmission rate and massive low latency. A number of factors affect the required data rates, such as the method used to construct a hologram, the type of display and the number of images that need to be synchronized. For example, a typical hologram based on the point cloud techniques requires 0.5–2 Gb/s, whereas a large-sized hologram may require up to a few Tb/s \cite{de2021survey}. The transmission of such holograms may be underpinned by using data compression techniques, but even then, holograms are likely to require vast amounts of bandwidth. Therefore, latency and reliability become two critical key performance indicators (KPIs) for HT. Within Metaverse, HT will be able to launch a number of human-type communications, including telesurgery, immersive education, teleoperations, and the Internet of Skills (IoS) \cite{park2020extreme}. Due to HT’s high sensitivity to latency, these applications have strict requirements for round-trip delay. Thus, it is important that the communication link for these applications be ultra-reliable to avoid packet losses which may lead to disastrous events. To overcome that, mobile B5G/6G networks have been considering URLLC as one of the new application scenarios to support the above applications \cite{wallace2021high}. It is expected that with 5G New Radio, the radio access networks (RANs) can achieve a 1 ms delay with 10-5-10-7 packet loss probability. This includes 0.5 ms uplink transmission and 0.5 ms downlink transmission \cite{li2020enabling}. In addition to that, 6G can provide URLLC with 1 Tbps of data rate and E2E delay of 0.1ms using THz frequency bands to support HT in Metaverse.

\begin{table*}[]
\centering
\caption{Summarising the role of state-of-the-art B5G/6G technologies in the context of Metaverse.}
\label{bcrtable}
\resizebox{\textwidth}{!}{\begin{tabular}{|c|c|c|c|}
\hline
\rowcolor{Gray1}{\color[HTML]{0E101A} \textbf{Authors \& Year}} &
  {\color[HTML]{0E101A} \textbf{Reference}} &
  \textbf{B5G/6G Services} &
  \textbf{Application scenario} \\ \hline
\begin{tabular}[c]{@{}c@{}}Mourtzis~\textit{et al}., \\ 2017 \end{tabular}  &
  {\color[HTML]{0E101A} \begin{tabular}[c]{@{}c@{}}\cite{mourtzis2017augmented}\end{tabular}} &
  HMDs, mobile edge, and cloud computing &
  {\color[HTML]{0E101A} \begin{tabular}[c]{@{}c@{}}AR-based tele-maintenance using a \\ cloud-based and serviced oriented system  \end{tabular}}
  \\ \hline

\rowcolor{Gray1}\begin{tabular}[c]{@{}c@{}}Cheng~\textit{et al}., \\ 2018 \end{tabular} &
  {\color[HTML]{0E101A} \begin{tabular}[c]{@{}c@{}}\cite{cheng2018industrial}\end{tabular}} &
  eMBB, mMTC, URLLC, and IIoT &
  {\color[HTML]{0E101A} \begin{tabular}[c]{@{}c@{}}AR/VR and digital twin based \\ industrial manufacturing  \end{tabular}}
  \\ \hline

\begin{tabular}[c]{@{}c@{}}Erol~\textit{et al}., \\ 2018 \end{tabular} &
  {\color[HTML]{0E101A} \begin{tabular}[c]{@{}c@{}}\cite{erol2018caching}\end{tabular}} &
  edge caching, edge mining, computational offloading& 
 AR/VR access pattern in virtual tourism \\ \hline

\rowcolor{Gray1}\begin{tabular}[c]{@{}c@{}}Ren~\textit{et al}., \\ 2020 \end{tabular} &
  {\color[HTML]{0E101A} \begin{tabular}[c]{@{}c@{}}\cite{ren2020edge} \end{tabular}} &
   {\color[HTML]{0E101A} \begin{tabular}[c]{@{}c@{}}D2D communication, cloud computing, \\ on-device computing, and computational offloading \end{tabular}}
   & 
  {\color[HTML]{0E101A} \begin{tabular}[c]{@{}c@{}}Edge-assisted multi-user cooperative environment \\ for mobile web AR applications \end{tabular}}
 \\ \hline
 
\begin{tabular}[c]{@{}c@{}}Pengnoo~\textit{et al}., \\ 2020 \end{tabular} &
{\color[HTML]{0E101A} \begin{tabular}[c]{@{}c@{}}\cite{pengnoo2020digital} \end{tabular}} &
digital twins, THz links, and metasurface reflectors &
{\color[HTML]{0E101A} \begin{tabular}[c]{@{}c@{}}Improving transmission links for potential  \\ data-intensive applications in Metaverse\end{tabular}}
\\ \hline
 
\rowcolor{Gray1}\begin{tabular}[c]{@{}c@{}}Zhou~\textit{et al}., \\ 2021 \end{tabular} &
  {\color[HTML]{0E101A} \begin{tabular}[c]{@{}c@{}}\cite{zhou20215g} \end{tabular}} &
  MEC, 5G antennas, computational handoff, RNI, and gNB/ng-eNB &
  {\color[HTML]{0E101A} \begin{tabular}[c]{@{}c@{}}Improving QoE for MAR based gaming such as \\ collaborative assembly of a virtual object \end{tabular}}
  \\ \hline
  
  \begin{tabular}[c]{@{}c@{}}Liu~\textit{et al}., \\ 2021\end{tabular} &
  {\color[HTML]{0E101A} \begin{tabular}[c]{@{}c@{}}\cite{liu2021learning} \end{tabular}} &
  MEC, THz network, RIS, and federated learning & 
  {\color[HTML]{0E101A} \begin{tabular}[c]{@{}c@{}}Empowering immersive VR experiences using \\ prediction, rendering, and transmission of $360^\circ$ content \end{tabular}}
 \\ \hline
  
  \rowcolor{Gray1}\begin{tabular}[c]{@{}c@{}}Fantacci~\textit{et al}., \\ 2021 \end{tabular} &
  {\color[HTML]{0E101A} \begin{tabular}[c]{@{}c@{}}\cite{fantacci2021edge} \end{tabular}} &
  HRLLC, THz channels, SNC, and edge computing &
  {\color[HTML]{0E101A} \begin{tabular}[c]{@{}c@{}}Improving QoS for edge-assisted VR experiences \\ by minimizing the E2E transmission delay \end{tabular}}
  \\ \hline
  
  \begin{tabular}[c]{@{}c@{}}Ng~\textit{et al}., \\ 2021\end{tabular} &
  {\color[HTML]{0E101A} \begin{tabular}[c]{@{}c@{}}\cite{ng2021unified}\end{tabular}} &
  edge intelligence, unified resource allocation, and VSPs & 
  {\color[HTML]{0E101A} \begin{tabular}[c]{@{}c@{}}Improving QoS in educational applications of Metaverse \\ by optimising resource allocation on users' demands  \end{tabular}} \\ \hline
  
  \rowcolor{Gray1}\begin{tabular}[c]{@{}c@{}}Ren~\textit{et al}., \\ 2022 \end{tabular} &
  {\color[HTML]{0E101A} \begin{tabular}[c]{@{}c@{}}\cite{ren2022distributed}\end{tabular}} &
  edge node localisation, service migration, load balancing, distributed edge AI &
  {\color[HTML]{0E101A} \begin{tabular}[c]{@{}c@{}}Orchestration of distributed edge services\\ for location-based mobile AR systems  \end{tabular}}
  \\ \hline
  
  \begin{tabular}[c]{@{}c@{}}Zhang~\textit{et al}., \\ 2022 \end{tabular} &
  {\color[HTML]{0E101A} \begin{tabular}[c]{@{}c@{}}\cite{zhang2022sear}\end{tabular}} &
  edge caching, computational offloading, decentralised computing&
  {\color[HTML]{0E101A} \begin{tabular}[c]{@{}c@{}} Scaling object recognition services for \\ seamless mobile AR experiences \end{tabular}}
  \\ \hline

\rowcolor{Gray1}\begin{tabular}[c]{@{}c@{}}Ghoshal~\textit{et al}., \\ 2022 \end{tabular} &
  {\color[HTML]{0E101A} \begin{tabular}[c]{@{}c@{}}\cite{ghoshal2022can}\end{tabular}} &
  mmWave, cloud computing, uplink data offloading& 
  {\color[HTML]{0E101A} \begin{tabular}[c]{@{}c@{}}Realising cloud-based multi-user AR experiences \\ with 5G mmWave \end{tabular}} \\ \hline

   \begin{tabular}[c]{@{}c@{}}Van~\textit{et al}., \\ 2022 \end{tabular} &
  {\color[HTML]{0E101A} \begin{tabular}[c]{@{}c@{}}\cite{van2022edge}\end{tabular}} &
  MEC, URLLC, digital twins, and edge caching&
  {\color[HTML]{0E101A} \begin{tabular}[c]{@{}c@{}} Improving QoE of digital twins in Metaverse by jointly \\  optimising the computing, communication, and storage resources \end{tabular}}
  \\ \hline

\rowcolor{Gray1}\begin{tabular}[c]{@{}c@{}}Bhattacharya~\textit{et al}., \\ 2022 \end{tabular} &
  {\color[HTML]{0E101A} \begin{tabular}[c]{@{}c@{}}\cite{bhattacharya2022Metaverse}\end{tabular}} &
  TI, URLLC, H2M control, blockchain, xAI, and decentralised computing &
  {\color[HTML]{0E101A} \begin{tabular}[c]{@{}c@{}}Improving quality and precision of tele-surgeries \\ for patients, virtual hospitals, and doctors in Metaverse  \end{tabular}}
  \\ \hline

\begin{tabular}[c]{@{}c@{}}Kang~\textit{et al}., \\ 2022 \end{tabular} &
  {\color[HTML]{0E101A} \begin{tabular}[c]{@{}c@{}}\cite{kang2022blockchain} \end{tabular}} &
  MEC, federated learning, IIoT, blockchain, and AoI & 
  {\color[HTML]{0E101A} \begin{tabular}[c]{@{}c@{}}Empowering immersive Metaverse experiences through \\ user-defined privacy-preserving incentive schemes \end{tabular}}
 \\ \hline

\end{tabular}}
\end{table*}

\

\begin{figure*}[ht]
\includegraphics[width= 1.0\textwidth]{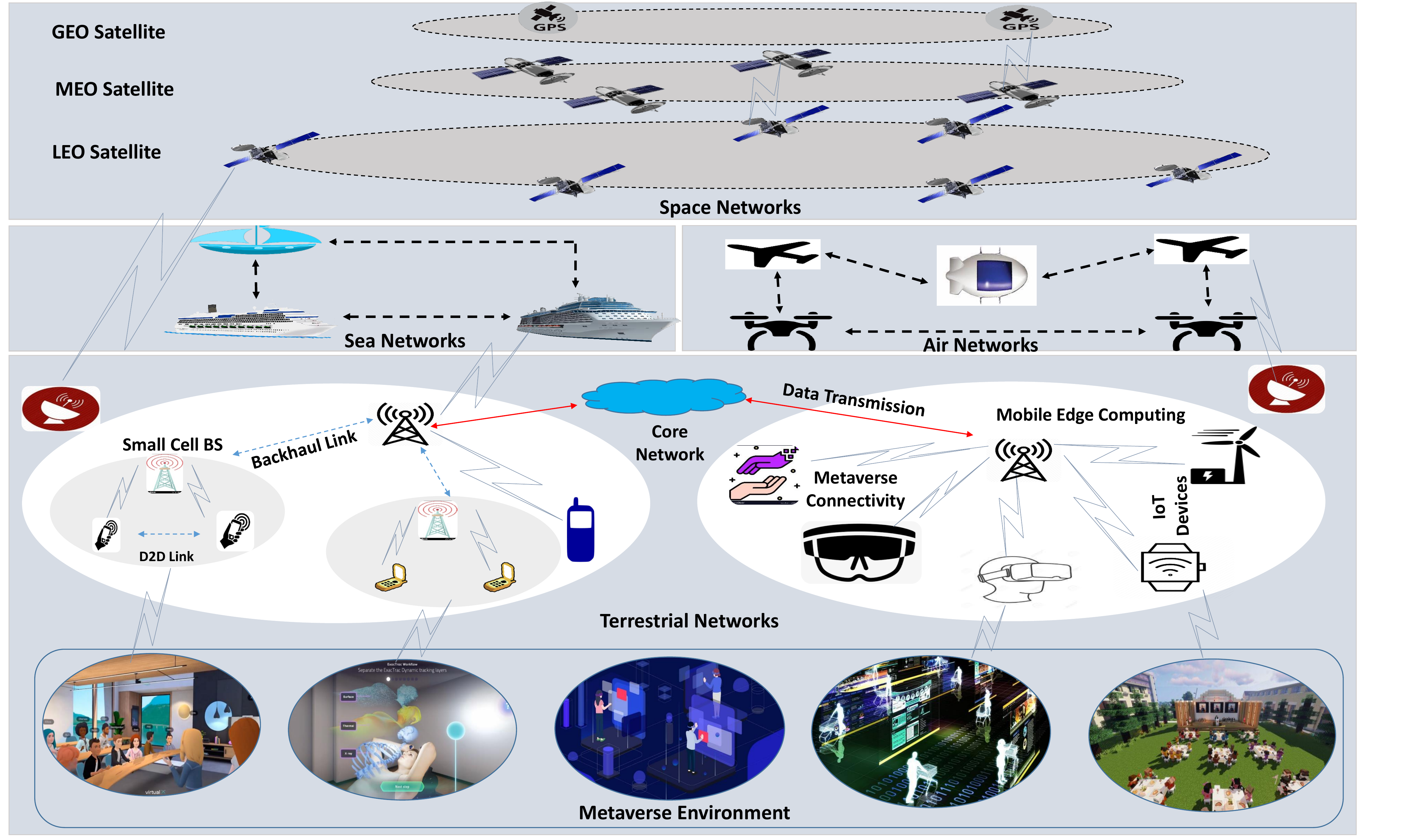}
\caption{An illustration of a large-scale SAGSIN in Realizing Metaverse.}
\centering\label{Fig9}
\end{figure*}

\section{Nexus of AI and 6G for Metaverse} 
This section outlines the interplay of B5G/6G and AI towards Metaverse. In subsections V-A, we discuss the role of 6G in realising the pervasive and ubiquitous AI services that are discussed in section III for Metaverse environments. Subsection V-B explains the role of AI in making autonomous 6G network infrastructures for Metaverse, followed by the joint role of AI and 6G for tactile and immersive experiences in subsections V-C. 
\subsection{6G-driven Pervasive AI for Metaverse}
Metaverse is envisioned to be extending people’s life by allowing them to live, work, and play seamlessly for unlimited durations. This can only be realized if the Metaverse has a strong relationship with “ubiquitous connectivity with intelligence,” i.e., B5G/6G, AI, and space-air-ground-sea integrated network (SAGSIN) as shown in Figure \ref{Fig9}. In essence, 6G will support pervasive intelligence by reducing the potential latency of virtual experiences and will allow developers to develop better virtual experiences in the Metaverse where millions of concurrent users can share an immersive 3D space with each other. In such cases, the location of sender and receiver might be remote; thus, the challenge is to support dense connections (e.g., $10^7$ users/$km^2$), where multiple users can transmit and receive $360^\circ$ video content or real-time human-like holographs in parallel \cite{chude2022enabling,tang2022roadmap}. Such a sheer amount of data fidelity will require ultra-high bandwidth with heterogeneous hardware for hosting AI services discussed in section III, requiring extremely low-latency communications for transferring model parameters or performing AI inferences.

Since 5G’s rollout, the first set of promising technologies along with architectural evaluation were polar codes, massive MIMO, mmWave, and MEC. These unique 5G services were able to support AR/VR applications by providing ubiquitous connectivity. AR/VR users leveraged 5G-eMBB service to achieve $<$100ms latency with a throughput of 100-200Mbps \cite{chakrabarti2021deep}. Several experiments were carried out further to evaluate 5G’s performance in realistic mobile environments. For example, the 5G-PPP (The 5G Infrastructure Public Private Partnership, 2018)\footnote{https://5g-ppp.eu/5g-trials-2/} consortium reported a data rate of approximately 3Gbps having 2ms of latency on their Bari-Matera installation in Italy\footnote{https://www.fastweb.it/fastweb-plus/digital-magazine/5g-matera-fastweb/}. This experiment was carried out with the combination of 5G services with LTE technology\footnote{http://www.barimatera5g.it/}. Another experiment provides data from around 180 trials and investigations conducted by the European 5G Observatory. In this scenario, the trials were conducted for a number of weeks in urban areas of Finland, therefore possibly with a reasonable density of users, and results suggest that the most realistic data rate was approximately ranged from 700 Mbps to 1 Gbps. According to the final experiment results, users' devices realized 1 to 4.5 Gbps in average data rates, with less than 5 ms latency. It has widely been acknowledged that AR/VR applications require latency lower than 5ms, with average throughput between 100 Mbps to a few Gbps, and reliability of at least 99\%, meaning that 5G performance is sufficient to support these needs \cite{torres2020immersive}.

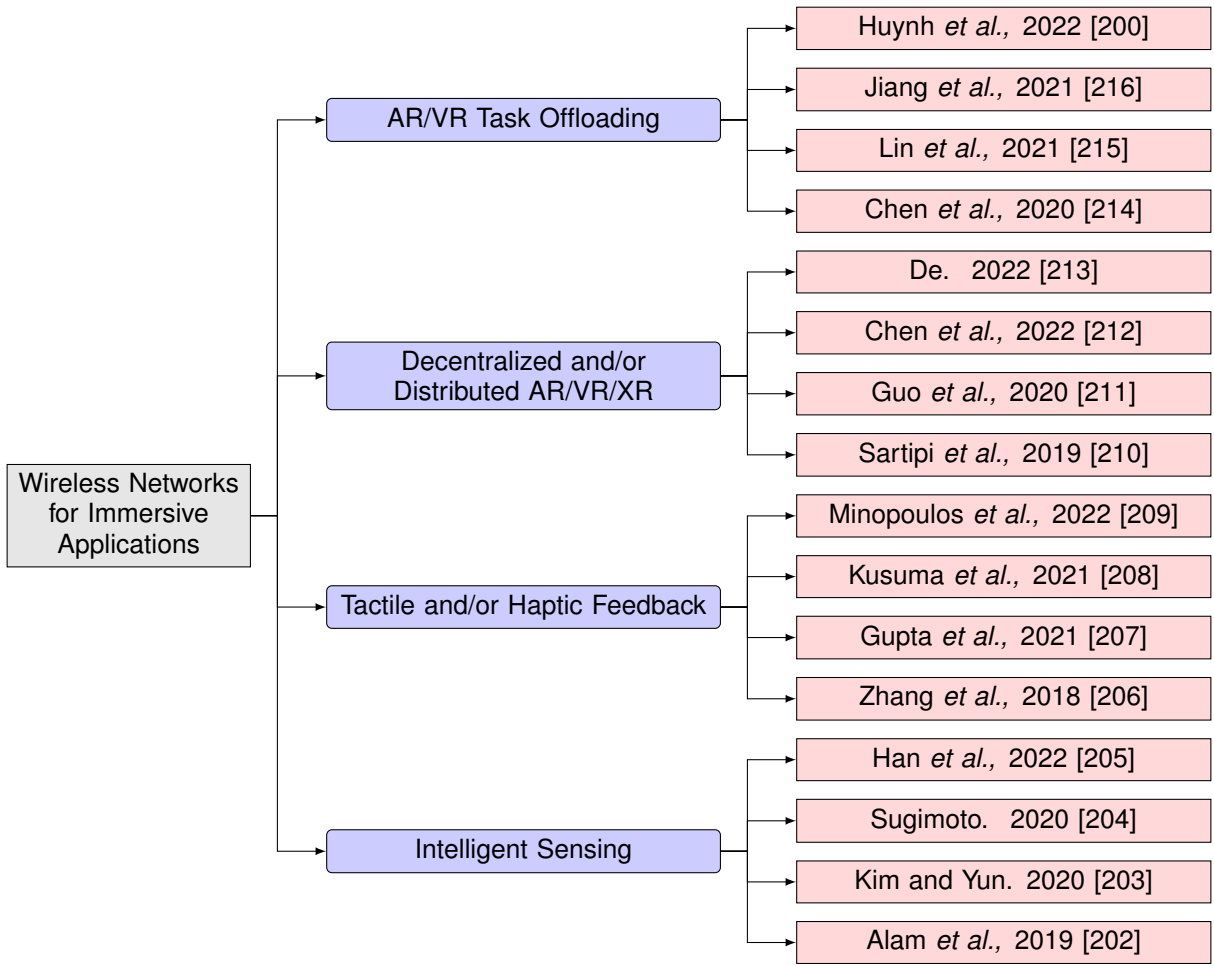
\begin{figure*}
    \centering

\tikzset{
    basic/.style  = {draw, text width=3cm, align=center,fill=gray!20, font=\sffamily, rectangle},
    root/.style   = {basic, rounded corners=2pt, thin, align=center, fill=green!30},
    onode/.style = {basic, thin, rounded corners=2pt, align=center, fill=green!60,text width=3cm,},
    tnode/.style = {basic, thin, align=left, fill=pink!60, text width=15em, align=center},
    xnode/.style = {basic, thin, rounded corners=2pt, align=center, fill=blue!20,text width=5cm,},
    wnode/.style = {basic, thin, align=left, fill=pink!10!blue!80!red!10, text width=6.5em},
    edge from parent/.style={draw=black, edge from parent fork right}

}

\begin{forest} for tree={
    grow=east,
    growth parent anchor=west,
    parent anchor=east,
    child anchor=west,
    edge path={\noexpand\path[\forestoption{edge},->, >={latex}] 
         (!u.parent anchor) -- +(10pt,0pt) |-  (.child anchor) 
         \forestoption{edge label};}
}
[Wireless Networks for Immersive Applications, basic,  l sep=10mm,
    [Intelligent Sensing, xnode,  l sep=10mm,
        [Alam \textit{et al.,}\, 2019 \cite{alam2015intelligent}, tnode]
        [Kim and Yun.\, 2020 \cite{kim2020motion}, tnode]
        [Sugimoto. \, 2020 \cite{sugimoto2022cloud}, tnode]
        [Han \textit{et al.,}\, 2022 \cite{han2022dynamic}, tnode] ]
    [Tactile and/or Haptic Feedback, xnode,  l sep=10mm,
        [Zhang \textit{et al.,}\, 2018 \cite{zhang2018towards}, tnode]
        [Gupta \textit{et al.,}\, 2021 \cite{gupta2019tactile}, tnode]
        [Kusuma \textit{et al.,}\, 2021 \cite{kusuma2021enabling}, tnode]
        [Minopoulos \textit{et al.,}\, 2022 \cite{minopoulos2021efficient}, tnode] ]
    [Decentralized and/or Distributed AR/VR/XR, xnode,  l sep=10mm,
        [Sartipi \textit{et al.,}\, 2019 \cite{sartipi2019decentralized}, tnode]
        [Guo \textit{et al.,}\, 2020 \cite{guo2020adaptive}, tnode]
        [Chen \textit{et al.,}\, 2022 \cite{chen2020federated}, tnode] 
        [De. \, 2022 \cite{de2022fedlens}, tnode]]
    [AR/VR Task Offloading, xnode,  l sep=10mm,
        [Chen \textit{et al.,}\, 2020 \cite{chen2020joint}, tnode]
        [Lin \textit{et al.,}\, 2021 \cite{lin2021task}, tnode]
        [Jiang \textit{et al.,}\, 2021 \cite{jiang2021reliable}, tnode]
        [Huynh \textit{et al.,}\, 2022 \cite{van2022edge}, tnode]
         ] ]
\end{forest}

    \caption{Taxonomy for the classification of literature on 5G/6G Wireless-empowered Immersive AR/VR Applications.}
    \label{fig:6G-taxonomy}
\end{figure*}

Metaverse is supposed to be accessible from anywhere at any time; thus, it is presumed that 5G networks will fail to provide coverage in some parts of the globe, such as remote areas, oceans, mountains, deep forests, and airspace. This shortcoming of 5G can be overcome by 6G, which is expected to provide ubiquitous global coverage as shown in Figure \ref{Fig9}. As part of 6G systems, data collection, transportation, and utilization will take place in real-time and anywhere at any given time, hence, catalyzing immersive applications and services for Metaverse. 6G will particularly emphasize ubiquitous AI, bringing artificial general intelligence into every aspect of Metaverse through a hyper-flexible architecture. On top of that, a 6G network will be well suited for several Metaverse applications involving deep learning because it can generate a great deal of data and perform computation and storage at the network edge \cite{loven2019edgeai}.
In summary, 6G networks can not only host personalized AI services but also cater to the needs of dense user connectivity with low-latency-high-reliability communications. For instance, a case study by Adeogun et al. entails a framework involving a short-range wireless isochronous real time (WIRT) environment for a 6G communication system. Authors developed a WIRT in-X sub-network with roundtrip communication latency lesser than 0.1ms with an outage probability $<10^{-6}$ \cite{adeogun2020towards}. A multi-GHz spectrum for enhancing spatial service availability is considered part of a dense IoT scenario with up to two devices per $m^2$. Based on their results, it can be observed that cycle times are 10x shorter than the latency target achieved via 5G radio technology, i.e., $<$0.1ms \cite{adeogun2020towards}. Therefore, Metaverse services involving novel communication modalities such as A2A, A2H, and H2A can leverage great benefits from 6G. We present a clear taxonomy of state-of-the-art for 5G/6G wireless-empowered immersive experiences in Figure \ref{fig:6G-taxonomy}. 

\subsection{AI-led Autonomous 6G Networks for Metaverse}

The success of the Metaverse experience will highly rely on the seamless and sustainable integration of physical space into virtual 3D space. Modern technologies (i.e., XR, AI, and 6G) can contribute to the interconnected Metaverse by ensuring interoperable coherence between the Metaverse and the physical world (e.g., physical objects in the real world are compatible with virtual objects in the Metaverse) \cite{khan2022Metaverse}. In this regard, a minor disconnect between Metaverse and the physical world can cause only a mere disturbance to the users; however, as the prevalence and utility of Metaverse increases, such a disconnect in immersion may leave a long-lasting impact on the life-critical applications of Metaverse. Therefore, a communication system for Metaverse must be capable of providing a stress-free and extremely reliable experience for sheer users. 

\subsubsection{Self-X Networks}
Based on the above-discussed Metaverse requirements in subsection IV-A, it is critical for 6G networks to have an intelligent and autonomous adjustment capability in order to meet various performance requirements, such as full coverage sensing, ultra-low latency, and ultra-high reliability, and persistent security \cite{benzaid2020ai}. At present, there is difficulty in dynamically adapting to the continual changes in user needs and network environments that alter the operating paradigms of today's networks, which mainly rely on rule-based algorithms as their core. A further constraint is the inability to effectively accumulate the QoE of network operations, which in turn prevents continuous improvement in the capabilities of the network \cite{rizwan2021zero}. In other words, it is not currently possible for networks to self-evolve under the existing operating procedures. Thereby, whenever there is a need to upgrade or improve something, much human expertise and effort are needed. As 6G networks grow in geography and complexity, manual intervention is unacceptable and unreliable for the networks to operate at such a scale. Therefore, to overcome this hurdle, european telecommunications standards institute (ETSI) recently initiated 2 groups, i.e., zero-touch service management (ZSM)\footnote{https://www.etsi.org/committee/zsm} and experiential networked intelligence (ENI)\footnote{https://www.etsi.org/committee/ENI}, having focused on the use of ML and DL to manage and orchestrate the resources of a network in a fully automated way. The key objective of ZSM and ENI is to add the component of “intelligence” into every block of future networks in order to realize intelligent and self-evolution abilities within the 6G-and-beyond networks \cite{andrus2019zero}. 

\subsubsection{AI for Networks}
The inception of Metaverse will exceptionally pose new challenges for 6G networks that will require support from AI, particularly deep reinforcement learning, and deep federated learning as shown in Figure \ref{Fig10}. In the traditional sense, the Metaverse is not like a digital twin, which is a virtual representation serving as the equivalent to a physical object \cite{yang2018digital}. In essence, Metaverse will offer possibilities for scenario creation that differ from reality and allow them to be run at their own time and with their own rules, requiring dynamic support from 6G and AI nexus. As such, several researchers have already explored the use of AI for supporting 6G networks in challenging environments. For example, the dutch double auction mechanism and DRL were combined in \cite{xu2021wireless} to improve transaction efficiency in the Metaverse. The ability to make intelligent decisions based on RL is a key enabler for the development of 6G networks for Metaverse, especially those with strict QoS requirements. Metaverse services can benefit from RL optimization techniques if they are well designed because they enable the ability of self-healing, self-optimizing, and self-organizing for the 6G networks. For instance, Tariq et al. propose a DRL-driven approach for proactive resource management and intelligent service provisioning in 6G networks for a digital-twin application \cite{tariq2022toward}. However, this approach can suffer from a significant deficiency in the performance of AI models. Particularly, for a highly distributed Metaverse application where users are only supposed to interact continuously for the purpose of AI model training without accessing the other user’s data due to strict security restrictions as shown in Figure \ref{Fig10}. To retain the performance of such AI models, recently Kang et al. proposed an architecture for a decentralized Metaverse application based on federated learning integrated with blockchain \cite{kang2022blockchain}. Based on their architecture, the local AI model can be optimized by horizontal cooperation of multiple end nodes (edge devices), which eventually retain the performance of AI models over time. Hence, the use of AI to optimize the network architecture and improve the performance of 6G networks is therefore proving to be a promising means of enhancing the Metaverse experience. In Figure \ref{fig:AI-taxonomy}, we present a clear taxonomy of state-of-the-art for AI-driven immersive experiences over wireless.

\subsection{AI and 6G for Tactile Internet in Metaverse}
In the Metaverse, the tactile Internet will dominate haptic communication for mission-critical virtual services with touch and actuation in real-time while facilitating timeliness of information delivery for intelligent and interconnected meta-worlds. TI is expected to underpin human-to-human and human-to-machine interactions in a way that the time duration between actions and feedback should be much lower than the physical world reaction times \cite{al2018experimental}. For this reason, it is necessary that the TI has a 1ms delay due to the nature of the haptic signals and how humans perceive them. There is, however, a limit to the latency that can be supported by pre-5G networks, which is approximately 25ms \cite{polachan2022assessing}. Therefore, B5G/6G networks are anticipated to outperform previous generations in terms of performance by providing a sustainable infrastructure that will ultimately realize the TI for immersive services. 5G can aid in TI realization for Metaverse by addressing the key communication and computation challenges leveraging the following 3 important enabling technologies: edge computing, network functions virtualization (NFV), and software-defined networking (SDN) \cite{aijaz2018toward}. These enabling technologies can be combined in order to provide sophisticated solutions for the network architecture. For instance, researchers in \cite{baktir2017can,rafique2020complementing} explore how SDN can be leveraged efficiently for the edge computing based 5G networks. In another study by Aijaz et al. \cite{aijaz2016realizing}, authors investigate how edge computing, SDN, and NFV can be utilized to create a novel, sustainable and general-purpose 5G architecture to realize TI applications. In essence, their approach is motivated by the virtualization concept of NFVs, which can dynamically instantiate an end-to-end 5G network according to the user’s applications’ requirements. Therefore, it can be observed that bringing computing capabilities closer to the user equipment and using SDN to control it centrally (i.e., limiting intermediate nodes) can foster latency reduction through edge computing and SDN \cite{rafique2020complementing,sharma2020toward}.
\begin{figure}[!t]
\includegraphics[width=0.5 \textwidth]{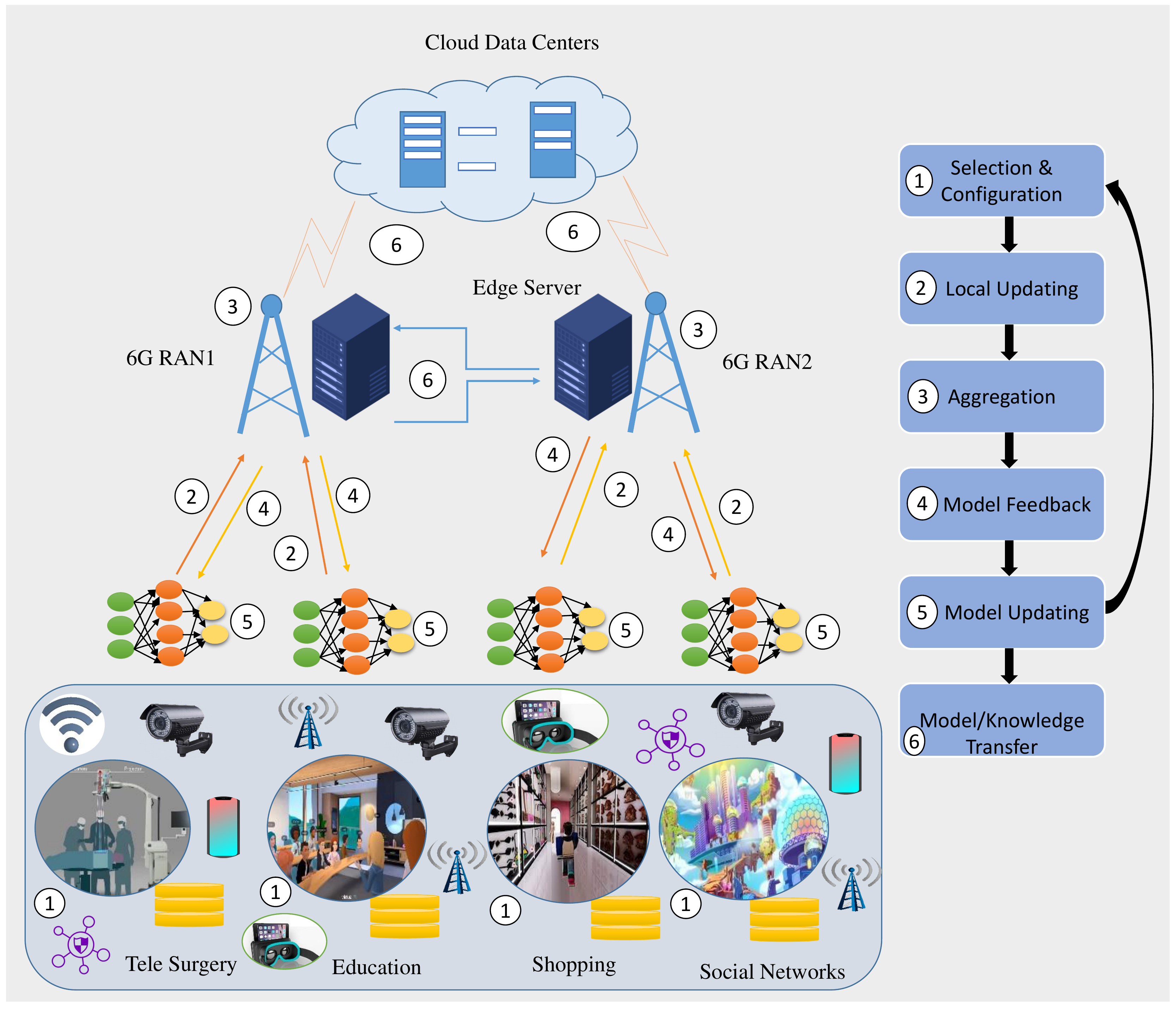}
\caption{FL architecture and distributed training procedure in  Metaverse.}
\centering\label{Fig10}
\end{figure}
It is also important to stress that the TI should not be limited to the context of 5G, as discussed above in relation to the Tactile Internet and 5G, but that it should take a broader position than that. Many research efforts are underway that aim to achieve the TI over other technologies (e.g., wireless communication with sub-GHz technology, wireless local area network (WLAN) and wireless body area network (WBAN), and combinations of technologies similar to these). Despite 5G's potential to enable URLLC applications in general, as mentioned in section IV-A, its design objectives were not customized to the TI’s specific features and requirements. In response to the shortcomings of 5G, both industry and academia have gathered a great deal of interest in next-generation 6G systems, which are expected to support a variety of immersive applications from multi-sensory communication to XR \cite{zhang2020challenges}. Apart from that, TI lays out a sheer amount of opportunities to leverage AI, particularly deep learning to segregate the perception of humans from the overwhelming transmission delays which are usually experienced in wide area networks (WANs) \cite{maier2019towards}. AI can certainly play a key role in fulfilling the 1ms latency requirements using these networks. Although communication-based enabling technologies can overcome the end-to-end latency challenge, their efficiency can not go beyond the scope of 150km due to their dependency on the speed of light. Therefore, deep learning-based predictions can be widely utilized to achieve 1ms latency by ruling out the 150km constraint on these networking technologies \cite{li2020enabling}.

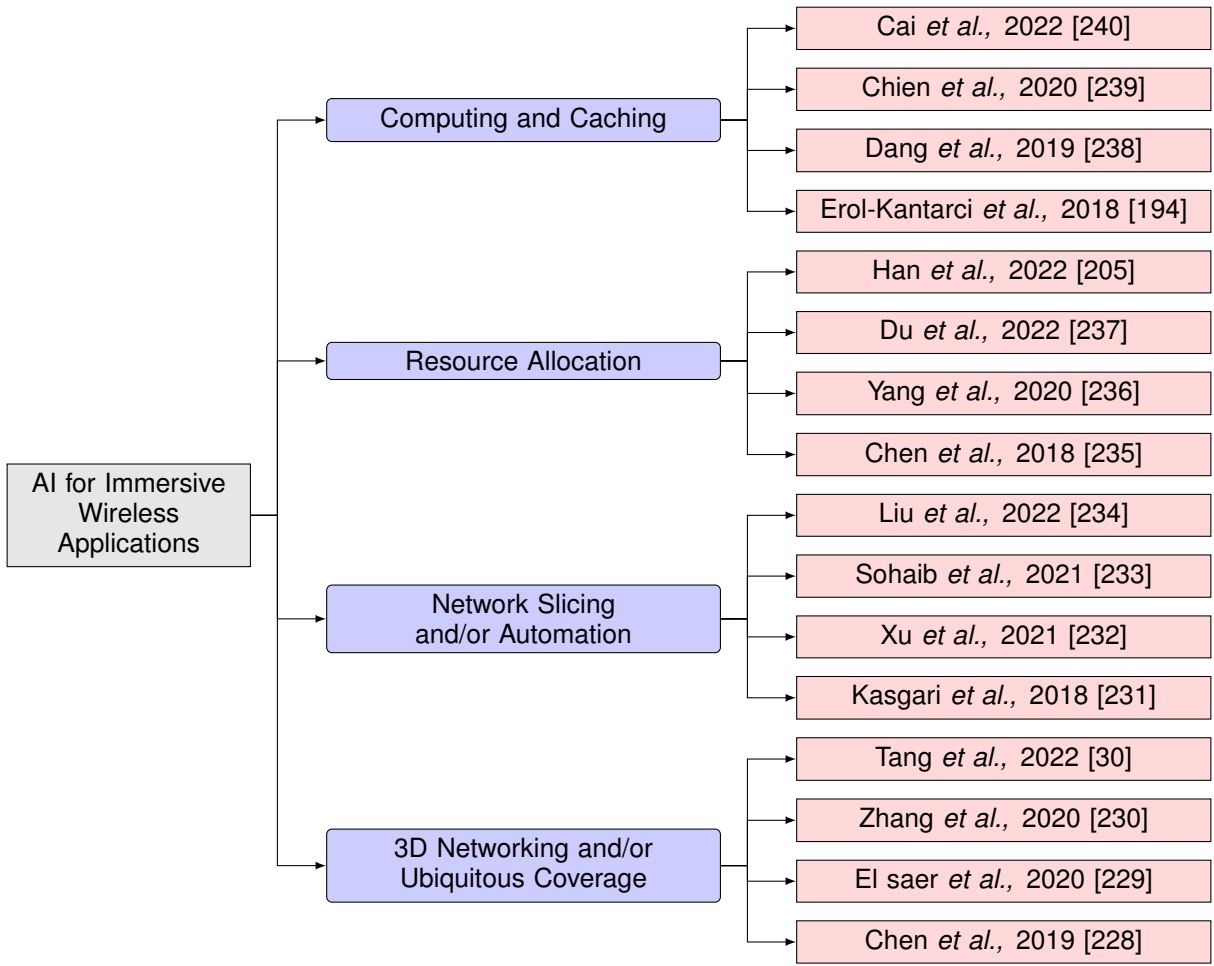
\begin{figure*}
    \centering

\tikzset{
    basic/.style  = {draw, text width=3cm, align=center,fill=gray!20, font=\sffamily, rectangle},
    root/.style   = {basic, rounded corners=2pt, thin, align=center, fill=green!30},
    onode/.style = {basic, thin, rounded corners=2pt, align=center, fill=green!60,text width=3cm,},
    tnode/.style = {basic, thin, align=left, fill=pink!60, text width=15em, align=center},
    xnode/.style = {basic, thin, rounded corners=2pt, align=center, fill=blue!20,text width=5cm,},
    wnode/.style = {basic, thin, align=left, fill=pink!10!blue!80!red!10, text width=6.5em},
    edge from parent/.style={draw=black, edge from parent fork right}

}

\begin{forest} for tree={
    grow=east,
    growth parent anchor=west,
    parent anchor=east,
    child anchor=west,
    edge path={\noexpand\path[\forestoption{edge},->, >={latex}] 
         (!u.parent anchor) -- +(10pt,0pt) |-  (.child anchor) 
         \forestoption{edge label};}
}
[AI for Immersive Wireless Applications, basic,  l sep=10mm,
    [3D Networking and/or Ubiquitous Coverage, xnode,  l sep=10mm,
        [Chen \textit{et al.,}\, 2019 \cite{chen2019deep}, tnode]
        [El saer \textit{et al.,}\, 2020 \cite{el20203d}, tnode]
        [Zhang \textit{et al.,}\, 2020 \cite{zhang2021method}, tnode]
        [Tang \textit{et al.,}\, 2022 \cite{tang2022roadmap}, tnode] ]
    [Network Slicing and/or Automation, xnode,  l sep=10mm,
        [Kasgari \textit{et al.,}\, 2018 \cite{kasgari2018stochastic}, tnode]
        [Xu \textit{et al.,}\, 2021 \cite{xu2021electrical}, tnode]
        [Sohaib \textit{et al.,}\, 2021 \cite{sohaib2021network}, tnode]
        [Liu \textit{et al.,}\, 2022 \cite{liu2022slicing4meta}, tnode] ]
    [Resource Allocation, xnode,  l sep=10mm,
        [Chen \textit{et al.,}\, 2018 \cite{chen2018virtual}, tnode]
        [Yang \textit{et al.,}\, 2020 \cite{yang2020resource}, tnode]
        [Du \textit{et al.,}\, 2022 \cite{du2022attention}, tnode] 
        [Han \textit{et al.,}\, 2022 \cite{han2022dynamic}, tnode] ]
    [Computing and Caching, xnode,  l sep=10mm,
        [Erol-Kantarci \textit{et al.,}\, 2018 \cite{erol2018caching}, tnode]
        [Dang \textit{et al.,}\, 2019 \cite{dang2019joint}, tnode]
        [Chien \textit{et al.,}\, 2020 \cite{chien2020q}, tnode]
        [Cai \textit{et al.,}\, 2022 \cite{cai2022joint}, tnode]
         ] ]
\end{forest}

    \caption{Taxonomy for the classification of literature on AI-empowered Immersive Wireless Experiences.}
    \label{fig:AI-taxonomy}
\end{figure*}

\section{Sustainable Metaverse} 
This section outlines the sustainability issues in the Metaverse with respect to virtual resources, virtual travel, digital twins, psychological barriers, and social sustainability, respectively. The issue of sustainability is at the helm of affairs when discussed business strategies concerning climate change. It is inevitable for the organizations to not think of a sustainable and efficient way for conducting business to make a positive impact on climate change. Although the concept of sustainability is not new, however, it needs to be evolved with respect to the innovation, changing times, and work style, accordingly.

With the announcement of Metaverse, the concerns about carbon emissions and energy consumption were also raised as the commerce of Metaverse is solely based on cryptocurrencies. However, it has been highlighted by several studies that the Metaverse holds potential for reducing carbon reductions through virtual interactions, replacing physical with digital goods, and digital twins, respectively. Furthermore, it has also been suggested that Metaverse, through immersive experience, can help overcome behavioral hurdles concerning climate change. 

Most of the studies use sustainability in the context of environment, however, sustainability should also be addressed from social point of view. Metaverse can help in enhancing social sustainability by bringing equity, inclusiveness, and accessibility on the table. It is the responsibility of the businesses and all stackholders involved to develop strategies and utilize Metaverse for improving both the environmental and social sustainability, respectively. This section briefly discusses the correlation between sustainability and Metaverse with the use of Digital-Twins and Industry 5.0. The topics covered in this subsection are summarized in figure \ref{FigS-D}.

\begin{figure*}[ht]
\includegraphics[width= \linewidth]{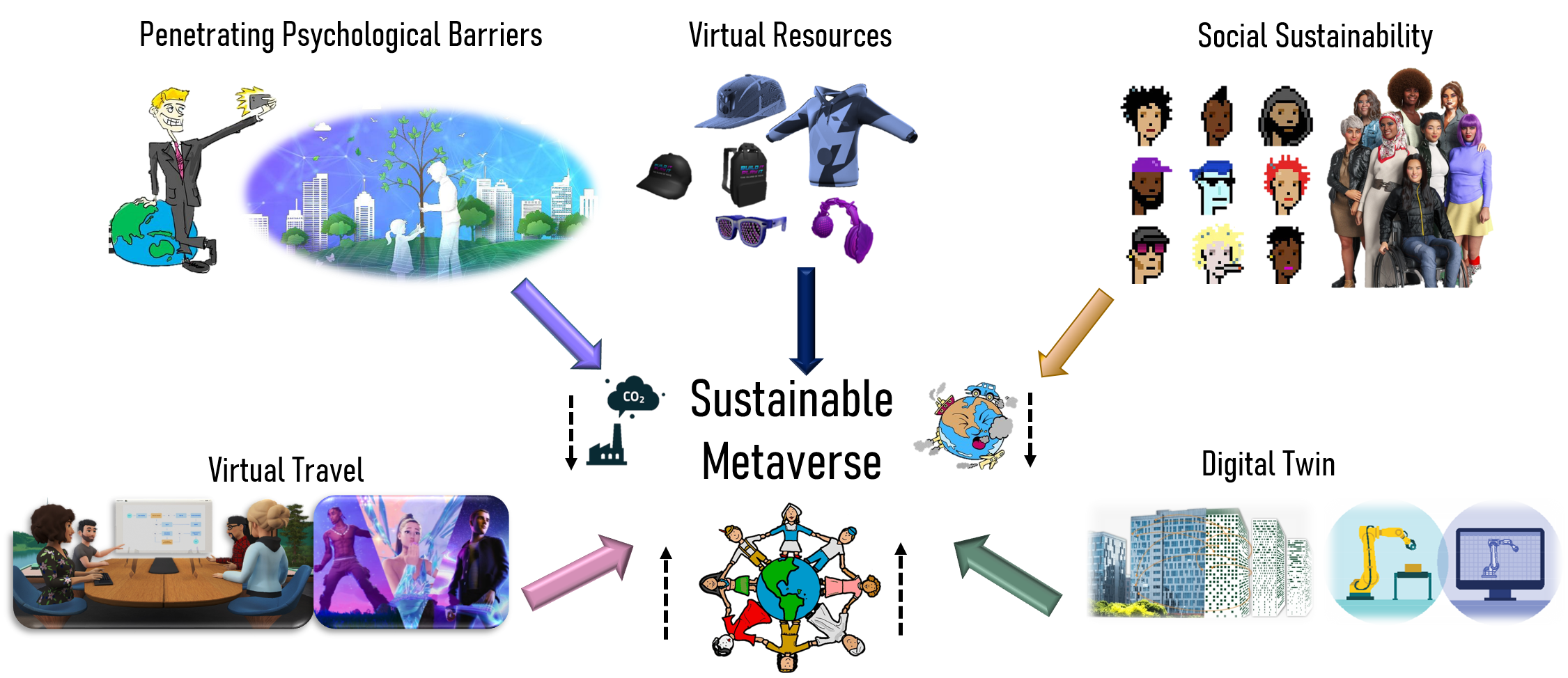}
\caption{Summarization of Sustainable Metaverse}
\centering\label{FigS-D}
\end{figure*}

\subsection{Virtual Resources in Metaverse}
The core assumption and intuition of Metaverse being sustainable lies in the fact that it can use virtual and digital alternatives as a substitute for real-world experiences and physical goods. It is assumed that the virtual experiences and digital products will be carbon-efficient and less resource-intensive in comparison to their counterparts. A paradigm shift in terms of budget allocation from consumers have already been noticed that shows positive signs for a significant sustainability impact. For instance, Denim trade globally consumes 4.7 billion $m^3$ of water while emitting 16.0 metric tons of carbon dioxide equivalent (MTCO2e) \cite{Denim2021}. If consumers opt for a digital denim for their avatars instead of buying the physical one, it can save substantial amount of water and carbon emissions. According to a study from EY Future Consumer Index\footnote{https://www.ey.com/en\_gl/consumer-products-retail/future-consumer-index-moving-out-of-brands-reach} 21\% of Metaverse consumers intend to buy digital items for their avatars in future rather than investing in physical items. The study \cite{Denim2021} suggested that the transition to virtual denim from physical one can reduce the CO2 emissions by 10\% which is equivalent to annual water consumption and annual emissions of around 350K internal combustion of automobiles in America \cite{GHGE2022,WFP2022}. Considering that a digital transformation from a single product could impact the resource and carbon efficiencies in a significant way, it can be assumed that the Metaverse is a potential candidate to achieve the required environmental sustainability. 

\subsection{Travelling through Metaverse}
Globally, air travel contributes to 2.5\% of CO2 emissions but it was cut in half due to the COVID19 pandemic \cite{Airline2022}. One of the earliest applications of Metaverse was to conduct a virtual meeting with a realistic VR experience. Recently, virtual concerts (Metaverse-based concerts) was organized which was attended by thousands of people in a virtual world \cite{Concert2022} that indicates that both the business and recreational travel can be replaced by Metaverse to a certain degree. Furthermore, with such an immersive experience, companies and businesses can promote the work-from-home culture to cut CO2 emissions. Although, physical collaboration and personal presence is important and recreating a live gathering would not be as real in Metaverse as it is in physical world, we assume that the Metaverse is not here to completely replace it. Rather it can be considered as an effective tool to reduce complexity and expense of discretionary trips while maintaining environmental sustainability.

\subsection{Digital Twins and Metaverse}
Digital twins is the biproduct of emerging technologies such as AI, IoT, and augment and virtual reality. With the satellite generated data, digital twin can transform the real-world entitites into virtual representations, hence propelling sustainability with respect to individuals, manufacturing assets, supply chains, and so forth. Currently, there are many applications streamlined that use digital twins to improve the environmental sustainability. For instance, European space agency is working towards a system that can represent Earth's digital twin to study human activities and its impact on the climate. The simulation model will then be used for making policies to ensure the improvement in environmental sustainability. Similarly, using digital twin for supply chain and manufacturing can improve the optimization for logistics, traceability, energy, processes and material inputs, accordingly. Many industries have already started to adapt digital twin in order to reduce energy and scrap usage that contribute directly to environmental sustainability \cite{Kamble2022}. Digital twins have also been widely used for medical records, modeling data from wearable sensors and forecast individual's health with respect to air pollution and air quality \cite{govind2021}. However, the digital twins have achieved best environmental sustainability in the field of building operations that could help in improving space utilization by 25\%, human productivity is increased by 20\%, maintenance and operational efficiency has been improved by 35\%, and lastly carbon emission for a building can be reduced by around 50\%, respectively \cite{Lydon2021, samad2021, zhao2021}. In recent interviews, Michael Jansen the CEO of Cityzenith (that is involved in making smart and sustainable cities throughout the world) has argued that urban digital twins and real-world Metaverse are synonymous and therefore, Metaverse would help in dealing with multisystem simulation and visualization of buildings and environment in digital space.

\subsection{Penetrating psychological barriers through Metaverse}
It has been suggested by many studies that environmental sustainability and climate change are highly correlated. Companies and industries might have the power to slow down the carbon emissions, however, the masses or general public can be equally held responsible for using the items that are not ecological friendly. Although, past years have recorded around two billion social media posts that address the climate conditions but a recent study showed that the general public bases their idea of climate change with respect to normal weather, therefore, their perception is quite short-term, i.e., the affect that started to come in play since 2 - 8 years \cite{Moore2018}. The use of Metaverse for providing immersive experience and simulation of climate change can tap into consciousness of general public to spur the action towards usage of eco-friendly products, hence, reducing the carbon emission footprint. The relationship of virtual reality and climate change psychology has been carried out and addressed in a recent study by Marowitz \cite{Markovits2021} to compel the general public for opting proenvironmental actions, respectively. 

\subsection{Social Sustainability with Metaverse}
Most of the sustainability issues are centered towards its environmental aspect, however, Metaverse addresses the unsustainable social aspects of the physical world and coalesce social rights such as diversity, accessibility, and equity within its ecosystem. The same was recently highlighted by the Chief Diversity Officer, Maxine Williams by her statement ``Diverse people shouldn't just participate in the Metaverse as consumers; they should be its architects and builders as well''\footnote{https://tech.fb.com/ideas/2022/02/being-intentional-about-diversity-equity-and-inclusion-in-the-Metaverse/}. Recently, a bloomberg report \cite{Cryptopunk2021} also highlighted this issue concerning CryptoPunk NFTs stating that the digital avatar prices varied based on skin color, gender, and race, accordingly. Companies like OvalEdge\footnote{https://www.ovaledge.com/} have identified the problem as unconcious bias which is quite similar to the ones associated with AI models. The company suggest to use metrics and comprehensive set of targel goals for the data scientist to reduce the bias and increase accessibility in Web3. Furthermore, stakeholders, civil society organizations, academia, investors, regulators, and businesses need to collaborate on this matter to make the Metaverse more democratized instead only decentralized. 

\subsection{Sustainable communication and connectivity}
The fifth generation communication system was created with a lot of hype and ambition to bring ubiquitous connectivity to the users. In recent Mobile World Congress (MWC), vice president of distributed edge at VMware quoted that ``5G is another phone in hands that is no different from LTE and 4G. The sixth generation communication system is about putting the focus back on humanity and sustainability''\footnote{https://www.trendingtopics.eu/how-will-6g-and-the-metaverse-change-the-world/}. The discussions on 6G always starts conversation in accordance to human, economic, and societal development. Furthermore, 6G emphasizes more on democratization instead of decentralization which is one of the major concerns with Metaverse and its coarse to sustainability. The sixth generation system is also being developed for achieving sustainability goals through better resource efficiency, less energy consumption, and less power requirement, accordingly. One of the driving forces for 6G advancement was the COVID-19 pandemic that made researchers and people realize about putting more emphasis on economic and societal needs of the broadband network instead of looking for nice presentation of devices and better performance. The 6G and metaverse are a good match for sustainable due to its characteristics of democratization and keeping human in the loop while breaking temporal and spatial barriers for multi-person communication. 
 \ 

\section{Applications and Usecases} 

\subsection{What Are the Use Cases of the Metaverse}

\subsubsection{Metaverse and Physical wold: How Do They Interact}
To understand the different uses of the Metaverse, one must first know what it is. Users can interact with varying spaces as their digital avatar in the Metaverse, an immersive 3D virtual world. Metaverse users can move around different areas as their digital avatars, just like they do in the real world. Users can also create, share, or trade experiences and assets using the Metaverse \cite{Jon2021}. Recently, the proverbial Metaverse was rumored worldwide, and people are taking a great deal of interest in its use cases. Many companies, such as Facebook, noticed it unveiled Meta as the brand name of their parent company at the Facebook Connect event in late October 2021. 
Social media giant Biggie has recently launched Metaverse tools for developers as well. There are three components to it, including a Headset, Presence Platform, and developers' AI toolkit. The interaction of VR/AR objects with users can be programmed. Similarly, Microsoft has invested in Metaverse to create an environment that fosters collaboration with Microsoft Mesh. Almost all major tech companies have jumped aboard the Metaverse bandwagon, doubling the demand for Metaverse use cases.

\subsection{Usecases}
Having a basic understanding of the Metaverse and why it is gaining popularity, you must be looking forward to discovering its use cases. Since the Metaverse is cutting-edge technology, many people wonder, “What are the benefits of a Metaverse?" In what ways will you use the Metaverse to your advantage? What roles will digital reality play in the real world? From identity verification to payments, the world around us has become more and more digital \cite{duan2021Metaverse}. In turn, a digital world like the Metaverse makes it possible for individuals and businesses to change how they view and use technology. Below are some examples of how the Metaverse could be used \cite{du2022rethinking, khan2022Metaverse}.
\subsubsection{Unlocking Marketing Prospects}
A possible use case for Metaverse would be the possibility of unlocking new marketing opportunities \cite{wang2022survey}. People can interact with digital avatars to do many things in the virtual world. 
Furthermore, people engage in the Metaverse's socializing, leisure, and learning activities.  Marketers could take advantage of exclusive marketing opportunities in different virtual worlds in the Metaverse. Marketers have already taken advantage of in-business opportunities in the Metaverse. Such as Anzu, which uses ads to track real-time views in gaming environments across mobile and console platforms. Also included in this list are brands like Paramount and WarnerMedia. In the Metaverse, the ads are reminiscent of real-life, and they are intertwined with the gameplay, where the ads can be found at the right places. For instance, billboard ads or characters wearing branded clothing offer promising brand exposure. Metaverse games have proved how Metaverse can open up new marketing prospects \cite{nevelsteen2018virtual}.

\subsubsection{Blockchain Use Cases}
Blockchain applications could provide advanced technology as the most prominent use case in the Metaverse \cite{gadekallu2022blockchain}.  Decentralizing the Metaverse through the adoption of large-scale Metaverse across various industries is a powerful foundation for the large-scale Metaverse. Blockchains enable the development of dApps and NFTs, power cryptocurrencies, and function as a distributed ledger for recording peer-to-peer transactions. Blockchain technology offers several benefits for engaging the NFT marketplaces fostering new and realistic avenues \cite{kiong2022Metaverse}. Users can communicate with other users and share more interactive Metaverses in the NFT marketplaces. Users can interact with NFTs to make informed purchasing decisions via the Metaverse. Metaverse blockchain application cases include the promotion of new NFTs or the development of a shared virtual area. The blockchain-based games trend has attracted a great deal of traction. Players in NFT or blockchain games can gain various in-game collectibles to exchange with other players or on external marketplaces. The Metaverse can power Blockchain-based online gaming \cite{mozumder2022overview}.

\subsubsection{Virtual Tourism}
Virtual tourism is another promising use case of the Metaverse \cite{gursoy2022Metaverse,um2022travel}. The development of advanced technologies has led to the point where one can enjoy traveling without physically reaching the destination. There is a big difference between experiencing such destinations in person and watching them on video. The Metaverse could be used to create immersive digital environments through virtual reality (VR) \cite{huang20193d,d2020markerless},  and augmented reality (AR) \cite{bajireanu2019mobile,nuzzi2020hands, wang2020avatarmeeting}. You may stir your audience's imagination by merging immersive digital reality with real content. As a result, people can experience their environment as physically present. The possibility for widespread adoption and acknowledgment of VR tourism is a possible mainstream use case.  There is an increasing amount of $360^\circ$ video content on video streaming platforms such as YouTube and other content hosting services. A serious drawback of using Metaverse for virtual tourism is its limited freedom, as you can notice with the use cases. Tourism destinations are not flexible enough to allow people to move around since they can only be viewed recorded.
\subsubsection{Web Real-time Communication}
A Metaverse can also facilitate real-time communication in web experiences, based on the search for blockchain use cases in a  \cite{kshetri2022web}. A real-time communication initiative for mobile applications and web browsers is known as real-time web communication. One of the most intriguing applications is the Metaverse to transform traditional audio and visual communication.
You can transfer information between clients through real-time web communication without requiring an intermediary server. Thanks to peer-to-peer communication in the Metaverse \cite{ryskeldiev2018distributed}, direct communication between browsers could become possible. In addition to providing a platform for designing new web communication standards, the Metaverse technology offers an intriguing application case. Web real-time communication can also be combined with numerous media streams, which is important for constructing virtual worlds.

\subsubsection{Virtual Office and Learning Spaces}
Remote working grew as a result of a global pandemic. The pandemic exposed professionals from various sectors to Skype calls, Microsoft Teams, and Zoom meetings. Remote professionals were able to communicate virtually with these tools \cite{bennett2022remote}. Metaverses offer many possibilities for developing virtual office spaces or learning environments \cite{ popescu2022virtual, ludlow2007second}. Working or learning together in the Metaverse is possible if you feel like you are in the same room. Virtuworx, for example, has created virtual office spaces by leveraging Metaverse use cases. With its hybrid of virtual and mixed reality environments, the company offers a meaningful and productive work environment. Team members can access various functions such as events, offices, conferences, virtual training, and trade exhibitions by using a completely configurable solution.
Metaverse benefits are found in the education sector as well. Minecraft and Second Life have been used to improve student learning experiences in universities. Students in architecture and medicine may also benefit from VR simulations in the Metaverse.

\subsection{Benefits of Metaverse}
It is equally important to know the benefits of the Metaverse after understanding its use cases. In the different examples of how a Metaverse might be used, an impression can be gained of its potential advantages \cite{lee2021all, ryskeldiev2018distributed}. As technology advanced, it answered many uncertainties like who would have imagined that people could make video calls over long distances to share data in seconds. If someone wondered about the benefits of the Metaverse in the past, you might not have found many answers, but at the moment, they seem evident to everyone. However, virtual spaces and digital communication avenues are used by millions of people worldwide to socialize remotely. Beyond adding real-world capabilities to virtual worlds, Metaverse has several benefits for the digital world. Here are a few of the Metaverse technology's key advantages.

\subsubsection{Innovation in Healthcare}
To understand how healthcare sectors fit into the Metaverse, we need to examine their promising benefits. For example, the Metaverse provides a unique opportunity for patients to interact with healthcare professionals regardless of their geographic location \cite{thomason2021metahealth}. Healthcare professionals can interact with patients in virtual worlds in real-time through the Metaverse. In addition, medical students can engage in engaging and comprehensive learning experiences through virtual reality simulation in the Metaverse \cite{tan2022Metaverse}.

\subsubsection{Metaverse and Exciting New Games}
The Metaverse will play a crucial role in gaming applications through play-to-earn models, which enable developers, publishers, and users to reap economic rewards from gaming experiences \cite{nevelsteen2018virtual, shin2022actualization}. On different online marketplaces, gamers can, for instance, create and trade in-game assets. Thus, the benefits of decentralization inherent in blockchain combined with Metaverse can revolutionize gaming. 

\subsubsection{Completely New Economy}
One of the most prominent assumptions about the Metaverse is the possibility of a creator economy. Trading assets across different spaces is made possible by the Metaverse \cite{hamilton2022deep, mackenzie2022criminology}. On another platform in the Metaverse, you can sell an NFT you created in a Metaverse game. A great deal of economic growth can be attributed to DeFi, NFTs, and blockchain games.

\subsection{Metaverse Market: State of the Art}
Without reflecting on the state of the current Metaverse, many people are intrigued by the question, “is the Metaverse the next Internet?" What number of projects are currently being developed? Is the Metaverse attracting attention from big names? Take Facebook, for example. Can we expect other tech giants to pay attention? There must be multiple organizations, creators, and developers creating the Metaverse, no matter who owns it. Metaverse adoption will also be mainly determined by its names. These are some of the top companies that have begun to develop a Metaverse \cite{mystakidis2022Metaverse}.

\subsubsection{\textbf{Facebook}}
Many tech enthusiasts have viewed Facebook's Metaverse announcement as a PR stunt. It is crucial to consider how a tech giant like Facebook can help create a more promising market for the Metaverse. Furthermore, Facebook has almost all of its resources to develop the Metaverse, from infrastructure to human interaction to discovery \cite{fernandez2022facebook}. The online networking behemoth has a massive advertising engine, Oculus headsets, and a burgeoning creator economy.

\subsubsection{\textbf{Epic Games}}
Epic Games, a well-known game production studio, competes for high positions in the Metaverse's future. Epic Games co-founder Tim Sweeney outlined a vision for the Metaverse a few years ago, and now the firm has the appropriate platform for it to thrive. Epic Games can potentially grow Metaverse growth with 1 billion dollar in funding from Sony in the future \cite{kim2021advertising}.

\subsubsection{\textbf{Icrosoft}}
Microsoft's efforts to develop the Metaverse cannot be ignored under any circumstance. The involvement of companies like Microsoft assuages concerns about the Metaverse's future. Microsoft is constructing Microsoft Mesh, a business-focused Metaverse \cite{cheng2022will}. Users would be able to connect to Microsoft products through a digital environment seamlessly. Using Microsoft Mesh, users can leverage Windows, Teams, and other services through VR.

\subsubsection{\textbf{Decentraland}}
Decentraland is an example of the resiliency of the Metaverse market. Metaverse is one of the first pioneering products that explicitly mentions the Metaverse. Users can trade in virtual real estate and create NFTs on Decentraland, developed in 2017. A recent auction in Decentraland featured virtual real estate worth more than 2 million \cite{vidal2022new,ante2021non}. This means Decentraland is likely to become one of the top names in the Metaverse of the future. Microservices and blockchain technologies are also included, along with edge computing and AI agents.

\section{Projects} 
An interconnected network of 3D virtual worlds is one way to define the concept definition of the Metaverse. Through a virtual reality headset, users enter these worlds, navigating the Metaverse with their eye movements, feedback controls, and voice instructions. The user is submerged by the headgear, producing a phenomenon called presence, which is achieved by simulating the genuine physical experience of being present. But before the Metaverse is widely and universally adopted, there are obstacles to be resolved. The virtual aspect of this environment is a significant obstacle. Although a VR headset is often required for entry into the Metaverse, this is not the only need. It is expected that the Metaverse industry will overcome these challenges as there is a tremendous growth in the sales of headsets supporting Metaverse concepts.

There are many players entering into the Metaverse universe in the recent years. The few significant ongoing Metaverse projects are shown in Table VII.

\begin{table*}[]
\centering
\caption{Ongoing Metaverse projects}
\label{bcrtable}
\resizebox{\textwidth}{!}{\begin{tabular}{|c|c|c|c|}
\hline
{\color[HTML]{0E101A} \textbf{Project Title}} &
  {\color[HTML]{0E101A} \textbf{Project sub-title}} &
  \textbf{Project Length} &
  \textbf{Project URL} \\ \hline
{\color[HTML]{0E101A} Decentraland} &
  {\color[HTML]{0E101A} \begin{tabular}[c]{@{}c@{}}Virtual advertisement for content, \\ goods and services\end{tabular}} &
  2015 - ongoing &
  https://decentraland.org/ \\ \hline
{\color[HTML]{0E101A} Oculus (Meta Quest)} &
  {\color[HTML]{0E101A} \begin{tabular}[c]{@{}c@{}}Meta's Metaverse platform - \\ VR equipment, experiences and games\end{tabular}} &
  2014 - ongoing &
  https://store.facebook.com/quest/ \\ \hline
{\color[HTML]{0E101A} Enjin} &
  {\color[HTML]{0E101A} NFTs for everyone} &
  2009 - ongoing &
  https://enjin.io/ \\ \hline
Silks &
  \begin{tabular}[c]{@{}c@{}}Horse racing and bidding in \\ Metaverse through NFTs\end{tabular} &
  Upcoming (2022 - 2023) &
  https://silks.io \\ \hline
\begin{tabular}[c]{@{}c@{}}University of Miami \\ - XR initiative\end{tabular} &
  Multiple XR and Metaverse based projects &
  2018 - ongoing &
  https://xr.miami.edu/projects/index.html \\ \hline
MirageXR &
  Learn through holograms &
  2021- ongoing &
  https://wekit-ecs.com/team \\ \hline
Highstreet &
  Metaverse shopping and product showcases &
  2021 - ongoing &
  https://www.highstreet.market/ \\ \hline
Metahero &
  \begin{tabular}[c]{@{}c@{}}3D scanning of humans to create their \\ own NFT avatars for Metaverse\end{tabular} &
  2021 - ongoing &
  https://metahero.io/ \\ \hline
\end{tabular}}
\end{table*}

\subsection{Decentraland}
Decentraland is a decentralized blockchain-based Metaverse. Decentraland focuses on creating, exhibiting, and selling real-world and NFT assets. Decentraland is governed by a Decentralized Autonomous Organization (DAO), which owns the necessary smart contracts based on which Decentraland operates. Users can vote using the DAO system to influence the various aspects of the Metaverse. The assets can be bought using multiple currencies, such as polygon and ethereum, while the parcels of land on the Metaverse can be purchased only using ethereum. 

Major brands such as Samsung, Atari, and Adidas have bought plots of land in Decentraland and established their products' showcases. The various brands in the Decentraland perform live events and immersive experiences. For example, Samsung recently hosted a live event of their Galaxy S22 unveiling in the Decentraland.

\subsection{Oculus}
Oculus is the online platform that holds the experiences and games accessed through Meta's VR headset platform (Quest). Though it is not a Metaverse of its own, it provides a variety of backgrounds and games that can be experienced through the headset. The significance of Oculus is that it serves as the ground for Meta's future planned expansion into the Metaverse concepts. Hence, it is vital to closely monitor the developments in Oculus even though it is not a Metaverse yet.

\subsection{Enjin}
Enjin is a marketplace and platform that deal with the creation of NFTs for Metaverses. The Enjin provides Software Development Kits (SDKs) that support these activities. Enjin works with two different types of currencies - Enjin token and Enjin coin. Furthermore, they provide integration support to integrate the trading of the created NFTs inside your Metaverse applications. Examples of Metaverses running on Enjin platform are AlterVerse, MotoBloq and Dvision Network. Enjin also actively funds Metaverse projects and such projects can apply for funding through their website.

\subsection{Silks}
Silks, a game platform for the Metaverse, mimics the thoroughbred horse racing industry in real life with the use of a blockchain-enabled Metaverse. A play-to-earn gaming economy powers the Silks Metaverse, enabling people to own racehorses and horse farms while earning tokens via skillful games and contributions to the Silks ecosystem. The users will be able to purchase, gather, exchange, and interact with digital assets that reflect genuine thoroughbred racehorses. Users will also be able to buy, build, and maintain horse farms, other real estates, and interactive digital assets. The project is underdevelopment and is expected to be fully functional by 2023.

\subsection{XR initiative}
XR initiative is a initiative from Miami University to explore use cases of XR and VR and their interoperability. The XR initiative is significant since it focuses on non-gaming domains such as education and healthcare. Though the XR initiative is not a complete Metaverse of its own, the various projects provide a possible look into the utility and usecases of Metaverse in the future. The projects part of this initiative include anesthesia training for nurses and technological training for persons suffering from autism spectrum disorder.

\subsection{MirageXR}
MirageXR enables educators and developers to create unique holographic training programs, enhancing workplace efficiency, enabling data analysis to evaluate performance, and assisting industrial organizations in saving money on training expenses. The app's core is technology-enhanced learning, which helps users go beyond simply remembering and comprehending the content of learning exercises by actually performing the action themselves. 

The project's authoring and performance analytics capabilities allow AR course designers to update activities, gauge engagement quickly, and assist learners in analyzing their actions by recording practice sessions to enhance practice. 

\subsection{Highstreet}
A Metaverse with a focus on business. Web browsers can be used to access the market. The Metaverse promises possibilities to interact with famous people who will introduce their brands on the Highstreet. There are face-to-face interactions available in the Highstreet Metaverse. The Highstreet is home to a unique cryptocurrency lounge where customers may exchange digital currency in a virtual reality setting. Every item sold on Highstreet is displayed to the customer in both digital and physical forms, connected via a product token. In addition to being able to wear a pair of shoes the user purchases in the real world, they may also be added to their avatar's wardrobe in the Metaverse.

\subsection{Metahero}
The Metahero project employs NFT smart contracts and 3D scanning technologies to enable the development of unique meta avatars and meta-objects. Metahero's fundamental technology is 3D scanning, which examines physical objects to gather information about their appearance and digitally depict them. Wolf Studio and Metahero have a partnership for 3D scanning. The concept of Metahero is to scan the user's body using 3D scanning machines and then create realistic NFTs. These NFTs will be utilized as the users' avatars inside the Metaverse. The project is in the development stage, where the users can sign up to be notified if a scanner is available near their location.

\section{Challenges and Future Research Directions} 
This section discuss and enlist the foreseeable challenges and future research trends concerning the role of AI, B5G/6G, and their interplay in Metaverse. 
\subsection{Role of AI in Metaverse}
This subsection highlights foreseeable challenges and future directions concerning the role of AI in Metaverse.

\subsubsection{Integration with Emerging Technologies}
It is apparent from the growth of Metaverse that more and more users will explore and adapt to the aforementioned technology. In order to handle scalablility, Metaverse needs to integrate with emerging technologies such as 5G/6G, Industry 5.0, Internet of Aerial and Ground Vehicles, and so forth. The repercussions or issues that may arise from such integration are yet to be seen but we can assume that the heterogeneity, bandwidth management, energy efficiency, and security would be a few of them. 

\subsubsection{Unique Immersive Experience}
Metaverse is knwon for realistic and immersive experience in digital world. However, as the marketplace gets saturated, the users will demand for unique aspects such as new digital landscapes, more interactive spaces, experiences based on reality, and real-time emotion mapping in digital world. Few of the aforementioned issues can be handled by large language models and methods like Dall-E, but these models and methods are not publicly accessible. Furthermore, the usage of these models cannot be handled by hand-held devices, therefore, the challenge of designing platforms that could provide mutual services and solutions will be in dire need.

\subsubsection{Democratization}
From the review of seminal works, it has been revealed that the Metaverse is far from being democratized at this point. With the growing and diverse users, Metaverse will have to opt for democratization in order to maintain sustainability. Researchers have suggested to use AI-based smart contracts to detect anomalies, but that is yet to be achieved, and the outcomes are yet to be realized. AI is itself subjected to biases, therefore, humans in the loop for making the process of democratization bias free needs to be integrated. 

\subsubsection{Intellectual Property Preservation}
Metaverse has been mainly introduced as an extension of our reality. This extension brings forth the businesses challenges that exist in the real-world such as brand protection, copyright issues, theft of intellectual property and more. Due to being in a digital space, the counterfeiting in Metaverse would be much easier, and reproducibility of the trademarks in the form of virtual images or NFTs would be much cheaper. The challenges for this category are not only related to AI, but also the law enforcement agencies and intellectual filing offices. Furthermore, methods needs to be devised to confront anonymity in the Metaverse as it will hinder the most for content and brand owners to enforce their intellectual property rights. 

\subsection{Role of B5G/6G in Metaverse}
This subsection highlights foreseeable challenges and future directions concerning the role of B5G/6G in Metaverse. 

\subsubsection{Massive, Highly dynamic, and Complex Metaverse}
It is beyond question that Metaverse will be build on top of the highly dynamic, massively dense, and exceptionally complex mobile networks. These networks will be composed of ultra-large scale and inherently heterogeneous equipments. There are, however, many features that are fixed in the architecture of current wireless networks, and the optimization tasks are thus defined to address particular challenges and services that have been identified and defined in advance \cite{fourati2021comprehensive,vittal2021self}. Consequently, the existing process of manually optimizing and configuring networks is not suitable for the Metaverse. Further, the original 5G service classes will be challenged by new immersive experiences such XR gaming, teleoperations, and holographic telepresense that will become available soon in the context of Metaverse \cite{nadir2021immersive}. Therefore, future mobile networks have to simultaneously deliver high transmission rates, high reliability, and low latency in order to effectively provide aforementioned Metaverse services. 

\subsubsection{Zero-touch Management for Metaverse}
Future wireless networks for Metaverse will not rely on manual interventions to avoid any minacious possibilities. Hence, ZTM will play a useful role by controlling, monitoring, and configuring the tasks for massive networks to attain a fully independent closed-loop automation by alleviating any opportunities of human intervention \cite{benzaid2020ai}. In current wireless networks, ZTM has been enabled for specific 5G/6G usage scenarios\cite{andrus2019zero}. However, Metaverse will expose diverse requirements occurring from heterogeneous user application. Hence it demands for a universal ZTM strategy to meet the requirements of several verticals in Metaverse.

\subsubsection{Space-air-ground-sea coverage for Metaverse}
The space-air-ground-sea integrated communication network will empower Metaverse in two folds:
\begin{itemize}
\item Non-terrestrial networks will provide ubiquitious wireless access to users from anywhere at anytime \cite{rinaldi2020non}.
\item Terrestrial networks will support Metaverse users to hop freely among space-air-ground-sea intergrated networks for an uninterrupted immersive experience \cite{lin2022next,tang2022roadmap}.
\end{itemize}

However, the highly frequent fidelity of momentous data among space-air-ground-sea can cause significant delays requiring massive bandwith and overwhelming backhaul resources\cite{chaccour2022seven}. Thus, it becomes imperative to explore new SAGSIN architectures and simulation platforms to underpin the diverse and extensive services and applications of Metaverse. Further, reliability could be compromised in existing SAGSIN architectures due to highly mobile nature of satellites which may eventually cause disruption to Metaverse applications. Hence, novel reliability evaluation metrics are needed to be designed for future services of Metaverse, as proposed by researchers in \cite{du2021optimal}. Figure \ref{meta} shows some of the requirements for seamless meta-immersion and indicators to evaluate the performance of services. 
\begin{figure}[!t]
\includegraphics[width=\linewidth]{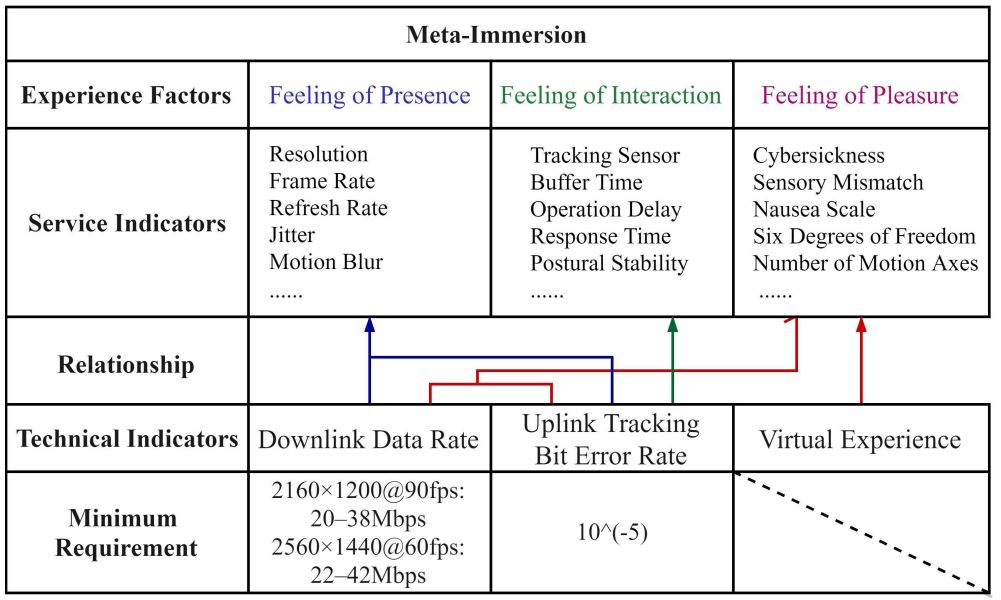}
\caption{Some KPIs and requirements for evaluating the immersiveness in a Metaverse environment \cite{du2021optimal}.}
\centering\label{meta}
\end{figure}
\subsubsection{Tactile Feedback in Metaverse}
It has been discussed in sections IV-D and V-C that certain applications of Metaverse, such as robotic surgery, avatar-to-avatar interaction, teleoperations, and telepresence need to rely on every tiny bits of tactile information occuring from the interaction between physical and meta environment in order to function normally. Although IEEE introduced a standard for tactile Internet (IEEE P1918.1) to normalize such activities, however, the requirements of tactile feedback are ultra-low latency such as in milliseconds\cite{aijaz2018toward}. The existing 5G-URLLC usecases are expected to meet these requirements. Irrespective of that, the tactile Internet poses a number of potential research challenges that have yet to be fully addressed, particularly in the context of Metaverse. For example, efficient routing and MAC protocols for delivering tactile information can be proposed for Metaverse usecases\cite{ghafoor2020mac}. In addition to that, intelligent and adaptive network traffic management can be explored for delivering multi-sensory audio-visual tactile information over wireless networks. Lastly, the computing capabilities can be accelerated for immersive services by apply compression techniques as pro-processors for tactile information.  

\subsection{Integrated Role of AI and 6G in Metaverse}
This subsection highlights foreseeable challenges and future directions concerning the integrated role of AI and 6G in Metaverse. 

\subsubsection{6G-enabled Edge AI in Metaverse}
It is anticipated that Metaverse will bring a complete shift of AI from cloud to the network edge, hence achieving on-device intelligence for Metaverse services. Edge AI can certainly benefit Metaverse in achieving low latency response while ensuring privacy of the user's data\cite{chang20226g,zawish2022towards,zawish2022complexity}. However, the offloading of Metaverse services will pose severe challenges for the existing wireless networks. For instance, Metaverse users based in remote regions with poor Internet connectivity might face a disconnect due to high latency. For mission-critical applications, this latency can potentially lead to worse consequences. Hence, modern communication systems can leverage 6G-uMBB services to support Metaverse applications on the end devices\cite{han2021abstracted}. Edge caching can also be explored to further reduce the transmission latency. Furthermore, Metaverse will enable collection of sheer amount of hetrogenous data (due to diverse users/services), hence edge devices can leverage the federated learning to perform model training locally without violating any privacy laws as shown in Figure \ref{Fig10}.
\subsubsection{AI-based Network Automation for Metaverse}
In a Metaverse composed of heterogeneous users and services, the network automation will be of key importance for delivering seamless QoE to users. AI has to play a vital role in enabling self-adaptive capabilities in 6G networks. Recently, AI-based frameworks have been widely used for empowering automation in 6G networks by allowing them to self-allocate, self-configure and self-optimise the network equipment and resources \cite{rizwan2021zero}. However, as anticipated in Metaverse, these abilities can be hindered by following challenges when leveraging AI: 
\begin{itemize}
    \item Lack of labeled training data to update the AI model given the changing context
    \item The complexity of underlying deep learning models to be ported on resource-constrained devices
    \item The limitation in availability of hardware to support new and emerging intelligent services for Metaverse
\end{itemize}

To address the aforementioned challenges, the possible research directions would be to adapt self-supervision learning techniques for training AI models, and distributing the AI models in the end-edge-cloud continuum to alleviate the computational load on end/edge devices.

\subsubsection{Energy efficient AI and 6G for Metaverse}
Energy efficiency will remain amongst key KPIs for Metaverse developers when proposing AI and 6G driven solutions. Based on the integrated role of AI and 6G, as discussed in section V, it can be said that both will be highly utilized for delivery of immersive services in Metaverse while tackling the density and mobility of users. However, in order to meet the Metaverse requirements for ubiquitous intelligence from anywhere at anytime, 6G networks will require abundant resources and their maintenance \cite{zhao2021}. While some of the resources might be highly leveraged, the others will rather remain redundant or unutilized most of the time, hence consuming the equivalent energy. To address this, network resources should be either adaptive or designed specifically according to the energy requirements of a certain Metaverse application \cite{benzaid2020ai,andrus2019zero}. Moreover, since most of the intelligence will be hosted locally, it will consume relatively higher energy than cloud-based intelligence, and will eventually cause the batteries to drain faster. In such cases, the end devices should be capable of harvesting energy from their surroundings in order to remain self-sufficient \cite{chen2018novel}. Apart from that, modern AI models require significant energy for their execution due to their nature of being highly parameterised. Therefore, a potential research direction would be to either remove the redundant parameters from AI models by trading the accuracy or adaptively offload the models to edge or cloud based on the intermittent energy available on the devices. 

\subsection{Sustainable Metaverse}
This subsection highlights the foreseeable challenges and future directions concerning sustainable Metaverse.
\subsubsection{Climate Change}
Although the Metaverse has the potential to reduce carbon emissions, substantially through the virtual interactions and digital presence, but it can also be harmful to the environment if used improperly. For instance, with the emergence of Metaverse, there would be a need to powerup data centers, thus the companies adopting Metaverse would need to shift to hyperscale data centers for meeting energy demands. Furthermore, most of the ecommerse associated with Metaverse is performed through NFTs and blockchain based processes which is an energy intensive process. Metaverse would need to move towards proof-of-stake where the transactions are concerned as they are less energy intensive. Furthermore, AI can be used to optimize the efficiency gains while proof-of-stake transactions are performed to further lower the energy consumption. 

\subsubsection{Digital Divide}
Studies have suggested that certain part of societies have better access to technology, therefore, the introduction to new technology in the market foster homogeneity instead of diversity. For instance, statista reported that Western Europe and North America makes up almost 90 percent of the VR set sales\footnote{https://www.statista.com/statistics/685774/ar-vr-headset-sales-share-by-region-worldwide/}. Very recently, PR Newswire suggested that North America will be the epicenter of Metaverse growth in next four years\footnote{https://www.prnewswire.com/news-releases/Metaverse-market-38-of-growth-to-originate-from-north-america--information-by-device-vr-and-ar-devices-and-computing-devices-and-geography--forecast-till-2026-301531082.html}. In order for Metaverse to be compliant with sustainable development goals, it needs to move forward towards equitable distribution of the technology, thus, lowering the digital divide in the digital world. 

\subsubsection{Resentment and Division}
Currently, the Metaverse is considered to be an extension of our reality or a digital space which is not dependent on the current mental state. However, continuous exposure and overuse of the Metaverse can create societal implications such as creating a parallel reality that affects the mental state and well-being of an individual. In order to make Metaverse sustainable, it needs to undergo governance and audit, so that an individual's mental state should not be overwhelmed and the individual's perception of digital world, i.e. Metaverse should be disjoint from its physical state, accordingly. 

\subsubsection{Security and Privacy}
The Metaverse not only collects two-dimensional Internet data but also the three dimensional environment data, which, if leaked can have serious implications in the physical world. Furthermore, the Metaverse enables the users to create their multiple personalities which can elevate the challenge of security and privacy to manifold. The use of AI and blockchain for not only managing personalities, identities but also that governance, data ownership, authentication and transactions needs to be undertaken. 

 \

\section{Lessons Learned} 
Based on the systematic review and tutorial like information presented in previous sections, the practical lessons and recommendations are as follows:

\subsubsection{Art and Immersive Experience}
A continuous advancement has been recorded in the field of AI, let it be in the form of self-supervised learning, explainable AI, generative adversarial networks, variational autoencoders, transformers, or large-language models \cite{DALLE1,GPT3,GauGAN}. All of the aforementioned methods have been used in some capacity or another to make the visualization of images/frames better and unique. Recently, large-language models such as Dall-E2 \cite{DALLE2} has taken the Internet by storm with the creation of realistic images via language understanding. It will help not only with the creation of scintillating art but also unique visual experience for the users.

\subsubsection{Circular Economy}
In simple words, circular economy is all for producing waste and circulating materials and products in order to preserve nature. Although many companies have move towards eco-friendly products but the impact is not substantial. We have discussed as to how the travel reduces CO2 emissions. In 2017 alone, the conference and exhibition industry was estimated to be 2.5 trillion dollars of worth, which requires substantial travel. Metaverse can help reduce the air travel by making the exhibition and conference industry digital/virtual, which will not only help in achieving sustainable development goals but will also help in cost savings for both the parties, hence a more circular approach \cite{Eshghie2022}. Similarly, a game could be designed that can motivate the consumers for not using plastic items and using recycled products for minting a fraction of a coin (mining approach).

\subsubsection{Sustainability in Metaverse}
It has been suggested by researchers as well as industry personnel that Metaverse can help in improving overall sustainability \cite{Deveci2022}. It can be the motivation for users to opt for digital clothing, virtual concerts, virtual meetings, and so forth. If used the right way, achieving sustainability through Metaverse would be inevitable. However, it could be used otherwise to disrupt the sustainability goals, therefore, regulations, guidelines and laws needs to be made and enforced for making the Metaverse a savior. Furthermore, the use of digital twins has already proven to be a positive way towards sustainability. In addition, the Metaverse also need to work on social sustainability so that the virtual system could achieve democratization instead of only moving towards decentralization.

\subsubsection{Layered Security in Metaverse}
Metaverse is without a doubt a game changer but with the increasing number of users, it will also be a nightmare for security personnel. It is estimated that cyber crimes would increase manifold as the Metaverse move toward commercialization. For instance, a successful business deal can turn to a disaster if the avatar (person) is impersonating someone else. Therefore, a layered security will be essential to secure the data as well as AI methods in Metaverse. Emerging technologies such as Federated learning, Private AI, and Blockchain along with encryption and privacy-preservation methods would be used to maintain security in Metaverse ecosystem \cite{PrivateAI, PGSL}.

\subsubsection{Wireless Interactivity in Metaverse}
The previous generations of wireless communication systems have concentrated mostly on improvement of connectivity. However, the future generations of mobile terrestrial systems will be blending connectivity with virtual and physical space, as well as ubiquitous intelligence. There is a potential for the wireless interactivity, which is made possible through future communication technologies, to play an important role in the establishment of links between cyberspace and physical space, as well as among humans. THz communication will be particularly useful for ultra high data rate communications within short distances that have absolutely zero error rates, as well as high data rates \cite{pengnoo2020digital,Liu_2021,fantacci2021edge}. Moreover, in the near future, AI will have a huge impact on the future of 6G-powered wireless communications by becoming both native and ubiquitous, which will enable all the components of 6G to be intelligent and user-friendly. As the Metaverse grows, 6G is also likely to be dominated by an increase in interactivity, which will be one of its key fundamentals \cite{zhang2019edge}. 6G-based interactivity will provide Metaverse users with multi-sensory experiences that are immersive and tactile, hence enriching their daily lives and enhancing their quality of life. 

\subsubsection{Diverse Service Requirements in Metaverse}
It is possible that the mobile networks will not be utilized to their full potential due to the diverse and sometimes conflicting requirements of the various Metaverse services and applications. In such cases, 6G networks allow slicing of the network into multiple virtual and standalone logical networks while sharing a same physical infrastructure \cite{wallace2021high,khan2020network}. Thus, each network slice can be customised to serve a Metaverse application in an efficient way by taking care of the energy efficiency, latency, dense connectivity, mobility, and bandwith. A key feature of 6G networks will be the provisioning of network slices in a dynamic manner \cite{chowdhury20206g}. Rather than categorizing Metaverse applications into eMBB, mMTC, and URLLC as in the traditional networks, 6G networks will be capable of providing dynamic types of service depending on the network traffic and the user requirements. 

\subsubsection{Self-sufficient Infrastructure in Metaverse}
Existing communication and computing infrastructures are limited in terms of scaling the services and applications according to dynamic requirements. As an example, Facebook space (a VR-based social network), is only able to handle a maximum of three people to socialise in the environment it provides. Metaverse will bring many essential use cases for which this is inefficient, including virtual classrooms,  holoconferences and events, and virtual tourism \cite{lee2022virtual,tayal2022virtual,kye2021educational, lee2021all}. With the advent of decentralised intelligence, multi-access edge computing, microservices, and reinforcement learning techniques, a ray of optimism shines through the way for such immersive applications \cite{ren2020edge,loven2019edgeai,dhelim2022edge,lim2022realizing}. A Metaverse environment supported by 6G networks will be made more responsive by automating the wireless networks with advanced deep learning methods. In particular, 6G combined with the incredible capabilities of reinforcement learning has the potential to truly transform the existing infrastrcture and making them suitable for Metaverse environments. Moreover, implementing AI as a native function of the network will enable 6G to become the first-generation network to promote large-scale deployment of self-optimized and automated networks \cite{wu2021toward,Gu2020}. As a result of the latest advances in aforementioned technologies, practices, and methods, the future wireless networks will be able to co-exist according to their state and achieve the goals set for them. 
 \

\section{Conclusion}
This survey paper provides a comprehensive review on the role of AI and 6G in realising the immersive experiences of Metaverse. In particular, we investigated several underlying technologies of AI and 6G, such as advancements in computer vision, learning paradigms, and wireless communication technologies in the context of Metaverse. Besides that, we explore the joint role of AI and 6G technologies in obtaining ubiquitous intelligence, tactile feedbacks, and self-optimising capabilities for several Metaverse services ranging from holographic telepresence to remote surgeries. Next, we highlight the sustainable aspect of Metaverse services followed by the applications, usecases, and on-going projects. Lastly, we enlighten numerous open issues, future directions, and lessons learned for potential researchers and developers of Metaverse applications and services.

\bibliographystyle{IEEEtran}
\bibliography{AI-6G}

\begin{thebibliography}{100}
\providecommand{\url}[1]{#1}
\csname url@samestyle\endcsname
\providecommand{\newblock}{\relax}
\providecommand{\bibinfo}[2]{#2}
\providecommand{\BIBentrySTDinterwordspacing}{\spaceskip=0pt\relax}
\providecommand{\BIBentryALTinterwordstretchfactor}{4}
\providecommand{\BIBentryALTinterwordspacing}{\spaceskip=\fontdimen2\font plus
\BIBentryALTinterwordstretchfactor\fontdimen3\font minus
  \fontdimen4\font\relax}
\providecommand{\BIBforeignlanguage}[2]{{%
\expandafter\ifx\csname l@#1\endcsname\relax
\typeout{** WARNING: IEEEtran.bst: No hyphenation pattern has been}%
\typeout{** loaded for the language `#1'. Using the pattern for}%
\typeout{** the default language instead.}%
\else
\language=\csname l@#1\endcsname
\fi
#2}}
\providecommand{\BIBdecl}{\relax}
\BIBdecl

\bibitem{khan2022Metaverse}
L.~U. Khan, Z.~Han, D.~Niyato, E.~Hossain, and C.~S. Hong, ``Metaverse for
  wireless systems: Vision, enablers, architecture, and future directions,''
  \emph{arXiv preprint arXiv:2207.00413}, 2022.

\bibitem{ning2021survey}
H.~Ning, H.~Wang, Y.~Lin, W.~Wang, S.~Dhelim, F.~Farha, J.~Ding, and
  M.~Daneshmand, ``A survey on metaverse: the state-of-the-art, technologies,
  applications, and challenges,'' \emph{arXiv preprint arXiv:2111.09673}, 2021.

\bibitem{lee2021all}
L.-H. Lee, T.~Braud, P.~Zhou, L.~Wang, D.~Xu, Z.~Lin, A.~Kumar, C.~Bermejo, and
  P.~Hui, ``All one needs to know about metaverse: A complete survey on
  technological singularity, virtual ecosystem, and research agenda,''
  \emph{arXiv preprint arXiv:2110.05352}, 2021.

\bibitem{lee2021creators}
L.-H. Lee, Z.~Lin, R.~Hu, Z.~Gong, A.~Kumar, T.~Li, S.~Li, and P.~Hui, ``When
  creators meet the metaverse: A survey on computational arts,'' \emph{arXiv
  preprint arXiv:2111.13486}, 2021.

\bibitem{bhattacharya2022Metaverse}
P.~Bhattacharya, M.~S. Obaidat, D.~Savaliya, S.~Sanghavi, S.~Tanwar, and
  B.~Sadaun, ``Metaverse assisted telesurgery in healthcare 5.0: An interplay
  of blockchain and explainable ai,'' in \emph{2022 International Conference on
  Computer, Information and Telecommunication Systems (CITS)}.\hskip 1em plus
  0.5em minus 0.4em\relax IEEE, 2022, pp. 1--5.

\bibitem{gadekallu2022blockchain}
T.~R. Gadekallu, T.~Huynh-The, W.~Wang, G.~Yenduri, P.~Ranaweera, Q.-V. Pham,
  D.~B. da~Costa, and M.~Liyanage, ``Blockchain for the metaverse: A review,''
  \emph{arXiv preprint arXiv:2203.09738}, 2022.

\bibitem{cao2022decentralized}
L.~Cao, ``Decentralized ai: Edge intelligence and smart blockchain, metaverse,
  web3, and desci,'' \emph{IEEE Intelligent Systems}, vol.~37, no.~3, pp.
  6--19, 2022.

\bibitem{ilyina2022Metaverse}
I.~A. Ilyina, E.~A. Eltikova, K.~A. Uvarova, and S.~D. Chelysheva,
  ``Metaverse-death to offline communication or empowerment of interaction?''
  in \emph{2022 Communication Strategies in Digital Society Seminar
  (ComSDS)}.\hskip 1em plus 0.5em minus 0.4em\relax IEEE, 2022, pp. 117--119.

\bibitem{vretos2019exploiting}
N.~Vretos, P.~Daras, S.~Asteriadis, E.~Hortal, E.~Ghaleb, E.~Spyrou, H.~C.
  Leligou, P.~Karkazis, P.~Trakadas, and K.~Assimakopoulos, ``Exploiting
  sensing devices availability in ar/vr deployments to foster engagement,''
  \emph{Virtual Reality}, vol.~23, no.~4, pp. 399--410, 2019.

\bibitem{lim2022realizing}
W.~Y.~B. Lim, Z.~Xiong, D.~Niyato, X.~Cao, C.~Miao, S.~Sun, and Q.~Yang,
  ``Realizing the metaverse with edge intelligence: A match made in heaven,''
  \emph{arXiv preprint arXiv:2201.01634}, 2022.

\bibitem{torres2020immersive}
M.~Torres~Vega, C.~Liaskos, S.~Abadal, E.~Papapetrou, A.~Jain, B.~Mouhouche,
  G.~Kalem, S.~Erg{\"u}t, M.~Mach, T.~Sabol \emph{et~al.}, ``Immersive
  interconnected virtual and augmented reality: A 5g and iot perspective,''
  \emph{Journal of Network and Systems Management}, vol.~28, no.~4, pp.
  796--826, 2020.

\bibitem{al2018experimental}
M.~Al~Ja'afreh, H.~Adharni, and A.~El~Saddik, ``Experimental qos optimization
  for haptic communication over tactile internet,'' in \emph{2018 IEEE
  International Symposium on Haptic, Audio and Visual Environments and Games
  (HAVE)}.\hskip 1em plus 0.5em minus 0.4em\relax IEEE, 2018, pp. 1--6.

\bibitem{polachan2022assessing}
K.~Polachan, J.~Pal, C.~Singh, and T.~Prabhakar, ``Assessing quality of control
  in tactile cyber-physical systems,'' \emph{IEEE Transactions on Network and
  Service Management}, 2022.

\bibitem{tariq2022toward}
M.~Tariq, F.~Naeem, and H.~V. Poor, ``Toward experience-driven traffic
  management and orchestration in digital-twin-enabled 6g networks,''
  \emph{arXiv preprint arXiv:2201.04259}, 2022.

\bibitem{aijaz2016realizing}
A.~Aijaz, M.~Dohler, A.~H. Aghvami, V.~Friderikos, and M.~Frodigh, ``Realizing
  the tactile internet: Haptic communications over next generation 5g cellular
  networks,'' \emph{IEEE Wireless Communications}, vol.~24, no.~2, pp. 82--89,
  2016.

\bibitem{barnett2018cisco}
T.~Barnett, S.~Jain, U.~Andra, and T.~Khurana, ``Cisco visual networking index
  (vni) complete forecast update, 2017--2022,'' \emph{Americas/EMEAR Cisco
  Knowledge Network (CKN) Presentation}, pp. 1--30, 2018.

\bibitem{ana2021study}
N.~Ana-Maria, M.~Alexandru, and P.~E. Cristian, ``Study of millimeter waves in
  5g,'' in \emph{2021 IEEE International Black Sea Conference on Communications
  and Networking (BlackSeaCom)}.\hskip 1em plus 0.5em minus 0.4em\relax IEEE,
  2021, pp. 1--4.

\bibitem{ghoshal2022can}
M.~Ghoshal, P.~Dash, Z.~Kong, Q.~Xu, Y.~C. Hu, D.~Koutsonikolas, and Y.~Li,
  ``Can 5g mmwave support multi-user ar?'' in \emph{International Conference on
  Passive and Active Network Measurement}.\hskip 1em plus 0.5em minus
  0.4em\relax Springer, 2022, pp. 180--196.

\bibitem{roh2014millimeter}
W.~Roh, J.-Y. Seol, J.~Park, B.~Lee, J.~Lee, Y.~Kim, J.~Cho, K.~Cheun, and
  F.~Aryanfar, ``Millimeter-wave beamforming as an enabling technology for 5g
  cellular communications: Theoretical feasibility and prototype results,''
  \emph{IEEE communications magazine}, vol.~52, no.~2, pp. 106--113, 2014.

\bibitem{lin2021wireless}
P.~Lin, Q.~Song, F.~R. Yu, D.~Wang, A.~Jamalipour, and L.~Guo, ``Wireless
  virtual reality in beyond 5g systems with the internet of intelligence,''
  \emph{IEEE Wireless Communications}, vol.~28, no.~2, pp. 70--77, 2021.

\bibitem{Gu2020}
B.~Gu, X.~Zhang, Z.~Lin, and M.~Alazab, ``Deep multiagent
  reinforcement-learning-based resource allocation for internet of controllable
  things,'' \emph{IEEE Internet of Things Journal}, vol.~8, no.~5, pp.
  3066--3074, 2021.

\bibitem{maier2022art}
M.~Maier, A.~Ebrahimzadeh, A.~Beniiche, and S.~Rostami, ``The art of 6g (tao
  6g): how to wire society 5.0,'' \emph{Journal of Optical Communications and
  Networking}, vol.~14, no.~2, pp. A101--A112, 2022.

\bibitem{hilty2020review}
D.~M. Hilty, K.~Randhawa, M.~M. Maheu, A.~J. McKean, R.~Pantera, M.~C.
  Mishkind, A.~Rizzo \emph{et~al.}, ``A review of telepresence, virtual
  reality, and augmented reality applied to clinical care,'' \emph{Journal of
  Technology in Behavioral Science}, vol.~5, no.~2, pp. 178--205, 2020.

\bibitem{pengnoo2020digital}
M.~Pengnoo, M.~T. Barros, L.~Wuttisittikulkij, B.~Butler, A.~Davy, and
  S.~Balasubramaniam, ``Digital twin for metasurface reflector management in 6g
  terahertz communications,'' \emph{IEEE access}, vol.~8, pp.
  114\,580--114\,596, 2020.

\bibitem{adhikari20226g}
M.~Adhikari and A.~Hazra, ``6g-enabled ultra-reliable low-latency communication
  in edge networks,'' \emph{IEEE Communications Standards Magazine}, vol.~6,
  no.~1, pp. 67--74, 2022.

\bibitem{Liu_2021}
X.~Liu, F.~Zhang, Z.~Hou, L.~Mian, Z.~Wang, J.~Zhang, and J.~Tang,
  ``Self-supervised learning: Generative or contrastive,'' \emph{IEEE
  Transactions on Knowledge and Data Engineering}, vol. Early Access, pp. 1--8,
  2021.

\bibitem{fantacci2021edge}
R.~Fantacci and B.~Picano, ``Edge-based virtual reality over 6g terahertz
  channels,'' \emph{IEEE Network}, vol.~35, no.~5, pp. 28--33, 2021.

\bibitem{chude2022enabling}
U.~K. Chude-Okonkwo, B.~S. Paul, and A.~A. Vasilakos, ``Enabling precision
  medicine via contemporary and future communication technologies: A survey,''
  \emph{IEEE Access}, 2022.

\bibitem{chakrabarti2021deep}
K.~Chakrabarti, ``Deep learning based offloading for mobile augmented reality
  application in 6g,'' \emph{Computers and Electrical Engineering}, vol.~95, p.
  107381, 2021.

\bibitem{tang2022roadmap}
F.~Tang, X.~Chen, M.~Zhao, and N.~Kato, ``The roadmap of communication and
  networking in 6g for the metaverse,'' \emph{IEEE Wireless Communications},
  2022.

\bibitem{aggarwal2021generative}
A.~Aggarwal, M.~Mittal, and G.~Battineni, ``Generative adversarial network: An
  overview of theory and applications,'' \emph{International Journal of
  Information Management Data Insights}, vol.~1, no.~1, p. 100004, 2021.

\bibitem{huang2019process}
X.~Huang, J.~Twycross, and F.~Wild, ``A process for the semi-automated
  generation of life-sized, interactive 3d character models for holographic
  projection,'' in \emph{2019 International Conference on 3D Immersion
  (IC3D)}.\hskip 1em plus 0.5em minus 0.4em\relax IEEE, 2019, pp. 1--8.

\bibitem{li2021animated}
Z.~Li, L.~Chen, C.~Liu, F.~Zhang, Z.~Li, Y.~Gao, Y.~Ha, C.~Xu, S.~Quan, and
  Y.~Xu, ``Animated 3d human avatars from a single image with gan-based texture
  inference,'' \emph{Computers \& Graphics}, vol.~95, pp. 81--91, 2021.

\bibitem{loven2019edgeai}
L.~Lov{\'e}n, T.~Lepp{\"a}nen, E.~Peltonen, J.~Partala, E.~Harjula,
  P.~Porambage, M.~Ylianttila, and J.~Riekki, ``Edgeai: A vision for
  distributed, edge-native artificial intelligence in future 6g networks,''
  \emph{The 1st 6G wireless summit}, pp. 1--2, 2019.

\bibitem{yang2022fusing}
Q.~Yang, Y.~Zhao, H.~Huang, Z.~Xiong, J.~Kang, and Z.~Zheng, ``Fusing
  blockchain and ai with metaverse: A survey,'' \emph{IEEE Open Journal of the
  Computer Society}, 2022.

\bibitem{wang2022survey}
Y.~Wang, Z.~Su, N.~Zhang, D.~Liu, R.~Xing, T.~H. Luan, and X.~Shen, ``A survey
  on metaverse: Fundamentals, security, and privacy,'' \emph{arXiv preprint
  arXiv:2203.02662}, 2022.

\bibitem{huynh2022artificial}
T.~Huynh-The, Q.-V. Pham, X.-Q. Pham, T.~T. Nguyen, Z.~Han, and D.-S. Kim,
  ``Artificial intelligence for the metaverse: A survey,'' \emph{arXiv preprint
  arXiv:2202.10336}, 2022.

\bibitem{chang20226g}
L.~Chang, Z.~Zhang, P.~Li, S.~Xi, W.~Guo, Y.~Shen, Z.~Xiong, J.~Kang,
  D.~Niyato, X.~Qiao \emph{et~al.}, ``6g-enabled edge ai for metaverse:
  Challenges, methods, and future research directions,'' \emph{arXiv preprint
  arXiv:2204.06192}, 2022.

\bibitem{jagatheesaperumal2022advancing}
S.~K. Jagatheesaperumal, K.~Ahmad, A.~Al-Fuqaha, and J.~Qadir, ``Advancing
  education through extended reality and internet of everything enabled
  metaverses: Applications, challenges, and open issues,'' \emph{arXiv preprint
  arXiv:2207.01512}, 2022.

\bibitem{dhelim2022edge}
S.~Dhelim, T.~Kechadi, L.~Chen, N.~Aung, H.~Ning, and L.~Atzori, ``Edge-enabled
  metaverse: The convergence of metaverse and mobile edge computing,''
  \emph{arXiv preprint arXiv:2205.02764}, 2022.

\bibitem{park2022Metaverse}
S.-M. Park and Y.-G. Kim, ``A metaverse: Taxonomy, components, applications,
  and open challenges,'' \emph{Ieee Access}, vol.~10, pp. 4209--4251, 2022.

\bibitem{mozumder2022overview}
M.~A.~I. Mozumder, M.~M. Sheeraz, A.~Athar, S.~Aich, and H.-C. Kim, ``Overview:
  technology roadmap of the future trend of metaverse based on iot, blockchain,
  ai technique, and medical domain metaverse activity,'' in \emph{2022 24th
  International Conference on Advanced Communication Technology (ICACT)}.\hskip
  1em plus 0.5em minus 0.4em\relax IEEE, 2022, pp. 256--261.

\bibitem{Coinbase:2021}
Coinbase, ``How coinbase thinks about the metaverse,''
  \url{https://blog.coinbase.com/how-coinbase-thinks-about-the-metaverse-16d8070f4841},
  2021.

\bibitem{huang2022analysis}
J.~Huang, P.~Sun, and W.~Zhang, ``Analysis of the future prospects for the
  metaverse,'' in \emph{2022 7th International Conference on Financial
  Innovation and Economic Development (ICFIED 2022)}.\hskip 1em plus 0.5em
  minus 0.4em\relax Atlantis Press, 2022, pp. 1899--1904.

\bibitem{David:2021}
D.~Pereira, ``How ai will shape the metaverse,''
  \url{https://towardsdatascience.com/how-ai-will-shape-the-metaverse-4ea7ae20c99},
  2021.

\bibitem{maccallum2019teacher}
K.~MacCallum and D.~Parsons, ``Teacher perspectives on mobile augmented
  reality: The potential of metaverse for learning,'' in \emph{World Conference
  on Mobile and Contextual Learning}, 2019, pp. 21--28.

\bibitem{sparkes2021Metaverse}
M.~Sparkes, ``What is a metaverse,'' 2021.

\bibitem{kshetri2022web}
N.~Kshetri, ``Web 3.0 and the metaverse shaping organizations’ brand and
  product strategies,'' \emph{IT Professional}, vol.~24, no.~02, pp. 11--15,
  2022.

\bibitem{valaskova2022virtual}
K.~Valaskova, V.~Machova, and E.~Lewis, ``Virtual marketplace dynamics data,
  spatial analytics, and customer engagement tools in a real-time interoperable
  decentralized metaverse,'' \emph{Linguistic and Philosophical
  Investigations}, vol.~21, pp. 105--120, 2022.

\bibitem{Jon1:2021}
J.~Radoff, ``The metaverse and artificial intelligence,''
  \url{https://medium.com/building-the-metaverse/the-metaverse-and-artificial-intelligence-ai-577343895411},
  2021.

\bibitem{Jon2:2021}
------, ``The metaverse value-chain,''
  \url{https://medium.com/building-the-metaverse/the-metaverse-value-chain-afcf9e09e3a7},
  2021.

\bibitem{Jon3:2021}
J.~Radof, ``9 megatrends shaping the metaverse,''
  \url{https://medium.com/building-the-metaverse/9-megatrends-shaping-the-metaverse-93b91c159375},
  May 2021.

\bibitem{aggarwal2022has}
K.~Aggarwal, M.~M. Mijwil, A.-H. Al-Mistarehi, S.~Alomari, M.~G{\"o}k, A.~M.~Z.
  Alaabdin, S.~H. Abdulrhman \emph{et~al.}, ``Has the future started? the
  current growth of artificial intelligence, machine learning, and deep
  learning,'' \emph{Iraqi Journal for Computer Science and Mathematics},
  vol.~3, no.~1, pp. 115--123, 2022.

\bibitem{lik2017interaction}
L.~Lik-Hang and H.~Pan, ``Interaction methods for smart glasses,'' \emph{ACM
  Comput. Surv}, vol.~1, no.~0, 2017.

\bibitem{lee2019hibey}
L.~H. Lee, K.~Y. Lam, Y.~P. Yau, T.~Braud, and P.~Hui, ``Hibey: Hide the
  keyboard in augmented reality,'' in \emph{2019 IEEE International Conference
  on Pervasive Computing and Communications (PerCom}.\hskip 1em plus 0.5em
  minus 0.4em\relax IEEE, 2019, pp. 1--10.

\bibitem{huang20193d}
X.-Y. Huang, M.-S. Tsai, and C.-C. Huang, ``3d virtual-reality interaction
  system,'' in \emph{2019 IEEE International Conference on Consumer
  Electronics-Taiwan (ICCE-TW)}.\hskip 1em plus 0.5em minus 0.4em\relax IEEE,
  2019, pp. 1--2.

\bibitem{d2020markerless}
E.~D’Antonio, J.~Taborri, E.~Palermo, S.~Rossi, and F.~Patane, ``A markerless
  system for gait analysis based on openpose library,'' in \emph{2020 IEEE
  International Instrumentation and Measurement Technology Conference
  (I2MTC)}.\hskip 1em plus 0.5em minus 0.4em\relax IEEE, 2020, pp. 1--6.

\bibitem{bajireanu2019mobile}
R.~Bajireanu, J.~A. Pereira, R.~J. Veiga, J.~D. Sardo, P.~J. Cardoso, R.~Lam,
  and J.~M. Rodrigues, ``Mobile human shape superimposition: an initial
  approach using openpose,'' \emph{International Journal of Computers}, vol.~4,
  2019.

\bibitem{nuzzi2020hands}
C.~Nuzzi, S.~Ghidini, R.~Pagani, S.~Pasinetti, G.~Coffetti, and G.~Sansoni,
  ``Hands-free: a robot augmented reality teleoperation system,'' in \emph{2020
  17th International Conference on Ubiquitous Robots (UR)}.\hskip 1em plus
  0.5em minus 0.4em\relax IEEE, 2020, pp. 617--624.

\bibitem{wang2020avatarmeeting}
X.~Wang, Y.~Wang, Y.~Shi, W.~Zhang, and Q.~Zheng, ``Avatarmeeting: An augmented
  reality remote interaction system with personalized avatars,'' in
  \emph{Proceedings of the 28th ACM International Conference on Multimedia},
  2020, pp. 4533--4535.

\bibitem{ponnusamy2022ai}
V.~Ponnusamy, A.~Vasuki, J.~C. Clement, and P.~Eswaran, ``Ai-driven information
  and communication technologies, services, and applications for
  next-generation healthcare system,'' \emph{Smart Systems for Industrial
  Applications}, pp. 1--32, 2022.

\bibitem{phillip2018new}
A.~Phillip, J.~S. Chan, and S.~Peiris, ``A new look at cryptocurrencies,''
  \emph{Economics Letters}, vol. 163, pp. 6--9, 2018.

\bibitem{zhu2022metaaid}
H.~Zhu, ``Metaaid: A flexible framework for developing metaverse applications
  via ai technology and human editing,'' \emph{arXiv preprint
  arXiv:2204.01614}, 2022.

\bibitem{dang2019deep}
Q.~Dang, J.~Yin, B.~Wang, and W.~Zheng, ``Deep learning based 2d human pose
  estimation: A survey,'' \emph{Tsinghua Science and Technology}, vol.~24,
  no.~6, pp. 663--676, 2019.

\bibitem{cao2017realtime}
Z.~Cao, T.~Simon, S.-E. Wei, and Y.~Sheikh, ``Realtime multi-person 2d pose
  estimation using part affinity fields,'' in \emph{Proceedings of the IEEE
  conference on computer vision and pattern recognition}, 2017, pp. 7291--7299.

\bibitem{fang2017rmpe}
H.-S. Fang, S.~Xie, Y.-W. Tai, and C.~Lu, ``Rmpe: Regional multi-person pose
  estimation,'' in \emph{Proceedings of the IEEE international conference on
  computer vision}, 2017, pp. 2334--2343.

\bibitem{mehta2017vnect}
D.~Mehta, S.~Sridhar, O.~Sotnychenko, H.~Rhodin, M.~Shafiei, H.-P. Seidel,
  W.~Xu, D.~Casas, and C.~Theobalt, ``Vnect: Real-time 3d human pose estimation
  with a single rgb camera,'' \emph{Acm transactions on graphics (tog)},
  vol.~36, no.~4, pp. 1--14, 2017.

\bibitem{moeslund2001survey}
T.~B. Moeslund and E.~Granum, ``A survey of computer vision-based human motion
  capture,'' \emph{Computer vision and image understanding}, vol.~81, no.~3,
  pp. 231--268, 2001.

\bibitem{hu2020fingertrak}
F.~Hu, P.~He, S.~Xu, Y.~Li, and C.~Zhang, ``Fingertrak: Continuous 3d hand pose
  tracking by deep learning hand silhouettes captured by miniature thermal
  cameras on wrist,'' \emph{Proceedings of the ACM on Interactive, Mobile,
  Wearable and Ubiquitous Technologies}, vol.~4, no.~2, pp. 1--24, 2020.

\bibitem{shin2021non}
Y.-j. Shin, H.-j. Lee, J.-h. Kim, D.-y. Kwon, S.-a. Lee, Y.-j. Choo, J.-h.
  Park, J.-h. Jung, H.-s. Lee, and J.-h. Kim, ``Non-face-to-face online home
  training application study using deep learning-based image processing
  technique and standard exercise program,'' \emph{The Journal of the
  Convergence on Culture Technology}, vol.~7, no.~3, pp. 577--582, 2021.

\bibitem{Spatial1}
``What is spatial computing? a short definition of spatial computing,''
  \url{https://www.techslang.com/definition/what-is-spatial-computing/}.

\bibitem{Spatial2}
Coinbase, ``Spatial computing,''
  \url{https://www.ptc.com/en/industry-insights/spatial-computing}.

\bibitem{shekhar2015spatial}
S.~Shekhar, S.~K. Feiner, and W.~G. Aref, ``Spatial computing,''
  \emph{Communications of the ACM}, vol.~59, no.~1, pp. 72--81, 2015.

\bibitem{zhang2016effect}
T.~Zhang, Y.-T. Li, and J.~P. Wachs, ``The effect of embodied interaction in
  visual-spatial navigation,'' \emph{ACM Transactions on Interactive
  Intelligent Systems (TiiS)}, vol.~7, no.~1, pp. 1--36, 2016.

\bibitem{Joe:2020}
J.~Bardi, ``What is virtual reality? [definition and examples,''
  \url{https://www.marxentlabs.com/what-is-virtual-reality/}, 2020.

\bibitem{kelly2021virtual}
J.~W. Kelly, L.~A. Cherep, A.~F. Lim, T.~Doty, and S.~B. Gilber, ``Who are
  virtual reality headset owners? a survey and comparison of headset owners and
  non-owners,'' in \emph{2021 IEEE Virtual Reality and 3D User Interfaces
  (VR)}.\hskip 1em plus 0.5em minus 0.4em\relax IEEE, 2021, pp. 687--694.

\bibitem{milgram1995augmented}
P.~Milgram, H.~Takemura, A.~Utsumi, and F.~Kishino, ``Augmented reality: A
  class of displays on the reality-virtuality continuum,'' in
  \emph{Telemanipulator and telepresence technologies}, vol. 2351.\hskip 1em
  plus 0.5em minus 0.4em\relax International Society for Optics and Photonics,
  1995, pp. 282--292.

\bibitem{speicher2019mixed}
M.~Speicher, B.~D. Hall, and M.~Nebeling, ``What is mixed reality?'' in
  \emph{Proceedings of the 2019 CHI conference on human factors in computing
  systems}, 2019, pp. 1--15.

\bibitem{zyda1999networked}
M.~Zyda, \emph{Networked virtual environments: design and
  implementation}.\hskip 1em plus 0.5em minus 0.4em\relax Addison-Wesley, 1999.

\bibitem{liu2012survey}
H.~Liu, M.~Bowman, and F.~Chang, ``Survey of state melding in virtual worlds,''
  \emph{ACM Computing Surveys (CSUR)}, vol.~44, no.~4, pp. 1--25, 2012.

\bibitem{narumi2011augmented}
T.~Narumi, S.~Nishizaka, T.~Kajinami, T.~Tanikawa, and M.~Hirose, ``Augmented
  reality flavors: gustatory display based on edible marker and cross-modal
  interaction,'' in \emph{Proceedings of the SIGCHI conference on human factors
  in computing systems}, 2011, pp. 93--102.

\bibitem{schmalstieg2016augmented}
D.~Schmalstieg and T.~Hollerer, \emph{Augmented reality: principles and
  practice}.\hskip 1em plus 0.5em minus 0.4em\relax Addison-Wesley
  Professional, 2016.

\bibitem{kruijff20043d}
E.~Kruijff, J.~J. LaViola, and I.~POUPYREV, ``3d user interfaces: theory and
  practice,'' 2004.

\bibitem{pierce2002comparing}
J.~S. Pierce and R.~Pausch, ``Comparing voodoo dolls and homer: exploring the
  importance of feedback in virtual environments,'' in \emph{Proceedings of the
  SIGCHI Conference on Human Factors in Computing Systems}, 2002, pp. 105--112.

\bibitem{lee2021towards}
L.-H. Lee, T.~Braud, S.~Hosio, and P.~Hui, ``Towards augmented reality driven
  human-city interaction: Current research on mobile headsets and future
  challenges,'' \emph{ACM Computing Surveys (CSUR)}, vol.~54, no.~8, pp. 1--38,
  2021.

\bibitem{langlotz2012sketching}
T.~Langlotz, S.~Mooslechner, S.~Zollmann, C.~Degendorfer, G.~Reitmayr, and
  D.~Schmalstieg, ``Sketching up the world: in situ authoring for mobile
  augmented reality,'' \emph{Personal and ubiquitous computing}, vol.~16,
  no.~6, pp. 623--630, 2012.

\bibitem{langlotz2011robust}
T.~Langlotz, C.~Degendorfer, A.~Mulloni, G.~Schall, G.~Reitmayr, and
  D.~Schmalstieg, ``Robust detection and tracking of annotations for outdoor
  augmented reality browsing,'' \emph{Computers \& graphics}, vol.~35, no.~4,
  pp. 831--840, 2011.

\bibitem{macintyre2002estimating}
B.~MacIntyre, E.~M. Coelho, and S.~J. Julier, ``Estimating and adapting to
  registration errors in augmented reality systems,'' in \emph{Proceedings IEEE
  Virtual Reality 2002}.\hskip 1em plus 0.5em minus 0.4em\relax IEEE, 2002, pp.
  73--80.

\bibitem{feiner1997touring}
S.~Feiner, B.~MacIntyre, T.~H{\"o}llerer, and A.~Webster, ``A touring machine:
  Prototyping 3d mobile augmented reality systems for exploring the urban
  environment,'' \emph{Personal Technologies}, vol.~1, no.~4, pp. 208--217,
  1997.

\bibitem{lee2018interaction}
L.-H. Lee and P.~Hui, ``Interaction methods for smart glasses: A survey,''
  \emph{IEEE access}, vol.~6, pp. 28\,712--28\,732, 2018.

\bibitem{wacker2020heatmaps}
P.~Wacker, A.~Wagner, S.~Voelker, and J.~Borchers, ``Heatmaps, shadows,
  bubbles, rays: Comparing mid-air pen position visualizations in handheld
  ar,'' in \emph{Proceedings of the 2020 CHI Conference on Human Factors in
  Computing Systems}, 2020, pp. 1--11.

\bibitem{xie2016large}
C.~Xie, Y.~Kameda, K.~Suzuki, and I.~Kitahara, ``Large scale interactive ar
  display based on a projector-camera system,'' in \emph{Proceedings of the
  2016 Symposium on Spatial User Interaction}, 2016, pp. 179--179.

\bibitem{roo2017inner}
J.~S. Roo, R.~Gervais, J.~Frey, and M.~Hachet, ``Inner garden: Connecting inner
  states to a mixed reality sandbox for mindfulness,'' in \emph{Proceedings of
  the 2017 CHI Conference on Human Factors in Computing Systems}, 2017, pp.
  1459--1470.

\bibitem{hartmann2020aar}
J.~Hartmann, Y.-T. Yeh, and D.~Vogel, ``Aar: Augmenting a wearable augmented
  reality display with an actuated head-mounted projector,'' in
  \emph{Proceedings of the 33rd Annual ACM Symposium on User Interface Software
  and Technology}, 2020, pp. 445--458.

\bibitem{chaturvedi2019peripheral}
I.~Chaturvedi, F.~H. Bijarbooneh, T.~Braud, and P.~Hui, ``Peripheral vision: a
  new killer app for smart glasses,'' in \emph{Proceedings of the 24th
  International Conference on Intelligent User Interfaces}, 2019, pp. 625--636.

\bibitem{milgram1994taxonomy}
P.~Milgram and F.~Kishino, ``A taxonomy of mixed reality visual displays,''
  \emph{IEICE TRANSACTIONS on Information and Systems}, vol.~77, no.~12, pp.
  1321--1329, 1994.

\bibitem{lopes2018adding}
P.~Lopes, S.~You, A.~Ion, and P.~Baudisch, ``Adding force feedback to mixed
  reality experiences and games using electrical muscle stimulation,'' in
  \emph{Proceedings of the 2018 chi conference on human factors in computing
  systems}, 2018, pp. 1--13.

\bibitem{reilly2015mapping}
D.~Reilly, A.~Echenique, A.~Wu, A.~Tang, and W.~K. Edwards, ``Mapping out work
  in a mixed reality project room,'' in \emph{Proceedings of the 33rd Annual
  ACM Conference on Human Factors in Computing Systems}, 2015, pp. 887--896.

\bibitem{ohta2015mixed}
M.~Ohta, S.~Nagano, H.~Niwa, and K.~Yamashita, ``Mixed-reality store on the
  other side of a tablet.'' in \emph{ISMAR}, 2015, pp. 192--193.

\bibitem{yue2017scenectrl}
Y.-T. Yue, Y.-L. Yang, G.~Ren, and W.~Wang, ``Scenectrl: Mixed reality
  enhancement via efficient scene editing,'' in \emph{Proceedings of the 30th
  annual ACM symposium on user interface software and technology}, 2017, pp.
  427--436.

\bibitem{lee2020towards}
L.-H. Lee, T.~Braud, S.~Hosio, and P.~Hui, ``Towards augmented reality-driven
  human-city interaction: Current research and future challenges,'' \emph{arXiv
  preprint arXiv:2007.09207}, 2020.

\bibitem{malinverni2017world}
L.~Malinverni, J.~Maya, M.-M. Schaper, and N.~Pares, ``The world-as-support:
  Embodied exploration, understanding and meaning-making of the augmented
  world,'' in \emph{Proceedings of the 2017 CHI Conference on Human Factors in
  Computing Systems}, 2017, pp. 5132--5144.

\bibitem{gardony2020eye}
A.~L. Gardony, R.~W. Lindeman, and T.~T. Bruny{\'e}, ``Eye-tracking for
  human-centered mixed reality: promises and challenges,'' in \emph{Optical
  Architectures for Displays and Sensing in Augmented, Virtual, and Mixed
  Reality (AR, VR, MR)}, vol. 11310.\hskip 1em plus 0.5em minus 0.4em\relax
  SPIE, 2020, pp. 230--247.

\bibitem{gustavsson2021implementation}
U.~Gustavsson, P.~Frenger, C.~Fager, T.~Eriksson, H.~Zirath, F.~Dielacher,
  C.~Studer, A.~P{\"a}rssinen, R.~Correia, J.~N. Matos \emph{et~al.},
  ``Implementation challenges and opportunities in beyond-5g and 6g
  communication,'' \emph{IEEE Journal of Microwaves}, vol.~1, no.~1, pp.
  86--100, 2021.

\bibitem{giordani2020toward}
M.~Giordani, M.~Polese, M.~Mezzavilla, S.~Rangan, and M.~Zorzi, ``Toward 6g
  networks: Use cases and technologies,'' \emph{IEEE Communications Magazine},
  vol.~58, no.~3, pp. 55--61, 2020.

\bibitem{rao2018impact}
S.~K. Rao and R.~Prasad, ``Impact of 5g technologies on industry 4.0,''
  \emph{Wireless personal communications}, vol. 100, no.~1, pp. 145--159, 2018.

\bibitem{glisic2006advanced}
S.~Glisic and J.-P. Makela, ``Advanced wireless networks: 4g technologies,'' in
  \emph{2006 IEEE Ninth International Symposium on Spread Spectrum Techniques
  and Applications}.\hskip 1em plus 0.5em minus 0.4em\relax IEEE, 2006, pp.
  442--446.

\bibitem{saad2019vision}
W.~Saad, M.~Bennis, and M.~Chen, ``A vision of 6g wireless systems:
  Applications, trends, technologies, and open research problems,'' \emph{IEEE
  network}, vol.~34, no.~3, pp. 134--142, 2019.

\bibitem{shrestha2020high}
P.~Shrestha, M.~Guidry, B.~Romanczyk, N.~Hatui, C.~Wurm, A.~Krishna, S.~S.
  Pasayat, R.~R. Karnaty, S.~Keller, J.~F. Buckwalter \emph{et~al.}, ``High
  linearity and high gain performance of n-polar gan mis-hemt at 30 ghz,''
  \emph{IEEE Electron Device Letters}, vol.~41, no.~5, pp. 681--684, 2020.

\bibitem{Andy2021}
\BIBentryALTinterwordspacing
A.~Boxall and T.~Lacoma. What is 6g, how fast will it be, and when is it
  coming? [Online]. Available:
  \url{https://www.digitaltrends.com/mobile/what-is-6g/}
\BIBentrySTDinterwordspacing

\bibitem{fernandez2019review}
T.~M. Fernandez-Carames and P.~Fraga-Lamas, ``A review on the application of
  blockchain to the next generation of cybersecure industry 4.0 smart
  factories,'' \emph{Ieee Access}, vol.~7, pp. 45\,201--45\,218, 2019.

\bibitem{cannavo2020blockchain}
A.~Cannavo and F.~Lamberti, ``How blockchain, virtual reality, and augmented
  reality are converging, and why,'' \emph{IEEE Consumer Electronics Magazine},
  vol.~10, no.~5, pp. 6--13, 2020.

\bibitem{french2020interaction}
A.~M. French, M.~Risius, and J.~P. Shim, ``The interaction of virtual reality,
  blockchain, and 5g new radio: Disrupting business and society,''
  \emph{Communications of the Association for Information Systems}, vol.~46,
  no.~1, p.~25, 2020.

\bibitem{nguyen2020blockchain}
D.~C. Nguyen, P.~N. Pathirana, M.~Ding, and A.~Seneviratne, ``Blockchain for 5g
  and beyond networks: A state of the art survey,'' \emph{Journal of Network
  and Computer Applications}, vol. 166, p. 102693, 2020.

\bibitem{el2020block5gintell}
A.~El~Azzaoui, S.~K. Singh, Y.~Pan, and J.~H. Park, ``Block5gintell: Blockchain
  for ai-enabled 5g networks,'' \emph{IEEE Access}, vol.~8, pp.
  145\,918--145\,935, 2020.

\bibitem{haddad2020blockchain}
Z.~Haddad, M.~M. Fouda, M.~Mahmoud, and M.~Abdallah, ``Blockchain-based
  authentication for 5g networks,'' in \emph{2020 IEEE International Conference
  on Informatics, IoT, and Enabling Technologies (ICIoT)}.\hskip 1em plus 0.5em
  minus 0.4em\relax IEEE, 2020, pp. 189--194.

\bibitem{Floridi2020}
L.~Floridi and M.~Chiriatti, ``Gpt-3: Its nature, scope, limits, and
  consequences,'' \emph{Minds and Machines}, vol.~30, pp. 681--694, 2020.

\bibitem{GPT3}
\BIBentryALTinterwordspacing
T.~Brown, B.~Mann, N.~Ryder, M.~Subbiah, J.~D. Kaplan, P.~Dhariwal,
  A.~Neelakantan, P.~Shyam, G.~Sastry, A.~Askell, S.~Agarwal, A.~Herbert-Voss,
  G.~Krueger, T.~Henighan, R.~Child, A.~Ramesh, D.~Ziegler, J.~Wu, C.~Winter,
  C.~Hesse, M.~Chen, E.~Sigler, M.~Litwin, S.~Gray, B.~Chess, J.~Clark,
  C.~Berner, S.~McCandlish, A.~Radford, I.~Sutskever, and D.~Amodei, ``Language
  models are few shot learners,'' in \emph{33rd Advances in Neural Information
  Processing Systems (NeurIPS)}, 2020, pp. 1--25. [Online]. Available:
  \url{https://proceedings.neurips.cc/paper/2020/hash/1457c0d6bfcb4967418bfb8ac142f64a-Abstract.html}
\BIBentrySTDinterwordspacing

\bibitem{DALLE1}
\BIBentryALTinterwordspacing
A.~Radford, J.~W. Kim, C.~Hallacy, A.~Ramesh, G.~Goh, S.~Agarwal, G.~Sastry,
  A.~Askell, P.~Mishkin, J.~Clark, G.~Krueger, and I.~Sutskever, ``Learning
  transferable visual models from natural language supervision,'' in \emph{38th
  International Conference on Machine Learning (ICML)}, 2021, pp. 8748--8763.
  [Online]. Available: \url{http://proceedings.mlr.press/v139/radford21a}
\BIBentrySTDinterwordspacing

\bibitem{DALLE2}
\BIBentryALTinterwordspacing
A.~Ramesh, M.~Pavlov, G.~Goh, S.~Gray, C.~Voss, A.~Radford, M.~Chen, and
  I.~Sutskever, ``Zero-shot text-to-image generation,'' in \emph{38th
  International Conference on Machine Learning (ICML)}, 2021, pp. 8821--8831.
  [Online]. Available: \url{https://proceedings.mlr.press/v139/ramesh21a.html}
\BIBentrySTDinterwordspacing

\bibitem{GauGAN}
T.~Park, M.-Y. Liu, T.-C. Wang, and J.-Y. Zhu, ``Gaugan: Semantic image
  synthesis with spatially adaptive normalization,'' in \emph{ACM SIGGRAPH 2019
  Real-Time Live}, 2019, pp. 1--1.

\bibitem{bishop_2016}
C.~M. BISHOP, \emph{Pattern recognition and machine learning}.\hskip 1em plus
  0.5em minus 0.4em\relax SPRINGER-VERLAG NEW YORK, 2016.

\bibitem{francois_2018}
V.~Francois-Lavet, P.~Henderson, I.~Riashat, M.~G. Bellemare, and J.~Pineau,
  \emph{An introduction to Deep Reinforcement Learning}.\hskip 1em plus 0.5em
  minus 0.4em\relax Now Foundations and Trends, 2018.

\bibitem{data2vec_2022}
A.~Baevski, W.-N. Hsu, Q.~Xu, A.~Babu, J.~Gu, and M.~Auli, ``data2vec: A
  general framework for self-supervised learning in speech, vision, and
  language,'' \emph{arXiV, preprint}, pp. 1--13, 2022.

\bibitem{FCN2015}
J.~Long, E.~Shelhamer, and T.~Darrell, ``Fully convolutional networks for
  semantic segmentation,'' \emph{IEEE Transations on Pattern Analysis and
  Machine Intelligence}, vol.~39, no.~4, pp. 640--651, 2017.

\bibitem{UPPSU2018}
T.~Xiao, Y.~Liu, B.~Zhou, Y.~Jiang, and J.~Sun, ``Unified perceptual parsing
  for scene understanding,'' in \emph{European Conference on Computer Vision},
  2018, pp. 432--448.

\bibitem{BiSeNet2018}
C.~Yu, J.~Wang, C.~Peng, C.~Gao, G.~Yu, and N.~Sang, ``Bisenet: Bilateral
  segmentation network for real-time semantic segmentation,'' in \emph{European
  Conference on Computer Vision}, 2018, pp. 334--349.

\bibitem{FPN2019}
A.~kirillov, R.~Girshick, K.~He, and P.~Dollar, ``Panoptic feature pyramid
  networks,'' in \emph{Proceedings of the IEEE Conference on Computer Vision
  and Pattern Recognition (CVPR)}, 2019, pp. 6399--6408.

\bibitem{SFNet2019}
X.~Li, A.~You, Z.~Zhu, H.~Zhao, M.~Yang, K.~Yang, S.~Tan, and Y.~Tong,
  ``Semantic flow for fast and accurate scene parsing,'' in \emph{European
  Conference on Computer Vision}, 2020, pp. 775--793.

\bibitem{SegFormer2021}
E.~Xie, W.~Wang, Z.~Yu, A.~Anandkumar, J.~M. Alvarez, and P.~Luo, ``Segformer:
  Simple and efficient design for semantic segmenattion with transformers,'' in
  \emph{35th Conference on Neural Information Processing Systems (NeurIPS)},
  2021, pp. 1--14.

\bibitem{FaPN2021}
S.~Huang, Z.~Lu, R.~Cheng, and C.~He, ``Fapn: Feature aligned pyramid network
  for dense image prediction,'' in \emph{Proceedings of the IEEE International
  Conference on Computer Vision (ICCV)}, 2021, pp. 864--873.

\bibitem{CondNet2021}
C.~Yu, Y.~Shao, C.~Gao, and N.~Sang, ``Condnet: Conditional classifier for
  scene segmentation,'' \emph{IEEE Signal Processing Letters}, vol.~28, pp.
  758--762, 2021.

\bibitem{Lawin2022}
H.~Yan, C.~Zhang, and M.~Wu, ``Lawin transformer: Improving semantic
  segmentation transfoermer with multiscale representations via large window
  attention,'' \emph{arXiv}, pp. 1--11, 2022.

\bibitem{abramson_johnson_2020}
D.~I. Abramson and J.~Johnson, ``Creating a conversational chatbot of a
  specific person,'' Dec 2020.

\bibitem{ZooBuilder2020}
A.~S. Fangbemi, Y.~F. Lu, M.~Y. Xu, X.~W. Luo, A.~Rolland, and C.~Raissi,
  ``Zoobuilder: 2d and 3d pose estimation for quadrupeds using synthetic
  data,'' \emph{ACM Siggraph}, vol.~39, pp. 1--2, 2020.

\bibitem{OpenPose2021}
Z.~Cao, G.~Hidalgo, T.~Simon, S.-E. Wei, and Y.~Sheikh, ``Openpose: Realtime
  multiperson 2d pose estimation using part affinity fields,'' \emph{IEEE
  Transactions on Pattern Analysis and Machine Intelligence}, vol.~43, pp.
  172--186, 2021.

\bibitem{PoseEstimator2018}
M.~R.~I. Hossain and J.~J. Little, ``Exploiting temporal information for 3d
  human pose estimation,'' in \emph{European Conference on Computer Vision
  (ECCV)}, 2018, pp. 69--86.

\bibitem{StyleGAN}
T.~Karras, S.~Laine, M.~Aittala, J.~Hellsten, and J.~Lehtinen, ``Analyzing and
  improving the image quality of stylegan,'' in \emph{Proceedings of the IEEE
  Conference on Computer Vision and Pattern Recognition (CVPR)}, 2020, pp.
  8110--8119.

\bibitem{She2021}
C.~She, C.~Sun, Z.~Gu, Y.~Li, C.~Yang, H.~Poor, Vincent, and B.~Vucetic, ``A
  tutorial on ultrareliable and low-latency communications in 6g: Integrating
  domain knowledge into deep learning,'' \emph{Proceedings of the IEEE}, vol.
  109, no.~3, pp. 204--246, 2021.

\bibitem{Dev2022}
K.~Dev, S.~A. Khowaja, P.~K. Sharma, B.~S. Chowdhry, S.~Tanwar, and G.~Fortino,
  ``Ddi: A novel architecture for joint active user detection and iot device
  identification in grant-free noma systems of 6g and beyond networks,''
  \emph{IEEE Internet of Things Journal}, vol.~9, no.~4, pp. 2906--2917, 2022.

\bibitem{She2020}
C.~She, R.~Dong, Z.~Gu, Y.~Li, W.~Hardjawana, C.~Yang, L.~Song, and B.~Vucetic,
  ``Deep learning for ultra-reliable and low-latency communication in 6g
  neworks,'' \emph{IEEE Network}, vol.~34, no.~5, pp. 219--225, 2020.

\bibitem{Alsenwi2021}
M.~Alsenwi, N.~H. Tran, M.~Bennis, S.~R. Pandey, A.~K. Bairagi, and C.~S.~a.
  Hong, ``Intelligence resource slicing for embb and urllc coexistence in 5g
  and beyond: A deep reinforcement learning based approach,'' \emph{IEEE
  Transactions on Wireless Communications}, vol.~20, no.~7, pp. 4585--4600,
  2021.

\bibitem{Tunze2020}
G.~B. Tunze, T.~Huynh-The, J.-M. Lee, and D.-S. Kim, ``Sparsely connected cnn
  for efficient automatic modulation recognition,'' \emph{IEEE Transactions on
  Vehicular Technology}, vol.~69, no.~12, pp. 15\,557--15\,568, 2020.

\bibitem{Huynh2020}
T.~Huynh-The, C.-H. Hua, Q.-V. Pham, and D.-S. Kim, ``Mcnet: An efficient cnn
  architecture for robust automation modulation classification,'' \emph{IEEE
  Communication Letters}, vol.~24, no.~4, pp. 811--815, 2020.

\bibitem{Luo2020}
C.~Luo, J.~Ji, Q.~Wang, X.~Chen, and P.~Li, ``Channel state information
  prediction for 5g wireless communications: A deep learning approach,''
  \emph{IEEE Transactions on Network Science and Engineering}, vol.~7, no.~1,
  pp. 227--236, 2020.

\bibitem{Guo2019}
S.~Guo, Y.~Lin, S.~Li, Z.~Chen, and H.~Wang, ``Deep spatial-temporal 3d
  convolutional neural networks for traffic data forecasting,'' \emph{IEEE
  Transactions on Intelligent Transportation Systems}, vol.~20, no.~10, pp.
  3913--3926, 2019.

\bibitem{wang2020}
X.~Wang, Y.~Han, V.~C.~M. Leung, D.~Niyato, X.~Yan, and X.~Chen, ``Convergence
  of edge computing and deep learning: A comprehensive survey,'' \emph{IEEE
  Communication Surveys and Tutorials}, vol.~22, no.~2, pp. 869--904, 2020.

\bibitem{IIFNet}
S.~A. Khowaja, K.~Dev, P.~Khuwaja, Q.-V. Pham, N.~M.~F. Qureshi, P.~Bellavista,
  and M.~Magarini, ``Iifnet: A fusion-based intelligent service for noisy
  preamble detection in 6g,'' \emph{IEEE Network}, vol.~36, no.~3, pp. 48--54,
  2022.

\bibitem{DBFL}
S.~A. Khowaja, K.~Dev, P.~Khuwaja, and P.~Bellavista, ``Toward energy-efficient
  distributed federated learning for 6g networks,'' \emph{IEEE Wireless
  Communications}, vol.~28, no.~6, pp. 34--40, 2021.

\bibitem{TeraPipe}
Z.~Li, S.~Zhuang, S.~Guo, D.~Zhuo, H.~Zhang, D.~Song, and I.~Stoica,
  ``Terapipe: Token-level pipeline parallelism for training large-scale
  language models,'' \emph{Proceedings of Machine Learning Research}, vol. 139,
  pp. 6543--6552, 2021.

\bibitem{HioTSP}
S.~A. Khowaja, A.~G. Prabono, F.~Setiawan, B.~N. Yahya, and S.-L. Lee,
  ``Contextual activity based healthcare internet of things, services, and
  people (hiotsp): An architectural framework for healthcare monitoring using
  wearable sensors,'' \emph{Computer Networks}, vol. 145, pp. 190--206, 2018.

\bibitem{FinTech}
P.~Khuwaja, S.~A. Khowaja, and K.~Dev, ``Adversarial learning networks for
  fintech applications using heterogeneous data sources,'' \emph{IEEE Internet
  of Things Journal}, vol. Early Access, pp. 1--8, 2021.

\bibitem{Req6G1}
D.~G. Morin, P.~Perez, and A.~G. Armada, ``Toward the distributed
  implementation of immersive augmented reality architectures on 5g networks,''
  \emph{IEEE Communications Magazine}, vol.~60, no.~2, pp. 46--52, 2022.

\bibitem{Req6G2}
F.~Hu, Y.~Deng, W.~Saad, M.~Bennis, and A.~H. Aghvami, ``Cellular-connected
  wireless virtual reality: Requirements, challenges, and solutions,''
  \emph{IEEE Communications Magazine}, vol.~58, no.~5, pp. 105--111, 2020.

\bibitem{siriwardhana2021survey}
Y.~Siriwardhana, P.~Porambage, M.~Liyanage, and M.~Ylianttila, ``A survey on
  mobile augmented reality with 5g mobile edge computing: architectures,
  applications, and technical aspects,'' \emph{IEEE Communications Surveys \&
  Tutorials}, vol.~23, no.~2, pp. 1160--1192, 2021.

\bibitem{sukhmani2018edge}
S.~Sukhmani, M.~Sadeghi, M.~Erol-Kantarci, and A.~El~Saddik, ``Edge caching and
  computing in 5g for mobile ar/vr and tactile internet,'' \emph{IEEE
  MultiMedia}, vol.~26, no.~1, pp. 21--30, 2018.

\bibitem{li2018network}
R.~Li, ``Network 2030: Market drivers and prospects,'' in \emph{Proc. 1st
  International Telecommunication Union Workshop on Network}, vol. 2030, 2018.

\bibitem{gotsch2018telehuman2}
D.~Gotsch, X.~Zhang, T.~Merritt, and R.~Vertegaal, ``Telehuman2: A cylindrical
  light field teleconferencing system for life-size 3d human telepresence.'' in
  \emph{CHI}, vol.~18, 2018, p. 552.

\bibitem{yang2018digital}
B.~Yang, Z.~Yu, J.~Lan, R.~Zhang, J.~Zhou, and W.~Hong, ``Digital
  beamforming-based massive mimo transceiver for 5g millimeter-wave
  communications,'' \emph{IEEE Transactions on Microwave Theory and
  Techniques}, vol.~66, no.~7, pp. 3403--3418, 2018.

\bibitem{yastrebova2018future}
A.~Yastrebova, R.~Kirichek, Y.~Koucheryavy, A.~Borodin, and A.~Koucheryavy,
  ``Future networks 2030: Architecture \& requirements,'' in \emph{2018 10th
  International Congress on Ultra Modern Telecommunications and Control Systems
  and Workshops (ICUMT)}.\hskip 1em plus 0.5em minus 0.4em\relax IEEE, 2018,
  pp. 1--8.

\bibitem{erel2019road}
M.~Erel-{\"O}z{\c{c}}evik and B.~Canberk, ``Road to 5g reduced-latency: A
  software defined handover model for embb services,'' \emph{IEEE Transactions
  on Vehicular Technology}, vol.~68, no.~8, pp. 8133--8144, 2019.

\bibitem{zhang2019will}
M.~Zhang, M.~Polese, M.~Mezzavilla, J.~Zhu, S.~Rangan, S.~Panwar, and M.~Zorzi,
  ``Will tcp work in mmwave 5g cellular networks?'' \emph{IEEE Communications
  Magazine}, vol.~57, no.~1, pp. 65--71, 2019.

\bibitem{milovanovic20215g}
D.~Milovanovic, Z.~Bojkovic, M.~Indoonundon, and T.~P. Fowdur, ``5g low-latency
  communication in virtual reality services: Performance requirements and
  promising solutions,'' \emph{WSEAS Transactions on Communications}, vol.~20,
  pp. 77--81, 2021.

\bibitem{chen2022standardization}
W.~Chen, J.~Montojo, J.~Lee, M.~Shafi, and Y.~Kim, ``The standardization of
  5g-advanced in 3gpp,'' \emph{IEEE Communications Magazine}, 2022.

\bibitem{yeh2022perspectives}
C.~Yeh, G.~Do~Jo, Y.-J. Ko, and H.~K. Chung, ``Perspectives on 6g wireless
  communications,'' \emph{ICT Express}, 2022.

\bibitem{viswanathan2020communications}
H.~Viswanathan and P.~E. Mogensen, ``Communications in the 6g era,'' \emph{IEEE
  Access}, vol.~8, pp. 57\,063--57\,074, 2020.

\bibitem{docomo2021white}
N.~DOCOMO, ``White paper: 5g evolution and 6g (version 3.0),'' \emph{NTT DOCOMO
  White Paper}, 2021.

\bibitem{zhang20196g}
Z.~Zhang, Y.~Xiao, Z.~Ma, M.~Xiao, Z.~Ding, X.~Lei, G.~K. Karagiannidis, and
  P.~Fan, ``6g wireless networks: Vision, requirements, architecture, and key
  technologies,'' \emph{IEEE Vehicular Technology Magazine}, vol.~14, no.~3,
  pp. 28--41, 2019.

\bibitem{yang2018communication}
X.~Yang, Z.~Chen, K.~Li, Y.~Sun, N.~Liu, W.~Xie, and Y.~Zhao,
  ``Communication-constrained mobile edge computing systems for wireless
  virtual reality: Scheduling and tradeoff,'' \emph{IEEE Access}, vol.~6, pp.
  16\,665--16\,677, 2018.

\bibitem{chaccour2022seven}
C.~Chaccour, M.~N. Soorki, W.~Saad, M.~Bennis, P.~Popovski, and M.~Debbah,
  ``Seven defining features of terahertz (thz) wireless systems: A fellowship
  of communication and sensing,'' \emph{IEEE Communications Surveys \&
  Tutorials}, vol.~24, no.~2, pp. 967--993, 2022.

\bibitem{petrov2022extended}
V.~Petrov, M.~Gapeyenko, S.~Paris, A.~Marcano, and K.~I. Pedersen, ``Extended
  reality (xr) over 5g and 5g-advanced new radio: Standardization,
  applications, and trends,'' \emph{arXiv preprint arXiv:2203.02242}, 2022.

\bibitem{kim2016performance}
J.~Kim, S.-C. Kwon, and G.~Choi, ``Performance of video streaming in
  infrastructure-to-vehicle telematic platforms with 60-ghz radiation and ieee
  802.11 ad baseband,'' \emph{IEEE Transactions on Vehicular Technology},
  vol.~65, no.~12, pp. 10\,111--10\,115, 2016.

\bibitem{kim2017strategic}
J.~Kim, J.-J. Lee, and W.~Lee, ``Strategic control of 60 ghz millimeter-wave
  high-speed wireless links for distributed virtual reality platforms,''
  \emph{Mobile Information Systems}, vol. 2017, 2017.

\bibitem{leng2019energy}
Y.~Leng, C.-C. Chen, Q.~Sun, J.~Huang, and Y.~Zhu, ``Energy-efficient video
  processing for virtual reality,'' in \emph{Proceedings of the 46th
  International Symposium on Computer Architecture}, 2019, pp. 91--103.

\bibitem{yang2019joint}
X.~Yang, Z.~Fei, J.~Zheng, N.~Zhang, and A.~Anpalagan, ``Joint multi-user
  computation offloading and data caching for hybrid mobile cloud/edge
  computing,'' \emph{IEEE Transactions on Vehicular Technology}, vol.~68,
  no.~11, pp. 11\,018--11\,030, 2019.

\bibitem{liu2018mec}
Y.~Liu, J.~Liu, A.~Argyriou, and S.~Ci, ``Mec-assisted panoramic vr video
  streaming over millimeter wave mobile networks,'' \emph{IEEE Transactions on
  Multimedia}, vol.~21, no.~5, pp. 1302--1316, 2018.

\bibitem{qian2016optimizing}
F.~Qian, L.~Ji, B.~Han, and V.~Gopalakrishnan, ``Optimizing 360 video delivery
  over cellular networks,'' in \emph{Proceedings of the 5th Workshop on All
  Things Cellular: Operations, Applications and Challenges}, 2016, pp. 1--6.

\bibitem{mangiante2017vr}
S.~Mangiante, G.~Klas, A.~Navon, Z.~GuanHua, J.~Ran, and M.~D. Silva, ``Vr is
  on the edge: How to deliver 360 videos in mobile networks,'' in
  \emph{Proceedings of the Workshop on Virtual Reality and Augmented Reality
  Network}, 2017, pp. 30--35.

\bibitem{sun2019communications}
Y.~Sun, Z.~Chen, M.~Tao, and H.~Liu, ``Communications, caching, and computing
  for mobile virtual reality: Modeling and tradeoff,'' \emph{IEEE Transactions
  on Communications}, vol.~67, no.~11, pp. 7573--7586, 2019.

\bibitem{zhao2020optimal}
L.~Zhao, Y.~Cui, C.~Guo, and Z.~Liu, ``Optimal streaming of 360 vr videos with
  perfect, imperfect and unknown fov viewing probabilities,'' in \emph{GLOBECOM
  2020-2020 IEEE Global Communications Conference}.\hskip 1em plus 0.5em minus
  0.4em\relax IEEE, 2020, pp. 1--6.

\bibitem{long2020optimal}
K.~Long, Y.~Cui, C.~Ye, and Z.~Liu, ``Optimal wireless streaming of
  multi-quality 360 vr video by exploiting natural, relative
  smoothness-enabled, and transcoding-enabled multicast opportunities,''
  \emph{IEEE Transactions on Multimedia}, vol.~23, pp. 3670--3683, 2020.

\bibitem{gupta2019millimeter}
S.~Gupta, J.~Chakareski, and P.~Popovski, ``Millimeter wave meets edge
  computing for mobile vr with high-fidelity 8k scalable 360 video,'' in
  \emph{2019 IEEE 21st International Workshop on Multimedia Signal Processing
  (MMSP)}.\hskip 1em plus 0.5em minus 0.4em\relax IEEE, 2019, pp. 1--6.

\bibitem{du2020mec}
J.~Du, F.~R. Yu, G.~Lu, J.~Wang, J.~Jiang, and X.~Chu, ``Mec-assisted immersive
  vr video streaming over terahertz wireless networks: A deep reinforcement
  learning approach,'' \emph{IEEE Internet of Things Journal}, vol.~7, no.~10,
  pp. 9517--9529, 2020.

\bibitem{chen2019liquid}
M.~Chen, W.~Saad, and C.~Yin, ``Liquid state based transfer learning for 360
  image transmission in wireless vr networks,'' in \emph{ICC 2019-2019 IEEE
  International Conference on Communications (ICC)}.\hskip 1em plus 0.5em minus
  0.4em\relax IEEE, 2019, pp. 1--6.

\bibitem{han2018propagation}
C.~Han and Y.~Chen, ``Propagation modeling for wireless communications in the
  terahertz band,'' \emph{IEEE Communications Magazine}, vol.~56, no.~6, pp.
  96--101, 2018.

\bibitem{liu2021learning}
X.~Liu, Y.~Deng, C.~Han, and M.~Di~Renzo, ``Learning-based prediction,
  rendering and transmission for interactive virtual reality in ris-assisted
  terahertz networks,'' \emph{IEEE Journal on Selected Areas in
  Communications}, vol.~40, no.~2, pp. 710--724, 2021.

\bibitem{chowdhury20206g}
M.~Z. Chowdhury, M.~Shahjalal, S.~Ahmed, and Y.~M. Jang, ``6g wireless
  communication systems: Applications, requirements, technologies, challenges,
  and research directions,'' \emph{IEEE Open Journal of the Communications
  Society}, vol.~1, pp. 957--975, 2020.

\bibitem{de2021survey}
C.~De~Alwis, A.~Kalla, Q.-V. Pham, P.~Kumar, K.~Dev, W.-J. Hwang, and
  M.~Liyanage, ``Survey on 6g frontiers: Trends, applications, requirements,
  technologies and future research,'' \emph{IEEE Open Journal of the
  Communications Society}, vol.~2, pp. 836--886, 2021.

\bibitem{park2020extreme}
J.~Park, S.~Samarakoon, H.~Shiri, M.~K. Abdel-Aziz, T.~Nishio, A.~Elgabli, and
  M.~Bennis, ``Extreme urllc: Vision, challenges, and key enablers,''
  \emph{arXiv preprint arXiv:2001.09683}, 2020.

\bibitem{wallace2021high}
J.~Wallace and A.~Valdivia, ``A high-performance 5g/6g infrastructure for
  augmented, virtual, and extended reality,'' in \emph{2021 International
  Conference on Computational Science and Computational Intelligence
  (CSCI)}.\hskip 1em plus 0.5em minus 0.4em\relax IEEE, 2021, pp. 1291--1296.

\bibitem{li2020enabling}
X.~Li, W.~Feng, J.~Wang, Y.~Chen, N.~Ge, and C.-X. Wang, ``Enabling 5g on the
  ocean: A hybrid satellite-uav-terrestrial network solution,'' \emph{IEEE
  Wireless Communications}, vol.~27, no.~6, pp. 116--121, 2020.

\bibitem{mourtzis2017augmented}
D.~Mourtzis, V.~Zogopoulos, and E.~Vlachou, ``Augmented reality application to
  support remote maintenance as a service in the robotics industry,''
  \emph{Procedia Cirp}, vol.~63, pp. 46--51, 2017.

\bibitem{cheng2018industrial}
J.~Cheng, W.~Chen, F.~Tao, and C.-L. Lin, ``Industrial iot in 5g environment
  towards smart manufacturing,'' \emph{Journal of Industrial Information
  Integration}, vol.~10, pp. 10--19, 2018.

\bibitem{erol2018caching}
M.~Erol-Kantarci and S.~Sukhmani, ``Caching and computing at the edge for
  mobile augmented reality and virtual reality (ar/vr) in 5g,'' \emph{Ad Hoc
  Networks}, pp. 169--177, 2018.

\bibitem{ren2020edge}
P.~Ren, X.~Qiao, Y.~Huang, L.~Liu, C.~Pu, S.~Dustdar, and J.-L. Chen, ``Edge ar
  x5: An edge-assisted multi-user collaborative framework for mobile web
  augmented reality in 5g and beyond,'' \emph{IEEE Transactions on Cloud
  Computing}, 2020.

\bibitem{zhou20215g}
P.~Zhou, B.~Finley, X.~Li, S.~Tarkoma, J.~Kangasharju, M.~Ammar, and P.~Hui,
  ``5g mec computation handoff for mobile augmented reality,'' \emph{arXiv
  preprint arXiv:2101.00256}, 2021.

\bibitem{ng2021unified}
W.~C. Ng, W.~Y.~B. Lim, J.~S. Ng, Z.~Xiong, D.~Niyato, and C.~Miao, ``Unified
  resource allocation framework for the edge intelligence-enabled metaverse,''
  \emph{arXiv preprint arXiv:2110.14325}, 2021.

\bibitem{ren2022distributed}
P.~Ren, L.~Liu, X.~Qiao, and J.~Chen, ``Distributed edge system orchestration
  for web-based mobile augmented reality services,'' \emph{IEEE Transactions on
  Services Computing}, no.~01, pp. 1--15, 2022.

\bibitem{zhang2022sear}
W.~Zhang, B.~Han, and P.~Hui, ``Sear: Scaling experiences in multi-user
  augmented reality,'' \emph{IEEE Transactions on Visualization and Computer
  Graphics}, vol.~28, no.~5, pp. 1982--1992, 2022.

\bibitem{van2022edge}
D.~Van~Huynh, S.~R. Khosravirad, A.~Masaracchia, O.~A. Dobre, and T.~Q. Duong,
  ``Edge intelligence-based ultra-reliable and low-latency communications for
  digital twin-enabled metaverse,'' \emph{IEEE Wireless Communications
  Letters}, 2022.

\bibitem{kang2022blockchain}
J.~Kang, D.~Ye, J.~Nie, J.~Xiao, X.~Deng, S.~Wang, Z.~Xiong, R.~Yu, and
  D.~Niyato, ``Blockchain-based federated learning for industrial metaverses:
  Incentive scheme with optimal aoi,'' \emph{arXiv preprint arXiv:2206.07384},
  2022.

\bibitem{alam2015intelligent}
M.~F. Alam, S.~Katsikas, and S.~Hadjiefthymiades, ``An intelligent and modular
  sensing system for augmented reality application,'' in \emph{2015 9th
  International Conference on Sensing Technology (ICST)}.\hskip 1em plus 0.5em
  minus 0.4em\relax IEEE, 2015, pp. 850--855.

\bibitem{kim2020motion}
S.~Kim and J.-H. Yun, ``Motion-aware interplay between wigig and wifi for
  wireless virtual reality,'' \emph{Sensors}, vol.~20, no.~23, p. 6782, 2020.

\bibitem{sugimoto2022cloud}
M.~Sugimoto, ``Cloud xr (extended reality: Virtual reality, augmented reality,
  mixed reality) and 5g mobile communication system for medical image-guided
  holographic surgery and telemedicine,'' in \emph{Multidisciplinary
  Computational Anatomy}.\hskip 1em plus 0.5em minus 0.4em\relax Springer,
  2022, pp. 381--387.

\bibitem{han2022dynamic}
Y.~Han, D.~Niyato, C.~Leung, C.~Miao, and D.~I. Kim, ``A dynamic resource
  allocation framework for synchronizing metaverse with iot service and data,''
  in \emph{ICC 2022-IEEE International Conference on Communications}.\hskip 1em
  plus 0.5em minus 0.4em\relax IEEE, 2022, pp. 1196--1201.

\bibitem{zhang2018towards}
Q.~Zhang, J.~Liu, and G.~Zhao, ``Towards 5g enabled tactile robotic
  telesurgery,'' \emph{arXiv preprint arXiv:1803.03586}, 2018.

\bibitem{gupta2019tactile}
R.~Gupta, S.~Tanwar, S.~Tyagi, and N.~Kumar, ``Tactile-internet-based
  telesurgery system for healthcare 4.0: An architecture, research challenges,
  and future directions,'' \emph{IEEE Network}, vol.~33, no.~6, pp. 22--29,
  2019.

\bibitem{kusuma2021enabling}
H.~M. Kusuma, V.~K. Shukla, and S.~Gupta, ``Enabling vr/ar and tactile through
  5g network,'' in \emph{2021 International Conference on Communication
  information and Computing Technology (ICCICT)}.\hskip 1em plus 0.5em minus
  0.4em\relax IEEE, 2021, pp. 1--6.

\bibitem{minopoulos2021efficient}
G.~Minopoulos, K.~E. Psannis, S.~Goudos, S.~Nikolaidis, G.~Kokkonis, and
  Y.~Ishibashi, ``Efficient integration of xr with haptic feedback and 5g
  networks,'' in \emph{2021 IEEE 9th International Conference on Information,
  Communication and Networks (ICICN)}.\hskip 1em plus 0.5em minus 0.4em\relax
  IEEE, 2021, pp. 240--244.

\bibitem{sartipi2019decentralized}
K.~Sartipi, R.~C. DuToit, C.~B. Cobar, and S.~I. Roumeliotis, ``Decentralized
  visual-inertial localization and mapping on mobile devices for augmented
  reality,'' in \emph{2019 IEEE/RSJ International Conference on Intelligent
  Robots and Systems (IROS)}.\hskip 1em plus 0.5em minus 0.4em\relax IEEE,
  2019, pp. 2145--2152.

\bibitem{guo2020adaptive}
F.~Guo, F.~R. Yu, H.~Zhang, H.~Ji, V.~C. Leung, and X.~Li, ``An adaptive
  wireless virtual reality framework in future wireless networks: A distributed
  learning approach,'' \emph{IEEE Transactions on Vehicular Technology},
  vol.~69, no.~8, pp. 8514--8528, 2020.

\bibitem{chen2020federated}
D.~Chen, L.~J. Xie, B.~Kim, L.~Wang, C.~S. Hong, L.-C. Wang, and Z.~Han,
  ``Federated learning based mobile edge computing for augmented reality
  applications,'' in \emph{2020 international conference on computing,
  networking and communications (ICNC)}.\hskip 1em plus 0.5em minus 0.4em\relax
  IEEE, 2020, pp. 767--773.

\bibitem{de2022fedlens}
D.~De, ``Fedlens: federated learning-based privacy-preserving mobile
  crowdsensing for virtual tourism,'' \emph{Innovations in Systems and Software
  Engineering}, pp. 1--14, 2022.

\bibitem{chen2020joint}
X.~Chen and G.~Liu, ``Joint optimization of task offloading and resource
  allocation via deep reinforcement learning for augmented reality in mobile
  edge network,'' in \emph{2020 IEEE International Conference on Edge Computing
  (EDGE)}.\hskip 1em plus 0.5em minus 0.4em\relax IEEE, 2020, pp. 76--82.

\bibitem{lin2021task}
P.~Lin, Q.~Song, F.~R. Yu, D.~Wang, and L.~Guo, ``Task offloading for wireless
  vr-enabled medical treatment with blockchain security using collective
  reinforcement learning,'' \emph{IEEE Internet of Things Journal}, vol.~8,
  no.~21, pp. 15\,749--15\,761, 2021.

\bibitem{jiang2021reliable}
Y.~Jiang, J.~Kang, D.~Niyato, X.~Ge, Z.~Xiong, and C.~Miao, ``Reliable coded
  distributed computing for metaverse services: Coalition formation and
  incentive mechanism design,'' \emph{arXiv preprint arXiv:2111.10548}, 2021.

\bibitem{adeogun2020towards}
R.~Adeogun, G.~Berardinelli, P.~E. Mogensen, I.~Rodriguez, and M.~Razzaghpour,
  ``Towards 6g in-x subnetworks with sub-millisecond communication cycles and
  extreme reliability,'' \emph{IEEE Access}, vol.~8, pp. 110\,172--110\,188,
  2020.

\bibitem{benzaid2020ai}
C.~Benzaid and T.~Taleb, ``Ai-driven zero touch network and service management
  in 5g and beyond: Challenges and research directions,'' \emph{IEEE Network},
  vol.~34, no.~2, pp. 186--194, 2020.

\bibitem{rizwan2021zero}
A.~Rizwan, M.~Jaber, F.~Filali, A.~Imran, and A.~Abu-Dayya, ``A zero-touch
  network service management approach using ai-enabled cdr analysis,''
  \emph{IEEE Access}, vol.~9, pp. 157\,699--157\,714, 2021.

\bibitem{andrus2019zero}
B.-M. Andrus, S.~A. Sasu, T.~Szyrkowiec, A.~Autenrieth, M.~Chamania, J.~K.
  Fischer, and S.~Rasp, ``Zero-touch provisioning of distributed video
  analytics in a software-defined metro-haul network with p4 processing,'' in
  \emph{Optical Fiber Communication Conference}.\hskip 1em plus 0.5em minus
  0.4em\relax Optica Publishing Group, 2019, pp. M3Z--10.

\bibitem{xu2021wireless}
M.~Xu, D.~Niyato, J.~Kang, Z.~Xiong, C.~Miao, and D.~I. Kim, ``Wireless
  edge-empowered metaverse: A learning-based incentive mechanism for virtual
  reality,'' \emph{arXiv preprint arXiv:2111.03776}, 2021.

\bibitem{aijaz2018toward}
A.~Aijaz, Z.~Dawy, N.~Pappas, M.~Simsek, S.~Oteafy, and O.~Holland, ``Toward a
  tactile internet reference architecture: Vision and progress of the ieee
  p1918. 1 standard,'' \emph{arXiv preprint arXiv:1807.11915}, 2018.

\bibitem{baktir2017can}
A.~C. Baktir, A.~Ozgovde, and C.~Ersoy, ``How can edge computing benefit from
  software-defined networking: A survey, use cases, and future directions,''
  \emph{IEEE Communications Surveys \& Tutorials}, vol.~19, no.~4, pp.
  2359--2391, 2017.

\bibitem{rafique2020complementing}
W.~Rafique, L.~Qi, I.~Yaqoob, M.~Imran, R.~U. Rasool, and W.~Dou,
  ``Complementing iot services through software defined networking and edge
  computing: A comprehensive survey,'' \emph{IEEE Communications Surveys \&
  Tutorials}, vol.~22, no.~3, pp. 1761--1804, 2020.

\bibitem{sharma2020toward}
S.~K. Sharma, I.~Woungang, A.~Anpalagan, and S.~Chatzinotas, ``Toward tactile
  internet in beyond 5g era: recent advances, current issues, and future
  directions,'' \emph{Ieee Access}, vol.~8, pp. 56\,948--56\,991, 2020.

\bibitem{zhang2020challenges}
C.~Zhang, G.~He, Y.~Chen, P.~Zhu, and S.~Dou, ``Challenges and road ahead for
  wireless networks to serve immersive human centric applications,'' in
  \emph{2020 IEEE 92nd Vehicular Technology Conference (VTC2020-Fall)}.\hskip
  1em plus 0.5em minus 0.4em\relax IEEE, 2020, pp. 1--5.

\bibitem{maier2019towards}
M.~Maier and A.~Ebrahimzadeh, ``Towards immersive tactile internet experiences:
  Low-latency fiwi enhanced mobile networks with edge intelligence,''
  \emph{Journal of Optical Communications and Networking}, vol.~11, no.~4, pp.
  B10--B25, 2019.

\bibitem{chen2019deep}
M.~Chen, W.~Saad, and C.~Yin, ``Deep learning for 360 content transmission in
  uav-enabled virtual reality,'' in \emph{ICC 2019-2019 IEEE International
  Conference on Communications (ICC)}.\hskip 1em plus 0.5em minus 0.4em\relax
  IEEE, 2019, pp. 1--6.

\bibitem{el20203d}
A.~El~Saer, C.~Stentoumis, I.~Kalisperakis, L.~Grammatikopoulos, P.~Nomikou,
  and O.~Vlasopoulos, ``3d reconstruction and mesh optimization of underwater
  spaces for virtual reality,'' \emph{The International Archives of
  Photogrammetry, Remote Sensing and Spatial Information Sciences}, vol.~43,
  pp. 949--956, 2020.

\bibitem{zhang2021method}
J.~Zhang and F.~Wu, ``A method of offline reinforcement learning virtual
  reality satellite attitude control based on generative adversarial network,''
  \emph{Wireless Communications and Mobile Computing}, vol. 2021, 2021.

\bibitem{kasgari2018stochastic}
A.~T.~Z. Kasgari and W.~Saad, ``Stochastic optimization and control framework
  for 5g network slicing with effective isolation,'' in \emph{2018 52nd Annual
  Conference on Information Sciences and Systems (CISS)}.\hskip 1em plus 0.5em
  minus 0.4em\relax IEEE, 2018, pp. 1--6.

\bibitem{xu2021electrical}
K.~Xu, ``Electrical automation teaching based on vr virtual reality
  technology,'' in \emph{2021 2nd Asia-Pacific Conference on Image Processing,
  Electronics and Computers}, 2021, pp. 925--928.

\bibitem{sohaib2021network}
R.~M. Sohaib, O.~Onireti, Y.~Sambo, and M.~A. Imran, ``Network slicing for
  beyond 5g systems: An overview of the smart port use case,''
  \emph{Electronics}, vol.~10, no.~9, p. 1090, 2021.

\bibitem{liu2022slicing4meta}
Y.-J. Liu, H.~Du, D.~Niyato, G.~Feng, J.~Kang, and Z.~Xiong, ``Slicing4meta: An
  intelligent integration framework with multi-dimensional network resources
  for metaverse-as-a-service in web 3.0,'' \emph{arXiv preprint
  arXiv:2208.06081}, 2022.

\bibitem{chen2018virtual}
M.~Chen, W.~Saad, and C.~Yin, ``Virtual reality over wireless networks:
  Quality-of-service model and learning-based resource management,'' \emph{IEEE
  Transactions on Communications}, vol.~66, no.~11, pp. 5621--5635, 2018.

\bibitem{yang2020resource}
Y.~Yang, L.~Feng, C.~Zhang, Q.~Ou, and W.~Li, ``Resource allocation for virtual
  reality content sharing based on 5g d2d multicast communication,''
  \emph{EURASIP Journal on Wireless Communications and Networking}, vol. 2020,
  no.~1, pp. 1--12, 2020.

\bibitem{du2022attention}
H.~Du, J.~Liu, D.~Niyato, J.~Kang, Z.~Xiong, J.~Zhang, and D.~I. Kim,
  ``Attention-aware resource allocation and qoe analysis for metaverse xurllc
  services,'' \emph{arXiv preprint arXiv:2208.05438}, 2022.

\bibitem{dang2019joint}
T.~Dang and M.~Peng, ``Joint radio communication, caching, and computing design
  for mobile virtual reality delivery in fog radio access networks,''
  \emph{IEEE Journal on Selected Areas in Communications}, vol.~37, no.~7, pp.
  1594--1607, 2019.

\bibitem{chien2020q}
W.-C. Chien, H.-Y. Weng, and C.-F. Lai, ``Q-learning based collaborative cache
  allocation in mobile edge computing,'' \emph{Future generation computer
  systems}, vol. 102, pp. 603--610, 2020.

\bibitem{cai2022joint}
Y.~Cai, J.~Llorca, A.~M. Tulino, and A.~F. Molisch, ``Joint
  compute-caching-communication control for online data-intensive service
  delivery,'' \emph{arXiv preprint arXiv:2205.01944}, 2022.

\bibitem{Denim2021}
M.~Zhao, Y.~Zhou, J.~Meng, H.~Zheng, Y.~Cai, Y.~Shan, D.~Guan, and Z.~Yang,
  ``Virtual carbon and water flows embodied in global fashion trade - a case
  study of denim products,'' \emph{Journal of Cleaner Production}, vol. 303, p.
  127080, 2021.

\bibitem{GHGE2022}
\BIBentryALTinterwordspacing
U.~E.~P. Agency. Greenhouse gas emissions from a typical passenger vehicle.
  [Online]. Available:
  \url{https://www.epa.gov/greenvehicles/greenhouse-gas-emissions-typical-passenger-vehicle}
\BIBentrySTDinterwordspacing

\bibitem{WFP2022}
\BIBentryALTinterwordspacing
W.~F. Network. National water footprint. [Online]. Available:
  \url{https://waterfootprint.org/en/water-footprint/national-water-footprint/}
\BIBentrySTDinterwordspacing

\bibitem{Airline2022}
\BIBentryALTinterwordspacing
C.~Nugent. Airlines' emissions halved during the pandemic. can the industry
  preserve some of those gains as travel rebounds. [Online]. Available:
  \url{https://time.com/6048871/pandemic-airlines-carbon-emissions/}
\BIBentrySTDinterwordspacing

\bibitem{Concert2022}
\BIBentryALTinterwordspacing
M.~Rasmussen. Touring the musical metaverse: Virtual concerts are here to stay.
  [Online]. Available:
  \url{https://www.virtualhumans.org/article/touring-the-musical-metaverse-virtual-concerts-are-here-to-stay}
\BIBentrySTDinterwordspacing

\bibitem{Kamble2022}
S.~S. Kamble, A.~Gunasekaran, H.~Parekh, V.~Mani, A.~Belhadi, and R.~Sharma,
  ``Digital twin for sustainable manufacturing supply chains: Current trends,
  future perspectives, and an implementation framework,'' \emph{Technological
  Forecasting and Social Change}, vol. 176, p. 121448, 2022.

\bibitem{govind2021}
H.~S. Govindasamy, R.~Jayaraman, B.~Taspinar, D.~Lehner, and M.~Wimmer, ``Air
  quality management: An exemplar for model-driven digital twin engineering,''
  in \emph{ACM/IEEE International Conference on Model Driven Engineering
  Languages and Systems Companion (MODEL-C)}, 2021, pp. 229--232.

\bibitem{Lydon2021}
G.~P. Lydon, S.~Caranovic, I.~Hischier, and A.~Schlueter, ``Coupled simulation
  of thermally active building systems to support a digital twin,''
  \emph{Energy and Buildings}, vol. 202, p. 109298, 2021.

\bibitem{samad2021}
S.~M.~E. Sepasgozar, ``Differentiating digital twin from digital shadow:
  Elucidating a paradigm shift to expedite a smart, sustainable built
  environment,'' \emph{Buildings}, vol.~11, no.~4, p. 151, 2021.

\bibitem{zhao2021}
L.~Zhao, H.~Zhang, Q.~Wang, and H.~Wang, ``Digital-twin-based evaluation of
  nearly zero-energy building for existing buildings based on scan-to-bim,''
  \emph{Advances in Civil Engineering}, vol. 2021, no. 6638897, pp. 1--11,
  2021.

\bibitem{Moore2018}
F.~C. Moore, N.~Obradovich, F.~Lehner, and P.~Baylis, ``Rapidly declining
  remarkability of temperature anomalies may obscure public perception of
  climate change,'' \emph{Proceedings of the National Academy of Sciences},
  vol. 116, no.~11, pp. 4905--4910, 2019.

\bibitem{Markovits2021}
D.~M. Markowitz and J.~N. Bailenson, ``Virtual reality and the psychology of
  climate change,'' \emph{Current Opinion in Psychology}, vol.~42, pp. 60--65,
  2021.

\bibitem{Cryptopunk2021}
\BIBentryALTinterwordspacing
M.~Egkolfopoulou and A.~Gardner. Even in the metaverse, not all identities are
  created equal. [Online]. Available:
  \url{https://www.bloomberg.com/news/features/2021-12-06/cryptopunk-nft-prices-suggest-a-diversity-problem-in-the-metaverse}
\BIBentrySTDinterwordspacing

\bibitem{Jon2021}
\BIBentryALTinterwordspacing
J.~Radoff. The metaverse value-chain. [Online]. Available:
  \url{https://medium.com/building-the-metaverse/the-metaverse-value-chain-afcf9e09e3a7}
\BIBentrySTDinterwordspacing

\bibitem{duan2021Metaverse}
H.~Duan, J.~Li, S.~Fan, Z.~Lin, X.~Wu, and W.~Cai, ``Metaverse for social good:
  A university campus prototype,'' in \emph{Proceedings of the 29th ACM
  International Conference on Multimedia}, 2021, pp. 153--161.

\bibitem{du2022rethinking}
H.~Du, B.~Ma, D.~Niyato, and J.~Kang, ``Rethinking quality of experience for
  metaverse services: A consumer-based economics perspective,'' \emph{arXiv
  preprint arXiv:2208.01076}, 2022.

\bibitem{nevelsteen2018virtual}
K.~J. Nevelsteen, ``Virtual world, defined from a technological perspective and
  applied to video games, mixed reality, and the metaverse,'' \emph{Computer
  Animation and Virtual Worlds}, vol.~29, no.~1, p. e1752, 2018.

\bibitem{kiong2022Metaverse}
L.~V. Kiong, \emph{Metaverse Made Easy: A Beginner's Guide to the Metaverse:
  Everything you need to know about Metaverse, NFT and GameFi}.\hskip 1em plus
  0.5em minus 0.4em\relax Liew Voon Kiong, 2022.

\bibitem{gursoy2022Metaverse}
D.~Gursoy, S.~Malodia, and A.~Dhir, ``The metaverse in the hospitality and
  tourism industry: An overview of current trends and future research
  directions,'' \emph{Journal of Hospitality Marketing \& Management}, pp.
  1--8, 2022.

\bibitem{um2022travel}
T.~Um, H.~Kim, H.~Kim, J.~Lee, C.~Koo, and N.~Chung, ``Travel incheon as a
  metaverse: Smart tourism cities development case in korea,'' in \emph{ENTER22
  e-Tourism Conference}.\hskip 1em plus 0.5em minus 0.4em\relax Springer, Cham,
  2022, pp. 226--231.

\bibitem{ryskeldiev2018distributed}
B.~Ryskeldiev, Y.~Ochiai, M.~Cohen, and J.~Herder, ``Distributed metaverse:
  creating decentralized blockchain-based model for peer-to-peer sharing of
  virtual spaces for mixed reality applications,'' in \emph{Proceedings of the
  9th augmented human international conference}, 2018, pp. 1--3.

\bibitem{bennett2022remote}
D.~Bennett, ``Remote workforce, virtual team tasks, and employee engagement
  tools in a real-time interoperable decentralized metaverse.''
  \emph{Psychosociological Issues in Human Resource Management}, vol.~10,
  no.~1, 2022.

\bibitem{popescu2022virtual}
G.~H. Popescu, C.~F. Ciurl{\u{a}}u, C.~I. Stan \emph{et~al.}, ``Virtual
  workplaces in the metaverse: Immersive remote collaboration tools, behavioral
  predictive analytics, and extended reality technologies.''
  \emph{Psychosociological Issues in Human Resource Management}, vol.~10,
  no.~1, 2022.

\bibitem{ludlow2007second}
P.~Ludlow and M.~Wallace, \emph{The Second Life Herald: The virtual tabloid
  that witnessed the dawn of the metaverse}.\hskip 1em plus 0.5em minus
  0.4em\relax MIT press, 2007.

\bibitem{thomason2021metahealth}
J.~Thomason, ``Metahealth-how will the metaverse change health care?''
  \emph{Journal of Metaverse}, vol.~1, no.~1, pp. 13--16, 2021.

\bibitem{tan2022Metaverse}
T.~F. Tan, Y.~Li, J.~S. Lim, D.~V. Gunasekeran, Z.~L. Teo, W.~Y. Ng, and D.~S.
  Ting, ``Metaverse and virtual health care in ophthalmology: Opportunities and
  challenges,'' \emph{The Asia-Pacific Journal of Ophthalmology}, vol.~11,
  no.~3, pp. 237--246, 2022.

\bibitem{shin2022actualization}
D.~Shin, ``The actualization of meta affordances: Conceptualizing affordance
  actualization in the metaverse games,'' \emph{Computers in Human Behavior},
  vol. 133, p. 107292, 2022.

\bibitem{hamilton2022deep}
S.~Hamilton, ``Deep learning computer vision algorithms, customer engagement
  tools, and virtual marketplace dynamics data in the metaverse economy.''
  \emph{Journal of Self-Governance \& Management Economics}, vol.~10, no.~2,
  2022.

\bibitem{mackenzie2022criminology}
S.~Mackenzie, ``Criminology towards the metaverse: Cryptocurrency scams, grey
  economy and the technosocial,'' \emph{The British Journal of Criminology},
  2022.

\bibitem{mystakidis2022Metaverse}
S.~Mystakidis, ``Metaverse,'' \emph{Encyclopedia}, vol.~2, no.~1, pp. 486--497,
  2022.

\bibitem{fernandez2022facebook}
P.~Fernandez, ``Facebook, meta, the metaverse and libraries,'' \emph{Library Hi
  Tech News}, 2022.

\bibitem{kim2021advertising}
J.~Kim, ``Advertising in the metaverse: Research agenda,'' \emph{Journal of
  Interactive Advertising}, vol.~21, no.~3, pp. 141--144, 2021.

\bibitem{cheng2022will}
R.~Cheng, N.~Wu, S.~Chen, and B.~Han, ``Will metaverse be nextg internet?
  vision, hype, and reality,'' \emph{arXiv preprint arXiv:2201.12894}, 2022.

\bibitem{vidal2022new}
D.~Vidal-Tom{\'a}s, ``The new crypto niche: Nfts, play-to-earn, and metaverse
  tokens,'' \emph{Finance Research Letters}, p. 102742, 2022.

\bibitem{ante2021non}
L.~Ante, ``Non-fungible token (nft) markets on the ethereum blockchain:
  Temporal development, cointegration and interrelations,'' \emph{Available at
  SSRN 3904683}, 2021.

\bibitem{fourati2021comprehensive}
H.~Fourati, R.~Maaloul, L.~Chaari, and M.~Jmaiel, ``Comprehensive survey on
  self-organizing cellular network approaches applied to 5g networks,''
  \emph{Computer Networks}, vol. 199, p. 108435, 2021.

\bibitem{vittal2021self}
S.~Vittal \emph{et~al.}, ``Self optimizing network slicing in 5g for slice
  isolation and high availability,'' in \emph{2021 17th International
  Conference on Network and Service Management (CNSM)}.\hskip 1em plus 0.5em
  minus 0.4em\relax IEEE, 2021, pp. 125--131.

\bibitem{nadir2021immersive}
Z.~Nadir, T.~Taleb, H.~Flinck, O.~Bouachir, and M.~Bagaa, ``Immersive services
  over 5g and beyond mobile systems,'' \emph{IEEE Network}, vol.~35, no.~6, pp.
  299--306, 2021.

\bibitem{rinaldi2020non}
F.~Rinaldi, H.-L. Maattanen, J.~Torsner, S.~Pizzi, S.~Andreev, A.~Iera,
  Y.~Koucheryavy, and G.~Araniti, ``Non-terrestrial networks in 5g \& beyond: A
  survey,'' \emph{IEEE access}, vol.~8, pp. 165\,178--165\,200, 2020.

\bibitem{lin2022next}
B.~Lin, J.~Duan, M.~Han, and L.~X. Cai, ``Next generation marine wireless
  communication networks,'' pp. 1--143, 2022.

\bibitem{du2021optimal}
H.~Du, D.~Niyato, J.~Kang, D.~I. Kim, and C.~Miao, ``Optimal targeted
  advertising strategy for secure wireless edge metaverse,'' \emph{arXiv
  preprint arXiv:2111.00511}, 2021.

\bibitem{ghafoor2020mac}
S.~Ghafoor, N.~Boujnah, M.~H. Rehmani, and A.~Davy, ``Mac protocols for
  terahertz communication: A comprehensive survey,'' \emph{IEEE Communications
  Surveys \& Tutorials}, vol.~22, no.~4, pp. 2236--2282, 2020.

\bibitem{zawish2022towards}
M.~Zawish, N.~Ashraf, R.~I. Ansari, S.~Davy, H.~K. Qureshi, N.~Aslam, and S.~A.
  Hassan, ``Towards on-device ai and blockchain for 6g enabled agricultural
  supply-chain management,'' \emph{arXiv preprint arXiv:2203.06465}, 2022.

\bibitem{zawish2022complexity}
M.~Zawish, S.~Davy, and L.~Abraham, ``Complexity-driven cnn compression for
  resource-constrained edge ai,'' \emph{arXiv preprint arXiv:2208.12816}, 2022.

\bibitem{han2021abstracted}
B.~Han, W.~Jiang, M.~A. Habibi, and H.~D. Schotten, ``An abstracted survey on
  6g: Drivers, requirements, efforts, and enablers,'' \emph{arXiv preprint
  arXiv:2101.01062}, 2021.

\bibitem{chen2018novel}
T.~Chen, M.~Zhao, Q.~Shi, Z.~Yang, H.~Liu, L.~Sun, J.~Ouyang, and C.~Lee,
  ``Novel augmented reality interface using a self-powered triboelectric based
  virtual reality 3d-control sensor,'' \emph{Nano Energy}, vol.~51, pp.
  162--172, 2018.

\bibitem{Eshghie2022}
M.~Eshghie, L.~Quan, G.~A. Kasche, F.~Jacobson, C.~Bassi, and C.~Artho,
  ``Circlechain: Tokenizing products with a role-based scheme for a circular
  economy,'' \emph{arXiv preprint arXiv:2205.11212}, 2022.

\bibitem{Deveci2022}
M.~Deveci, A.~R. Mishra, I.~Gokasar, P.~Rani, D.~Pamucar, and E.~Ozcan, ``A
  decision support system for assessing and prioritizing sustainable urban
  transportation in metaverse,'' \emph{IEEE Transactions on Fuzzy Systems},
  vol. Early Access, pp. 1--10, 2022.

\bibitem{PrivateAI}
S.~A. Khowaja, K.~Dev, N.~M.~F. Qureshi, P.~Khuwaja, and L.~Foschini, ``Toward
  industrial private ai: A two-tier framework for data and model security,''
  \emph{IEEE Wireless Communications}, vol.~29, no.~2, pp. 76--83, 2022.

\bibitem{PGSL}
S.~A. Khowaja, I.~H. Lee, K.~Dev, M.~A. Jarwar, and N.~M.~F. Qureshi, ``Get
  your foes fooled: Proximal gradient split learning for defense against model
  inversion attacks on iomt data,'' \emph{IEEE Transactions on Network Science
  and Engineering}, vol. Early Access, pp. 1--10, 2022.

\bibitem{zhang2019edge}
Y.~Zhang, X.~Ma, J.~Zhang, M.~S. Hossain, G.~Muhammad, and S.~U. Amin, ``Edge
  intelligence in the cognitive internet of things: Improving sensitivity and
  interactivity,'' \emph{IEEE Network}, vol.~33, no.~3, pp. 58--64, 2019.

\bibitem{khan2020network}
L.~U. Khan, I.~Yaqoob, N.~H. Tran, Z.~Han, and C.~S. Hong, ``Network slicing:
  Recent advances, taxonomy, requirements, and open research challenges,''
  \emph{IEEE Access}, vol.~8, pp. 36\,009--36\,028, 2020.

\bibitem{lee2022virtual}
H.~Lee, D.~Woo, and S.~Yu, ``Virtual reality metaverse system supplementing
  remote education methods: Based on aircraft maintenance simulation,''
  \emph{Applied Sciences}, vol.~12, no.~5, p. 2667, 2022.

\bibitem{tayal2022virtual}
S.~Tayal, K.~Rajagopal, and V.~Mahajan, ``Virtual reality based metaverse of
  gamification,'' in \emph{2022 6th International Conference on Computing
  Methodologies and Communication (ICCMC)}.\hskip 1em plus 0.5em minus
  0.4em\relax IEEE, 2022, pp. 1597--1604.

\bibitem{kye2021educational}
B.~Kye, N.~Han, E.~Kim, Y.~Park, and S.~Jo, ``Educational applications of
  metaverse: possibilities and limitations,'' \emph{Journal of Educational
  Evaluation for Health Professions}, vol.~18, 2021.

\bibitem{wu2021toward}
J.~Wu, R.~Li, X.~An, C.~Peng, Z.~Liu, J.~Crowcroft, and H.~Zhang, ``Toward
  native artificial intelligence in 6g networks: System design, architectures,
  and paradigms,'' \emph{arXiv preprint arXiv:2103.02823}, 2021.

\end{thebibliography}

\end{document}